\RequirePackage{fix-cm}
\documentclass[smallextended]{svjour3}       
\smartqed  
\usepackage{graphicx}
\usepackage{makeidx}  
\usepackage{amsmath} 
\usepackage{amssymb}
\usepackage{enumerate}
\usepackage{algorithm}
\usepackage{algpseudocode}
\usepackage{graphicx} 
\usepackage{caption}
\usepackage{subcaption}
\usepackage{wrapfig}
\usepackage{sidecap}
\usepackage{framed}
\usepackage{bm}
\usepackage{multirow}
\usepackage{ctable}
\usepackage{empheq}
\usepackage{enumitem}

\newenvironment{Proof}[1]{\par\noindent\textit{Proof:}\space#1}{\hfill $\blacksquare$}

\DeclareMathOperator*{\argmin}{arg\,min}
\DeclareMathOperator*{\argmax}{arg\,max}
\def\bbbr{{\rm I\!R}}
\def\bbbh{{\rm I\!H}}

\def\bbbone{{\mathchoice {\rm 1\mskip-4mu l} {\rm 1\mskip-4mu l}
{\rm 1\mskip-4.5mu l} {\rm 1\mskip-5mu l}}}
\newcommand{\overbar}[1]{\mkern 1.5mu\overline{\mkern-1.5mu#1\mkern-1.5mu}\mkern 1.5mu}

\begin{document}

\title{An Incremental Off-policy Search in a Model-free Markov Decision Process Using a Single Sample Path
}


\author{Ajin George Joseph         \and
        Shalabh Bhatnagar 
}


\institute{Ajin George Joseph \at
              Indian Institute of Science, Bangalore, INDIA, 560012. \\
              Tel.: +91-80-22932368\\
              Fax:  +91-80-23602911\\
              \email{ajin@csa.iisc.ernet.in}           
           \and
           Shalabh  Bhatnagar \at
              Indian  Institute of Science, Bangalore, INDIA, 560012.
}


\maketitle

\begin{abstract}
In this paper, we consider a modified version of the control problem in a model free Markov decision process (MDP) setting with large state and action spaces. The control problem most commonly addressed in the contemporary literature is to find an optimal policy which maximizes the value function, \emph{i.e.}, the long run discounted reward of the MDP. The current settings also assume access to a generative model of the MDP with the hidden premise that observations of the system behaviour in the form of sample trajectories can be obtained with ease from the model. In this paper, we consider a modified version, where the cost function is the expectation of a non-convex function of the value function without access to the generative model. Rather, we assume that a sample trajectory generated using a priori chosen behaviour policy is made available. In this restricted setting, we solve the modified control problem in its true sense, \emph{i.e.}, to find the best possible  policy given this limited information. We propose a stochastic approximation algorithm based on the well-known cross entropy method which is data (sample trajectory) efficient, stable, robust as well as computationally and storage efficient. We provide a proof of convergence of our algorithm to a policy which is globally optimal relative to the behaviour policy. We also present experimental results to corroborate our claims and we demonstrate the superiority of the solution produced by our algorithm compared to the state-of-the-art algorithms under appropriately chosen behaviour policy.
\end{abstract}
\section{Summary of Notation}\label{sec:summary}
\vspace*{0mm}
We use $\mathbf{x}$ for random variable and $x$ for deterministic variable. For set $A$,  $I_{A}$ represents the indicator function of $A$, \emph{i.e.}, $I_{A}(x) = 1$ if $x \in A$ and $0$ otherwise. Let $f_{\theta}(\cdot)$ denote the \textit{probability density function} parametrized by $\theta$. Let $\mathbb{E}_{\theta}[\cdot]$ and $P_{\theta}$ denote the \textit{expectation} and the induced \textit{probability measure} \emph{w.r.t.} $f_{\theta}$. For $\rho \in (0,1)$, let $\gamma_{\rho}(J, \theta)$ denote the $(1-\rho)$-quantile of $J(\mathbf{x})$ \emph{w.r.t.} $f_{\theta}$, \emph{i.e.},
\begin{equation}
\gamma_{\rho}(J, \theta) \triangleq \sup\{l: P_{\theta}(J(\mathbf{x}) \geq l) \geq \rho \}.
\end{equation}
Let $supp(f) \triangleq \overline{\{x | f(x) \neq 0\}}$ denote the support of $f$ \normalsize and $interior(A)$ be the \textit{interior} of set $A$. Let $\mathcal{N}_{d}(a, B)$ represent the multivariate Gaussian distribution with mean vector $a$ and covariance matrix $B$. A function $L:\bbbr^{m} \rightarrow \bbbr$ is \textit{Lipschitz continuous}, if $\exists K \geq 0$ \emph{s.t.} $\vert L(x) - L(y) \vert \leq K\Vert x - y \Vert$, $\forall x, y \in \bbbr^{m}$, where $\Vert \cdot \Vert$ is a norm defined on $\bbbr^{m}$. Also, for a matrix $A = [a_{ij}]_{1 \leq i \leq m, 1 \leq j \leq n} \in \bbbr^{m \times n}$, we define the norm $\Vert A \Vert_{\infty} \triangleq \max_{1 \leq i \leq m} \sum_{1 \leq j \leq n}\vert a_{ij} \vert$ and for invertible matrices, we define the \textit{condition number} $\kappa(A) \triangleq \Vert A \Vert_{\infty} \Vert A^{-1} \Vert_{\infty}$. Also, $\vert A \vert \triangleq [\vert a_{ij} \vert]_{1 \leq i \leq m, 1 \leq j \leq n}$. Similarly, for $x \in \bbbr^{m}$, the sup norm $\Vert x \Vert_{\infty}$ is defined as $\Vert x \Vert_{\infty} \triangleq \sup_{i}{\vert x_i \vert}$ and $\vert x \vert \triangleq (\vert x_i \vert)_{1 \leq i \leq m}$. 
\section{Introduction and Preliminaries}\label{sec:intro}
A discrete time Markov decision process (MDP) \cite{sutton1998reinforcement,bertsekas1995dynamic} is a 4-tuple ($\mathbb{S}$, $\mathbb{A}$, $R$, $P$), where $\mathbb{S}$ denotes the set of \textit{states} and $\mathbb{A}$ is the set of \textit{actions}. Also, $R: \mathbb{S} \times \mathbb{A} \times \mathbb{S} \rightarrow \bbbr$ is the \textit{reward function} where $R(s, a, s^{'})$ represents the reward obtained in state $s$ after taking action $a$ and transitioning to state $s^{'}$. Without loss of generality, we assume that the same choice of actions is available for all the states. We also assume that the reward function is bounded, \emph{i.e.}, $\Vert R \Vert_{\infty} < \infty$. We let $P:\mathbb{S} \times \mathbb{A} \times \mathbb{S} \rightarrow [0,1]$ denote the \textit{transition probability kernel}, where $P(s, a, s^{'})$ is the probability of next state being $s^{'}$ conditioned on the fact that the current state is $s$ and action taken is $a$. We assume that the state and action spaces are finite. A \textit{stationary random policy} (SRP) $\pi(\cdot \vert s)$ is a probability distribution over the action space $\mathbb{A}$ conditioned on state $s \in \mathbb{S}$. A given policy $\pi$ along with the transition kernel $P$ determines the state dynamics of the system. For a given policy $\pi$, the system behaves as a homogeneous Markov chain with transition probabilities 
\begin{equation}\label{eq:pmstoch}
P_{\pi}(s,s^{'}) = \sum_{a \in \mathbb{A}}\pi(a \vert s)P(s, a, s^{'}), s, s^{'} \in \mathbb{S}.
\end{equation}
In this paper, we  consider only stationary randomized policies. We also assume that given a SRP $\pi$, the Markov chain induced  by $P_{\pi}$ is ergodic, \emph{i.e.}, the Markov chain is irreducible and aperiodic.\vspace*{0mm}\\

The two fundamental questions most commonly addressed in the MDP literature are: $1$. \textit{Prediction problem} and $2$. \textit{Control problem}.\vspace*{2mm}\\
\textbf{Prediction problem: }
For a given SRP $\pi$ and \textit{discount factor }$\gamma \in (0,1)$, the objective is to evaluate the long-run $\gamma$-discounted cost $V^{\pi} \in \bbbr^{\vert \mathbb{S} \vert}$ which is defined as
\begin{equation}\label{eqn:discost}
V^{\pi}(s) \triangleq \mathbb{E}_{\pi}\left[\sum_{k=0}^{\infty}\gamma^{k}R(\mathbf{s}_{k}, \mathbf{a}_{k}, \mathbf{s}_{k+1}) \Big\vert s_{0} = s\right], \hspace*{3mm} s \in \mathbb{S},
\end{equation}
where the random variable $\mathbf{s}_{k}$ represents the state at instant $k$, the random variable $\mathbf{a}_{k}$ represents the action chosen at instant $k$ and the random variable $\mathbf{s}_{k+1}$ represents the transitioned state after instant $k$, \emph{i.e.}, the state at instant $k+1$. Further, $\mathbb{E}_{\pi}[\cdot]$ is the expectation \emph{w.r.t.} the probability distribution induced by $P_{\pi}$ with initial state $\mathbf{s}_{0} = s$. Note that the cost evaluation in (\ref{eqn:discost}) is realistic and prudent. Since MDP is a sequential decision making paradigm, the discount factor $\gamma$ controls the width of the window of future events to be considered to guide the decision process. For $\gamma$ close to $0$, only the rewards pertaining to the first few transitions count as the effect of the future rewards whose weights are geometric in $\gamma$ is minimal. However, the case of $\gamma$ very close to $1$ requires a very long window to be considered. 

For a given policy $\pi$, the value function $V^{\pi}$ satisfies the following \textit{Bellman equation} (written in vector-matrix notation):
\begin{equation}\label{eq:bellman}
V^{\pi} = T^{\pi}V^{\pi},
\end{equation}
where $T^{\pi}$ called the \textit{Bellman operator} is defined as $T^{\pi}V \triangleq R^{\pi} + \gamma P_{\pi} V$ and $R^{\pi}(s) \triangleq \sum_{a \in \mathbb{A}}\pi(a \vert s)\sum_{s^{\prime} \in \mathbb{S}}P(s, a, s^{\prime})R(s, a, s^{\prime})$. Hence $V^{\pi}$ can be directly computed as $V^{\pi} = (I-\gamma P_{\pi})^{-1}R^{\pi}$. The computational complexity of the above direct computation is $O(\vert \mathbb{S} \vert^{3})$ and the space complexity is $O(\vert \mathbb{S} \vert^{2})$. An alternate procedure to solve the prediction problem is \textit{value iteration} that is based on the contraction mapping theorem. It is easy to see that the Bellman operator $T^{\pi}$ is a contraction mapping with the contraction constant $\gamma$. Hence by the contraction mapping theorem, $(T^{\pi})^{k}V \rightarrow V^{\pi}$ as $k \rightarrow \infty$, $\forall V \in \bbbr^{\vert \mathbb{S} \vert}$. The computational complexity of this successive approximation procedure is $O(\vert \mathbb{S} \vert^{2})$ per iteration and the space complexity is $O(\vert \mathbb{S} \vert^{2})$ as well. The state space $\mathbb{S}$ can be huge, for example, in cases where the state is represented as a high-dimensional vector. The cardinality of the state space in such a case is exponential in the dimension resulting in a corresponding exponential upsurge in computational effort and storage requirement. In such cases, the above method can become well-nigh intractable. This predicament is referred to in the literature as the \textit{curse of dimensionality}. One commonly employed heuristic to circumvent the curse is the \textit{state aggregation} \cite{bertsekas1989adaptive} technique. However, it also suffers dearly when the state space is huge.\vspace*{2mm}\\
\textbf{Control problem: } The objective for this problem is to find the optimal stationary policy $\pi^{*}$ of the MDP, where
\begin{equation} 
\pi^{*}(s) \in \argmax_{\pi}V^{\pi}(s), \hspace*{3mm} s \in \mathbb{S}.
\end{equation}
The existence of an optimal stationary policy is proven in \cite{puterman2014markov}. The optimal value function $V^{*} (= V^{\pi^{*}})$ satisfies the \textit{Bellman optimality equation} given by: $T V^{*} = V^{*}$, where the \textit{Bellman optimality operator} $T$ is defined as $T V(s) \triangleq max_{a \in \mathbb{A}}{\sum_{s^\prime \in \mathbb{S}}P(s, a, s^\prime)(R(s, a, s^\prime) + \gamma V(s^\prime))}$. The primary numerical methods which solve the control problem are the \textit{value iteration} and \textit{policy iteration}. A detailed description of these methods is available in \cite{puterman2014markov}. In a nutshell, policy iteration can be characterized as generating a sequence of improving policies $\{\pi_k\}_{k \in \mathbb{N}}$ with $\pi_k$ converging to $\pi^{*}$ after a finite number of steps. Value iteration on the other hand involves repeated application of the Bellman optimality operator, which requires  multiple extensive passes over the state space and the convergence is only guaranteed asymptotically. The computational complexities of policy iteration and value iteration are $O(\vert \mathbb{S} \vert^{2}\vert \mathbb{A} \vert + \vert \mathbb{S} \vert^{3})$ and $O(\vert \mathbb{S} \vert^{2} \vert \mathbb{A} \vert)$ respectively. The space complexity of both the methods is the same and it is $O(\vert \mathbb{S} \vert + \vert \mathbb{A} \vert)$. The super-linear dependency of the methods on the size of state space results in the curse of dimensionality. A recently proposed policy iteration method based on stochastic factorization \cite{barreto2014policy}  has reduced the dependency to linear terms. However, when $\mathbb{S}$ is very large, stochastic factorization also becomes intractable.
\subsection{Model Free Algorithms}
In the above section, the  prediction and control algorithms are numerical methods that assume that the probability transition function $P$ and the reward function $R$ are available. In most of the practical scenarios, it is unrealistic to assume that accurate knowledge of $P$ and $R$ is realizable. However, the behaviour of the system can be observed and one needs to either predict the value of a given policy or find the optimal control using the available observations. The observations are in the form of a sample trajectory $\{s_0, a_0, r_0, s_1, a_1, r_1, s_2, a_2, \dots \}$, where $s_i \in \mathbb{S}$ is the state and $r_i = R(s_i, a_i, s_{i+1})$ is the immediate reward at time instant $i$. Model free algorithms are basically of three types: (i) Indirect methods, (ii) Direct methods and (iii) Policy search methods. The last of these methods searches in the policy space to find the optimal policy where the performance measure used for comparison is the estimate of the value function induced from the observations. Prominent algorithms in this category are actor-critic \cite{konda2003onactor}, policy gradient \cite{baxter2001infinite}, natural actor-critic \cite{bhatnagar2009natural} and fast policy search \cite{mannor2003cross}. Indirect methods are based on the certainty equivalence of computing where initially the transition matrix and the expected reward vector are estimated using the observations and subsequently, model based approaches mentioned in the above section are applied on the estimates. A few indirect methods are control learning \cite{sato1982learning,sato1988learning,kumar1982optimal}, priority sweeping \cite{moore1993prioritized}, adaptive real-time dynamic programming (ARTDP) \cite{barto1995learning} and PILCO \cite{deisenroth2011pilco}. For the case of direct methods which are more appealing, the model is not  estimated, rather the control policy is adapted iteratively using a shadow utility function derived from the  instantiation of the internal dynamics of the MDP. The algorithms in this class are generally referred to in the literature as the \textit{reinforcement learning} algorithms. Prominent reinforcement learning algorithms include \textit{temporal difference (TD) learning} \cite{sutton1988learning} (prediction method), \textit{Q-learning} \cite{watkins1989learning} and \textit{SARSA} \cite{singh1996reinforcement} (control methods). There are two variants of the prediction algorithm depending on how the sample trajectory is generated. They are \textit{on-policy} and \textit{off-policy} algorithms. In the on-policy case, the sample trajectory is generated using the policy $\pi$ which is being evaluated, \emph{i.e.},  $s_{i+1} \sim P(s_i, a_i, \cdot)$, where $a_i \sim \pi(\cdot \vert s_{i})$ and $r_i = R(s_i, a_i, s_{i+1})$. In the off-policy case, the sample trajectory is generated using a policy $\pi_b$ which is possibly different from the policy $\pi$ that is being evaluated, \emph{i.e.}, $s_{i+1} \sim P(s_i, a_i, \cdot)$, where $a_i \sim \pi_b(\cdot \vert s_{i})$ and $r_i = R(s_i, a_i, s_{i+1})$. 

Model free algorithms are shown to be robust, stable and exhibit good convergence behaviour under realistic assumptions. However, they suffer from the curse of dimensionality which arises due to the space complexity. Note that the space complexity of the above mentioned learning algorithms is $O(\vert \mathbb{S} \vert)$, 
which becomes unmanageably large with increasing state space.
\subsection{Linear Function Approximation (LFA) Methods for Model Free Markov Decision Process}
To tackle the curse of dimensionality and to achieve tractability, it is imperative to eliminate the dependency both in terms of the computational and storage requirements of the learning methods on the cardinalities of state and action spaces. An efficient approach is to compactly yet effectively represent the system in a lower $k_1$-dimensional space, where $k_1 \ll \vert \mathbb{S} \vert$. A well understood dimensionality reduction technique is the linear function approximation. 
Here, we choose a collection of {\em prediction features} $\{\phi_i\}_{i=1}^{k_1}$, where $\phi_i \in \bbbr^{\vert \mathbb{S} \vert}$. In this case, the prediction task becomes a projection where
\begin{equation}
\Pi V^{\pi} = \argmin_{h \in \bbbh^{\Phi}} \Vert V^{\pi} - h \Vert^{2},
\end{equation}
where $\bbbh^{\Phi} \triangleq \{\Phi x \vert x \in \bbbr^{k_1}\} \subset \bbbr^{\vert \mathbb{S} \vert}$ is the space of representable functions with $\Phi \triangleq (\phi_1, \dots, \phi_{k_1}) \in \bbbr^{\vert \mathbb{S} \vert \times k_1}$ and the norm $\Vert \cdot \Vert$ is chosen appropriately  according to the domain. Note that $\bbbh^{\Phi}$ is a linear function space. Further, we define $\phi(s) \triangleq (\phi_1(s), \dots, \phi_{k_{1}}(s))^{\top}$, $s \in \mathbb{S}$. Note that $\phi_i$ can be viewed as a function from $\mathbb{S}$ to $\bbbr$. Similarly, the control problem becomes $\pi^{*}(s) \in \argmax_{\pi}\Pi V^{\pi}(s), \hspace*{1mm} \forall s \in \mathbb{S}$. Note that in the case of large and complex MDPs, the features are not hard-coded, instead one employs compact representations in the form of basis functions. Examples of basis functions include radial basis functions and Fourier basis.

To address the computational and storage concerns arising due to large action space, a sagacious approach is to employ a parametrized class of SRPs $\{\pi_{w} \vert w \in \mathbb{W} \subset \bbbr^{k_{2}}\}$, where $k_{2} \in \mathbb{N}$, instead of an exact representation. The most commonly used is the Gibbs (or Boltzmann) ``soft-max" class of policies. In this case, for a given $w \in \mathbb{W} \subset \bbbr^{k_{2}}$, the SRP $\pi_{w}$ is defined as
\begin{equation}\label{eq:stplcy}
\pi_{w}(a | s) = \frac{\exp{(w^{\top}\psi(s, a)/\tau)}}{\sum_{b \in \mathbb{A}}\exp{(w^{\top}\psi(s,b)/\tau)}},
\end{equation}
where $\{\psi(s, a) \in \bbbr^{k_{2}} \vert s \in \mathbb{S}, a \in \mathbb{A}\}$ is a given {\em policy feature set } and $\tau \in \bbbr_{+}$ is fixed \textit{a priori}.

The accuracy of the function approximation method depends on the representational/expressive ability of $\bbbh^{\Phi}$. For example, when $k_{1} = \vert \mathbb{S} \vert$, the representational ability is utmost, since $\bbbh = \bbbr^{\vert \mathbb{S} \vert}$. In general, $k_1 \ll n$ and hence $\bbbh \subset \bbbr^{\vert \mathbb{S} \vert}$. So for an arbitrary policy $\pi$, where $V^{\pi} \notin \bbbh^{\Phi}$, the prediction of the value function $V^{\pi}$ shall always incur an unavoidable \textit{approximation error} ($e_{appr}$) given by $\inf_{h \in \bbbh^{\Phi}}\Vert V^{\pi} - h \Vert$. Given $\bbbh^{\Phi}$, one cannot perform no better than $e_{appr}$. The prediction features $\{\phi_i\}$ are hand-crafted using prior domain knowledge and their choice is critical in approximating the value function. There is an abundance of literature available on the topic. In  this paper, we assume that an appropriately chosen feature set is available \textit{a priori}. Also note that the convergence of the prediction methods is in asymptotic sense. But in most practical scenarios, the algorithm has to be terminated after a finite number of steps. This incurs an \textit{estimation error} ($e_{est}$) which however decays to zero, asymptotically.

Even though LFA produces sub-optimal solutions, since the search is conducted on a restricted subspace of $\bbbr^{\vert \mathbb{S} \vert}$, it yields large computational and storage benefits. So some degree of trade-off between accuracy and tractability is indeed unavoidable.
\subsection{Off-Policy Prediction Using LFA}\label{subsec:offplcy}
\noindent
\textbf{Setup: }Given $w, w_b \in \mathbb{W}$ and an observation of the system dynamics in the form of a sample trajectory $\{s_{0}, a_0, r_{0}, s_{1}, a_1, r_{1}, s_{2}, \dots \}$, where at each instant $k$, $a_{k} \sim \pi_{w_b}(\cdot \vert s_{k})$, $s_{k+1} \sim $ $P(s_k, a_k, \cdot)$ and $r_{k}$ = $R(s_{k}, a_{k}, s_{k+1})$, the goal is to estimate the value function $V^{\pi_{w}}$ of the target policy $\pi_{w}$ (that is possibly different from $\pi_{w_b}$). We assume that the Markov chains defined by $P_{w}$ and $P_{w_b}$ are ergodic. Further, let $\nu_{w}$ and $\nu_{w_b}$ be the stationary distributions of the Markov chains with transition probability matrices $P_{w}$ and $P_{w_b}$ respectively, \emph{i.e.}, $\lim_{k \rightarrow \infty}P_{w}(\mathbf{s}_k = s) = \nu_{w}(s)$ and $\nu_{w}^{\top}P_{w} = \nu_{w}^{\top}$ and likewise for $\nu_{w_b}$. Note that for brevity the notations have been simplified here, \emph{i.e.}, $P_{w} \triangleq P_{\pi_{w}}$ and $P_{w_b} \triangleq P_{\pi_{w_b}}$. We follow the new notation for the rest of the paper. Similarly, $V^{w} \triangleq V^{\pi_w}$.

In the off-policy learning case, the projection is \emph{w.r.t.} the norm $\Vert \cdot \Vert_{\nu_{w_b}}$, where $\Vert V \Vert^{2}_{\nu_{w_b}} = <V, V>_{\nu_{w_b}}$. The inner product is defined as $<V_1, V_2>_{\nu} = V_{1}^{\top}D^{\nu}V_2$, where $V_1, V_2 \in \bbbr^{\vert \mathbb{S} \vert}, \nu \in [0,1]^{\vert \mathbb{S} \vert}$ is a probability mass function over $\mathbb{S}$ and $D^{\nu}$ is a $\vert \mathbb{S} \vert \times \vert \mathbb{S} \vert$ diagonal matrix with $D^{\nu}_{ii} = \nu(i)$, $1 \leq i \leq \vert \mathbb{S} \vert$.
Thus the norm $\Vert \cdot \Vert_{\nu_{w_b}}$ is in fact the Euclidean norm weighted with the stationary distribution $\nu_{w_b}$ of the behaviour policy $\pi_b$, \emph{i.e.}, $\Vert V \Vert_{\nu_{w_b}} \triangleq \sqrt{\sum_{s \in \mathbb{S}}\nu_{w_b}(s)V^{2}(s)}$. So
\begin{equation}\label{eq:hproj}
h_{w | w_b} \triangleq \Pi^{w_b}V^{w} = \argmin_{h \in \bbbh^{\Phi}} \Vert V^{w} - h \Vert_{\nu_{w_b}}^{2},
\end{equation}
where $\Pi^{w_b}$ denotes the projection operator \emph{w.r.t.} $\Vert \cdot \Vert_{\nu_{w_b}}$ whose closed form expression can be derived as follows:
\begin{equation*}
\begin{aligned}
\nabla_{x}\Vert V^{w} - & h \Vert_{\nu_{w_b}}^{2} = 0\\
&\Rightarrow \nabla_x(V^{w} - \Phi x)^{\top}D^{\nu_{w_b}}(V^{w} - \Phi x) = 0\\
&\Rightarrow \Phi^{\top}D^{\nu_{w_b}}(V^{w} - \Phi x) = 0\\
&\Rightarrow \Phi^{\top}D^{\nu_{w_b}}\Phi x  = \Phi^{\top}D^{\nu_{w_b}}V^{w}\\
&\Rightarrow x = (\Phi^{\top}D^{\nu_{w_b}}\Phi)^{-1}\Phi^{\top}D^{\nu_{w_b}}V^{w} \\
&\Rightarrow \Phi x = \Phi(\Phi^{\top}D^{\nu_{w_b}}\Phi)^{-1}\Phi^{\top}D^{\nu_{w_b}}V^{w}.
\end{aligned}
\end{equation*}
\begin{equation}
\therefore \Pi^{w_b} = \Phi(\Phi^{\top} D^{\nu_{w_b}} \Phi)^{-1}\Phi^{\top}D^{\nu_{w_b}}.\hspace*{3cm}
\end{equation}
$\circledast$ \textbf{Assumption (A1)} \textit{The prediction features $\{\phi_i\}_{i=1}^{k_1}$ are linearly independent.}\vspace*{2mm}\\
\textbf{Algorithms: }The evaluation of $\Pi^{w_b}$ requires knowledge of the stationary distribution $\nu_{w_b}$ which can only be derived if the transition matrix $P_{w_b}$ is available. However, in model free learning $P_{w_b}$ is hidden and hence all the state-of-the-art methods can only derive an approximation to the projection. Two pertinent algorithms are \textit{off-policy TD($\lambda$)} and \textit{off-policy LSTD($\lambda$)}. The algorithms return a prediction vector $x \in \bbbr^{k_{1}}$ \emph{s.t.} $\Phi x \approx h_{w \vert w_b}$. The major technique used in both the algorithms is to correct the discrepancies between the target and behaviour policies using \textit{importance sampling} \cite{glynn1989importance}. Here we introduce the sampling ratio at time $k$ to be $\rho_{k} \triangleq \frac{\pi_{w}(a_{k} \vert s_{k})}{\pi_{w_b}(a_{k} \vert s_{k})}$, where we use the convention $0/0 = 0$. \vspace*{2mm}\\
$\bullet$ \textbf{Off-policy TD($\lambda$):} Off-policy TD($\lambda)$ \cite{yu2012least,yu2015convergence}, where $\lambda \in [0,1]$ is one of the fundamental algorithms to approximate value function using linear architecture. The algorithm is defined as follows:
\begin{subequations}\label{eqn:tdalg}
\begin{align}
&\mathbf{x}_{k+1} := \mathbf{x}_k + \alpha_{k+1}\delta_{k+1}(\mathbf{x}_k)\mathbf{e}_{k},\\
&\mathbf{e}_{k+1} := \gamma\lambda \rho_{k} \mathbf{e}_k + \phi(s_k), \hspace*{8mm}
\end{align}
\end{subequations}
where $\mathbf{e}_{k}, \mathbf{x}_{k} \in \bbbr^{k_{1}}$ and $\delta_{k+1} \triangleq \rho_{k}r_{k} + \gamma\rho_{k}\mathbf{x}_{k}^{\top}\phi(s_{k+1}) - \mathbf{x}_{k}^{\top}\phi(s_k)$ is called the \textit{temporal difference error}. The learning rate $\alpha_{k}$ is non-negative, deterministic and satisfies $\sum_{k} \alpha_k = \infty$, $\sum_{k} \alpha_{k}^{2} < \infty$. The vector $\mathbf{e}_{k} \in \bbbr^{k_{1}}$  is called the \textit{eligibility trace} and is used for variance reduction. Eligibility traces accelerate the learning process by integrating temporal differences from multiple time steps. The convergence analysis of the off-policy TD($\lambda$) method is provided in \cite{yu2012least}. However, the analysis assumes that the iterates $\mathbf{x}_{k} \in \bar{B}_{r}(0), \forall k \geq 0$, with $r > 0$ being sufficiently large. The convergence of the un-constrained case for $\lambda$ close to $1$ is proved in \cite{yu2015convergence}.\vspace*{2mm}\\
$\bullet$ \textbf{Off-policy LSTD($\lambda$):} Off-policy least squares temporal difference (LSTD) with eligibility traces \cite{yu2012least} is another relevant algorithm in this category. The procedure is described below:
\begin{subequations}\label{eqn:lstd}
\begin{align}
&\mathbf{e}_{k+1} := \gamma\lambda \rho_{k} \mathbf{e}_k + \phi(s_k), \hspace*{5mm}\\
&\mathbf{A}_{k+1} := \mathbf{A}_{k} + \frac{1}{k+1}\left(\mathbf{e}_{k}(\phi(s_{k})-\gamma\rho_{k}\phi(s_{k+1}))^{\top} - \mathbf{A}_k\right),\\
&\mathbf{b}_{k+1} := \mathbf{b}_{k} + \frac{1}{k+1}(\rho_{k}r_{k}\mathbf{e}_k -   \mathbf{b}_k), \hspace*{3mm}\\
&\mathbf{x}_{k+1} := \mathbf{A}^{-1}_{k+1}\mathbf{b}_{k+1},\hspace*{3mm}
\end{align}
\end{subequations}
where $\mathbf{A}_{k} \in \bbbr^{k_{1} \times k_{1}}$ and $\mathbf{e}_{k}$, $\mathbf{b}_{k}$, $\mathbf{x}_{k} \in \bbbr^{k_{1}}$. In some cases, the matrix $\mathbf{A}_{k}$ may not be of full rank. To avoid such singularities, initialize $\mathbf{A}_0$ with $\delta \bbbone_{k_1 \times k_1}$, $\delta > 0$.

Contrary to the earlier algorithm, the off-policy LSTD($\lambda$) is shown to be stable with well defined limiting behaviour for all $\lambda \in [0,1]$ under pragmatic assumptions. The only restriction imposed is that the target policy $\pi_{w}$ is \emph{absolutely continuous} ($\prec$) \emph{w.r.t.} the behaviour policy $\pi_b$, \emph{i.e.}, 
\begin{equation}
\pi_{w} \prec \pi_{w_b} \hspace*{5mm}\Leftrightarrow \hspace*{5mm}\pi_{w_b}(a \vert s) = 0 \Rightarrow \pi_{w}(a \vert s) = 0, \hspace*{2mm}\forall a \in \mathbb{A}, \forall s \in \mathbb{S}.
\end{equation}
The contrapositive form of the above statement implies that $\pi_{w}(a \vert s) \neq 0 \Rightarrow \pi_{w_b}(a \vert s) \neq 0$, $\forall a \in \mathbb{A}, \forall s \in \mathbb{S}$. This means that for a given state $s \in \mathbb{S}$, every action feasible under the target policy $\pi_{w}$ is also feasible under the behaviour policy $\pi_{w_b}$. The following result from \cite{yu2012least} characterizes the limiting behaviour of the off-policy LSTD($\lambda$) algorithm:
\begin{theorem}\label{thm:offplcy}
For a given target policy vector $w \in \mathbb{W}$ and a behaviour policy vector $w_b \in \mathbb{W}$, the sequence $\{\mathbf{x}_{k}\}$ generated by the off-policy LSTD($\lambda$) algorithm defined in Equation (\ref{eqn:lstd}) converges to the limit $x_{w \vert w_b}$ with probability one, where
\begin{equation}\label{eq:offls}
\begin{aligned}
x_{w | w_b} = A^{-1}_{w \vert w_b} & b_{w \vert w_b}, \hspace*{3mm} \textrm{with} \hspace*{45mm}\\ 
A_{w \vert w_b} &= \Phi^{\top}D^{\nu_{w_b}}(I-\gamma \lambda P_{w})^{-1}(I-\gamma P_{w})\Phi \hspace*{5mm}\textrm{ and }\\ b_{w \vert w_b} &= \Phi^{\top}D^{\nu_{w_b}}(I-\gamma\lambda P_{w})^{-1}R^{w}.
\end{aligned}
\end{equation}
Here $D^{\nu_{w_b}}$ is the diagonal matrix with $D^{\nu_{w_b}}_{ii}=\nu_{w_b}(i)$, $1 \leq i \leq \vert \mathbb{S} \vert$, where $\nu_{w_b}$ is the stationary distribution of the Markov chain $P_{w_b}$ induced by the behaviour policy $\pi_{w_b}$, \emph{i.e.}, $\nu_{w_b}$ satisfies $\nu_{w_b}^{\top} P_{w_b} = \nu_{w_b}^{\top}$ and $R^{w}(s) \in \bbbr^{\vert \mathbb{S} \vert}$ is the expected reward, \emph{i.e.}, $R^{w} \triangleq \Sigma_{s^{\prime} \in \mathbb{S}, a \in \mathbb{A}}\pi_{w}(a \vert s)P(s, a, s^{\prime})R(s, a, s^{\prime})$.
\end{theorem}
It is also important to note that in the on-policy LSTD($\lambda$), where both $\pi_w$ and $\pi_{w_b}$ are the same, the limit point $x_{w | w}$ is given by $x_{w \vert w} = A_{w \vert w}^{-1}b_{w \vert w}$, where
\begin{equation}\label{eq:onls}
\begin{aligned}
A_{w \vert w} &= \Phi^{\top}D^{\nu_{w}}(I-\gamma \lambda P_{w})^{-1}(I-\gamma P_{w})\Phi \hspace*{2mm} \textrm{ and }\\  b_{w \vert w} &= \Phi^{\top}D^{\nu_{w}}(I-\gamma\lambda P_{w})^{-1}R^{w}.
\end{aligned}
\end{equation}

\subsection{The Control Problem of Interest}\label{sec:sec1}
In this section, we define a variant of the control problem which is the topic of interest in this paper.\\\\
\textbf{Problem Statement: } 
\begin{equation}\label{eq:modctrl}
\textrm{ Find } w^{*} \in \argmax_{w \in \mathbb{W} \subset \bbbr^{k_{2}}} \mathbb{E}_{\nu_{w}}\left[L(h_{w \vert w})\right],
\end{equation}\\
where  $L:\bbbr^{\vert \mathbb{S} \vert} \rightarrow \bbbr^{\vert \mathbb{S} \vert}$ is a performance function.  We assume that $L$ is bounded and continuous. Note that since $h_{w \vert w} \in \bbbr^{\vert \mathbb{S} \vert}$, we have $L(h_{w \vert w}) \in \bbbr^{\vert \mathbb{S} \vert}$, \emph{i.e.}, $L(h_{w \vert w})$ can be viewed as a mapping from the the state space $\mathbb{S}$ to the scalars. In the case of finite MDP (both $\mathbb{S}$ and $\mathbb{A}$ are finite),  we have $\mathbb{E}_{\nu_{w}}\left[L(h_{w \vert w})\right] = \sum_{s \in \mathbb{S}}\nu_{w}(s)L(h_{w \vert w})(s)$. Thus the objective function in Equation (\ref{eq:modctrl}) is scalar-valued and hence the optimization problem defined in Equation (\ref{eq:modctrl}) is indeed well-defined.\\

\noindent
$\circledast$ \textbf{Assumption (A2)} \textit{The Markov chain under any SRP $\pi_{w}, w \in \bbbr^{k_2}$ is ergodic, \emph{i.e.}, irreducible and aperiodic.}\vspace*{4mm}\\
$\circledast$ \textbf{Assumption (A3)} \textit{$\mathbb{W}$ is a compact subset of $\bbbr^{k_{2}}$.}
\subsection{Motivation}
\begin{figure}
	\centering
	\includegraphics[width=75mm, height=50mm]{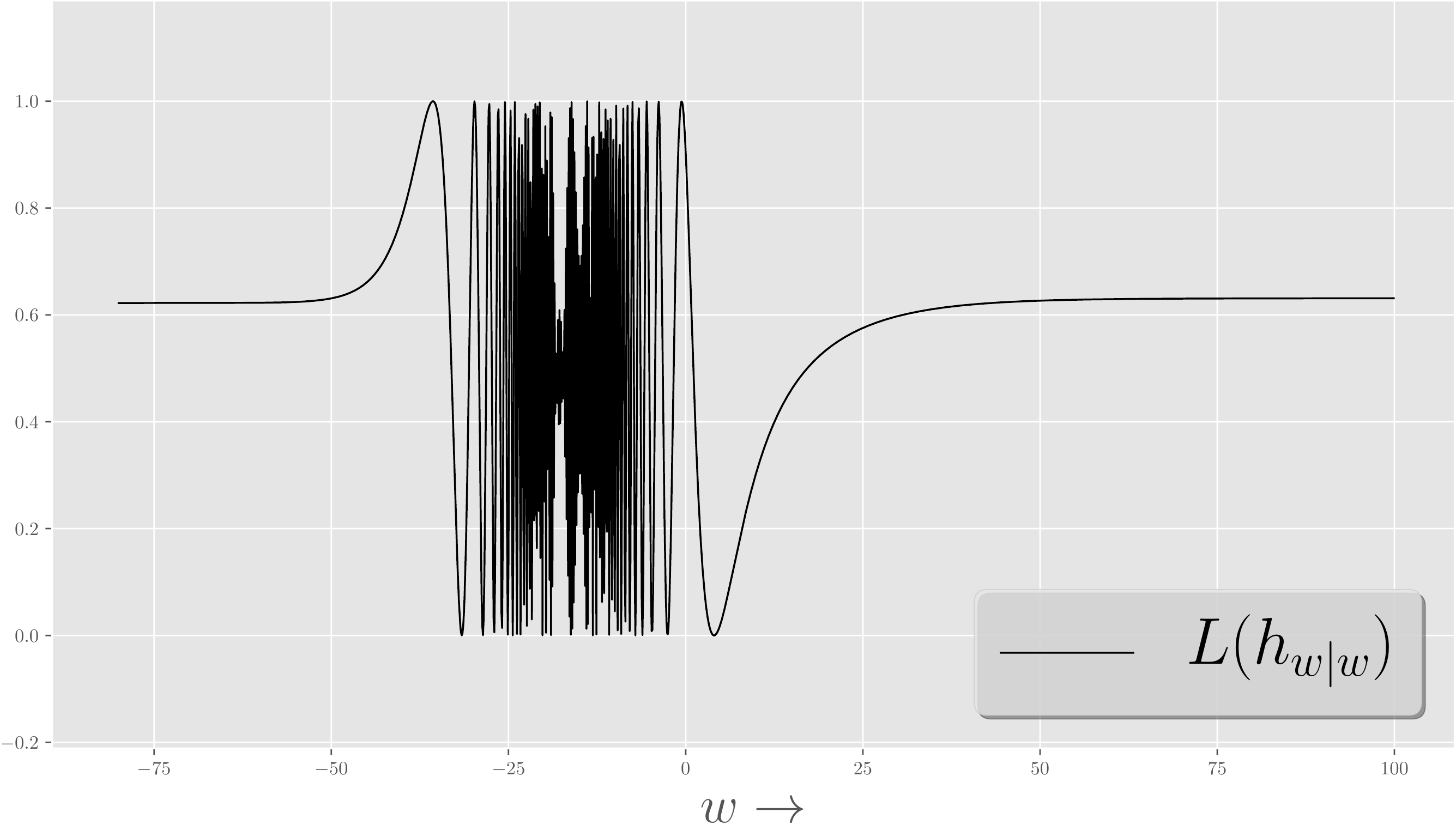}
	\caption{$\mathbb{S} = \{0,1,2,F\}$, $\mathbb{A} = \{0,1\}$, $k_1=1$, $k_2=1$, $\gamma = 0.99, \tau = 10$, $\lambda = 0.00125$, $\psi(s, a)=s*a, \Phi=(1, 0, 1, 0)^{\top}$, $L(h_{w \vert w})(s) = \sin^{2}{(\frac{\pi}{2} s)}$, $P(0,0,0)=P(1,0,1)=P(2,0,2)=1.0$, $P(0,1,1)=P(1,1,2)=P(2,1,F)=1.0$. The remaining transition probabilities are zero. }\label{fig:exctrl}
\end{figure}

We demonstrate here a practical situation where the optimization problem of the kind (\ref{eq:modctrl}) arises. We consider here a special case of the self-drive system. The goal is to propel an automotive (equipped with sensors to detect the vehicular traffic) from source $0$ to destination $F$ (where there are multiple intersections in between) in minimum time without any accidents. Here, the collection of junctions represents the state space, \emph{i.e.}, $\mathbb{S} = \{0,1,2,3,F\}$. The automotive travels with a constant velocity between subsequent intersections and the choice of the velocities is restricted to the discrete, finite set $\{1,2,3\}$. The velocity is chosen randomly by the automotive from the above set at each intersection. The purpose of the randomness is to capture the uncertainty in the traffic conditions during the subsequent stretch of the trip. 
At each intersection, the automotive senses the vehicular traffic at the intersection and has to make a choice of whether to halt or not. 
\begin{figure}[h]
	\begin{subfigure}[h]{0.45\textwidth}
		\includegraphics[width=50mm, height=30mm]{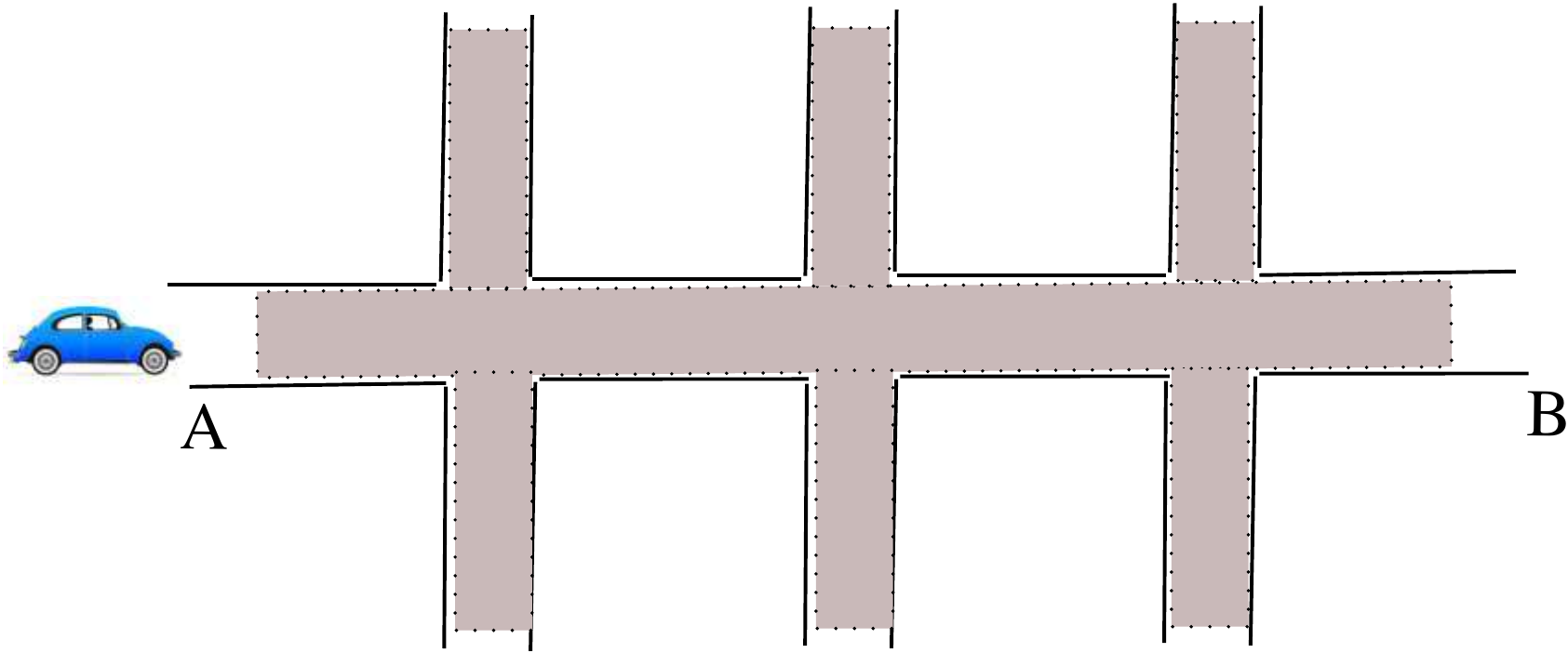}
	\end{subfigure}
	\begin{subfigure}[h]{0.5\textwidth}
		\includegraphics[width=60mm, height=14mm]{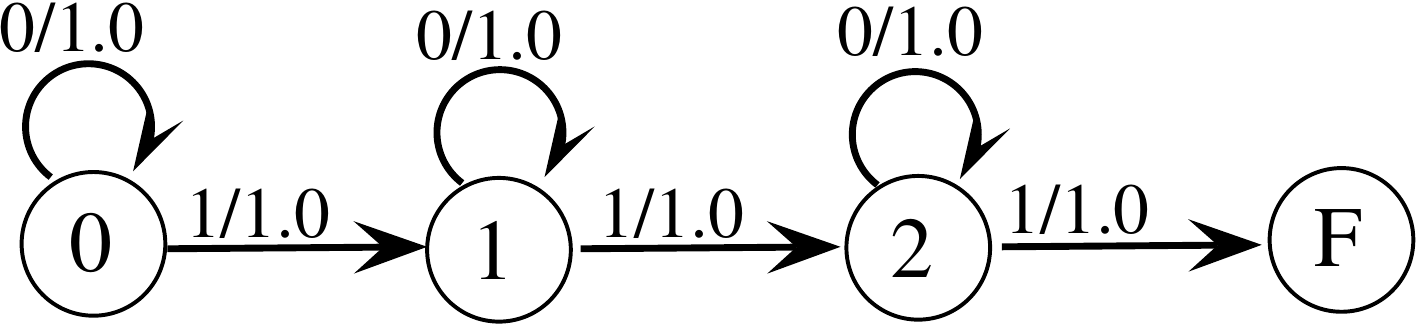}
	\end{subfigure}
	\caption{Self-drive system}	
\end{figure}    
So the action space is $\mathbb{A} = \{0 \textrm{ (halt)}, 1 \textrm{ (proceed)}\}$. Here, the performance of the task is evaluated based on the overall time the automotive takes to cover the distance to the destination. Hence the reward function is taken as the velocity chosen by the automotive to traverse the subsequent stretch. This indeed makes sense since the time is directly dependent on the velocity with distance being constant.  This optimization problem can be modeled using a finite horizon cost function. Now, suppose that the task is further rewarded based on the overall time it takes to complete the trip. In this case, the final payoff is dependent on the value function (in this case, the value function is time), then the role of the performance function $L$ is to capture this particular aspect. If the payoffs are  further based on the maintenance cost incurred (which cannot be integrated into the reward function due to the presence of multiple operating components and hence considering the net maintenance cost at the end of the episode is more worthwhile), the performance function might not be unimodal in general. This is further confirmed by Fig. \ref{fig:exctrl}, where we provide the plot of the objective function of the self-drive MDP which exhibits a complex landscape with many local optima.
This particular problem is more relevant in the context of neural computation, where distinct neural substrates in regions of prefrontal and anterior striatum have been identified with human habitual learning (model free reinforcement learning) \cite{o2015structure,balleine1998goal,lee2014neural}. The human brain is a complex network of computing components and one is inclined to believe that the value function obtained through the habitual learning  will be further evaluated using a performance function (similar to the activation function found in the artificial neural networks) before relaying to the subsequent level in the network.

The control problem in Equation (\ref{eq:modctrl}) is harder due to the application of the  performance function $L$ on the approximate value function. Hence we cannot apply the existing direct model free methods like LSPI or off-policy Q-learning \cite{maei2010toward}. Note that the LSPI algorithm (Fig. $8$ of \cite{lagoudakis2003least}) is a policy iteration method, where at each iteration an improved policy parameter is deduced from the projected Q-value of the previous policy parameter. So one cannot directly incorporate the operator $\mathbb{E}_{\nu_w}$ into the LSPI iteration. Similar compatibility issues are found with the off-policy Q-learning also \cite{maei2010toward}. However, policy search methods are a direct match for this problem. Not all policy search methods can provide quality solutions. The pertinent issue is the non-convexity of $\mathbb{E}_{\nu_w}\left[L(h_{w \vert w})\right]$ which presents a landscape with many local optima. Any gradient based method like the state-of-the-art simultaneous perturbation stochastic approximation (SPSA) \cite{spall1992multivariate} algorithm or the policy gradient methods can only provide sub-optimal solutions.  In this paper, we try to solve the control problem in its \emph{true sense}, \emph{i.e.}, find a solution close to the global optimum of the optimization problem (\ref{eq:modctrl}). We employ a stochastic approximation variant of the well known cross entropy (CE) method proposed in \cite{predictsce2016,genstochce2016,ajincedet} to achieve the \emph{true sense} behaviour. The CE method has in fact been applied to the model free control setting before in \cite{mannor2003cross}, where the algorithm is termed the \textit{fast policy search}. However, the approach in \cite{mannor2003cross} has left several practical and computational challenges uncovered. The method in \cite{mannor2003cross} assumes access to a generative model, \emph{i.e.}, the real MDP system itself or a simulator/computational model of the MDP under consideration, which can be configured with moderate ease (with time constraints) and the observations recorded. The existence of generative models for extremely complex MDPs is highly unlikely, since it demands accurate knowledge about the transition dynamics of the MDP. Now regarding the computational aspect, the algorithm in \cite{mannor2003cross} maintains an evolving $\vert \mathbb{S} \vert \times \vert \mathbb{A} \vert$ matrix $P^{(t)} \triangleq (P^{(t)}_{sa})_{s \in \mathbb{S}, a \in \mathbb{A}}$, where $P^{(t)}_{sa}$ is the probability of taking action $a$ in state $s$ at time $t$. At each discrete time instant $t$, the algorithm generates multiple sample trajectories using $P^{(t)}$, each of finite length, but sufficiently long. For each trajectory, the discounted cost is calculated and then averaged over those multiple trajectories to deduce the subsequent iterate $P^{(t+1)}$. This however is an expensive operation, both computation and storage wise. Another pertinent issue is the number of sample trajectories required at each time instant $t$. There is no analysis pertaining to finding a bound on the trajectory count. This implies a brute-force approach has to be adopted and this further burdens the algorithm. A more recent global optimization algorithm called the model reference adaptive search (MRAS) has also been applied in the model free control setting \cite{chang2013simulation}. However, it also suffers from similar issues as the earlier approach.\\

Here, we illustrate using a real life scenario, the hardness incurred in assuming a generative model. We consider a legacy water delivery system \cite{feinberg2012handbook,fracasso2014optimized,ikonen2011scheduling,ertin2001dynamic}.
\begin{figure}[h]
	\centering	
	\fbox{\includegraphics[width=0.75\textwidth, height=40mm]{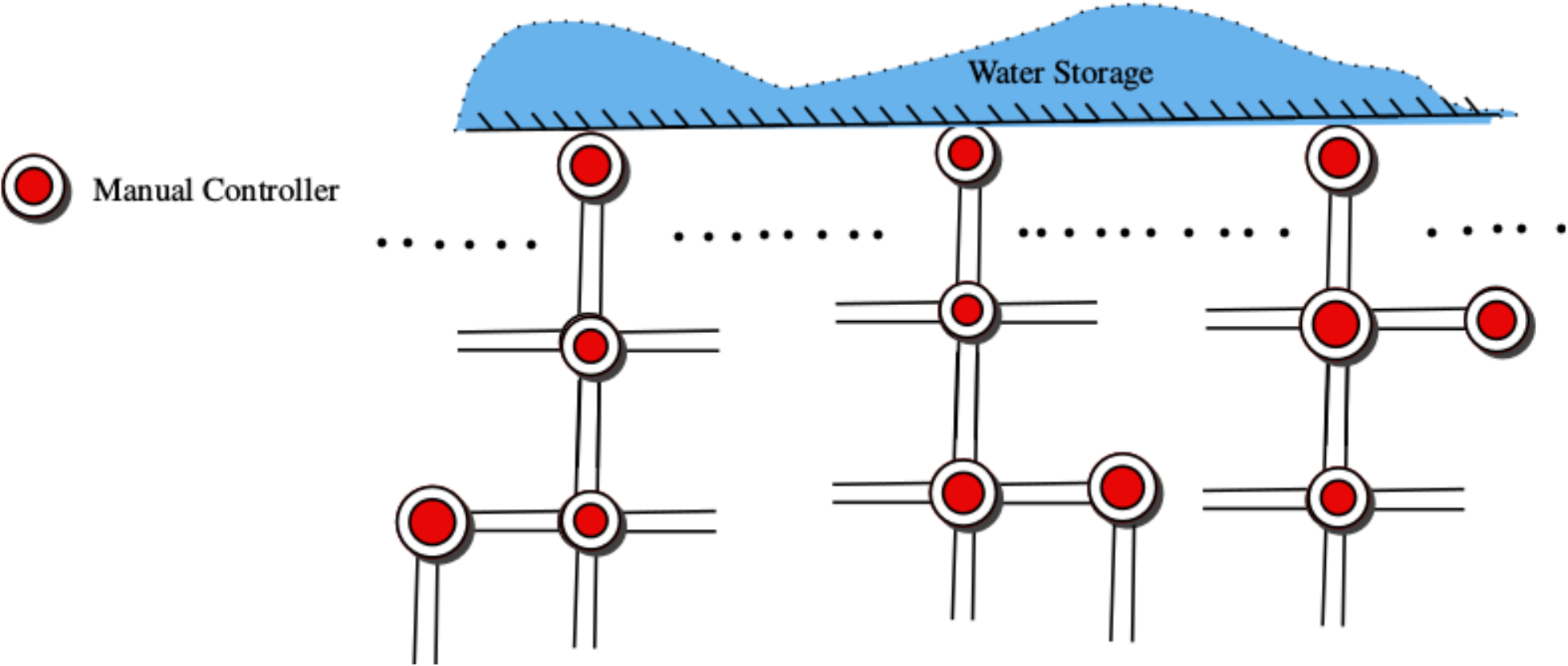}}
	\caption{Water delivery network: The system consists of a water reservoir and a web of manual controllers. The quantity of water in the reservoir is stochastic in nature  and so is the consumption of the water by the end users. The end usage of the system includes agriculture, household activities, power generation etc. The reward function is a complex function with positive weights on profits from  effective utilization and negative weights on spill overs.}\label{fig:waterdelv}
\end{figure}%
The legacy water delivery systems in most cases are not electronically controlled, which implies that a manual intervention is required to adjust the various throughput levels. The reservoir operators have to rely on agreed upon rules, their judgement and experience to calibrate the network. Fig.\ref{fig:waterdelv} shows a water delivery network where there is a web of manual controllers. The state space is the net output (quantity of water delivered) of the delivery system. Intuitively, one might expect the dynamics of the system to be Markovian in character since the immediate future output is indeed dependent on the current quantity of the reservoir and its current consumption rate. So the state variable takes real values and the underlying MDP is continuous.
The reward function is a complex function with positive weights on profits from  effective utilization (agriculture, drinking purpose, power generation, \emph{etc}) and negative weights on spill overs, kinetic energy losses and factors engendering physical damage to the network like excessive pipe pressure.  The objective is to find a configuration for the network of controllers (which is indeed a vector with each co-ordinate deciding the amount of calibration required for the corresponding controller) which provides optimum expected discounted reward. Here the configurations represent the action space and thus are also continuous. The reconfiguration of the whole system as and when  demanded by the algorithm requires heavy human labor, which is a luxury one cannot afford. On the other hand, developing a simulator for this system requires understanding all the sources of water for the reservoir which depends on a wide variety of environmental factors and also the consumption statistics of the end users, both of which require observations for a long period of time notwithstanding the human labor incurred. Therefore, it is hard in general to develop a simulator/generative model for MDPs with large state  and action spaces with complex, opaque and perplexing transition dynamics. Examples where similar issues arise can be found in manual human control, social sciences, biological systems, unmanned aerial vehicles \cite{bagnell2001autonomous} and mechanical systems which wear out quickly like low-cost robots \cite{deisenroth2011pilco}.\\

Other relevant work in the literature includes Bellman-residual minimization based fitted policy iteration using a single trajectory \cite{antos2008learning} and value-iteration based fitted policy iteration using a single trajectory \cite{antos2007value}. However, those approaches fall prey to the curse of dimensionality arising from large action spaces.  Also, they are abstract in the sense that a generic function space is considered and the value function approximation step is expressed as a formal optimization problem. In the above methods which are almost similar in their approach, considerable effort is dedicated to addressing the approximation power of the function space and sample complexity.

In this paper, we address the above mentioned practical and computational concerns. We focus on two key objectives: 
\begin{enumerate}
	\item 
	To reduce the total number of policy evaluations.
	\item
	To find a high performing policy without presuming an unlimited access to the generative model.
\end{enumerate}
By accomplishing the above mentioned objectives, we try to chisel down the requirements inherent in most of the reinforcement learning algorithms and thus enable them to operate in real-time scenarios. We provide here a brief narrative of the approach we follow to realize the above objectives.

To accomplish the former objective, the ubiquitous choice is to employ the stochastic approximation (SA) version of the CE method instead of the naive CE method used in \cite{mannor2003cross}. The SA version of CE is a zero-order optimization method which is incremental, adaptive, robust and stable with the additional attractive attribute of convergence to the global optimum of the objective function. It has been demonstrated empirically in \cite{predictsce2016,ajincedet} that the method exhibits efficient utilization of the samples and possesses better rate of convergence than the naive CE method. The effective sample utilization implies that the method requires minimum number of objective function evaluations. These attributes are appealing in the context of the control problem we consider here, especially in effectively addressing the former objective.  The adaptive nature of the algorithm apparently eliminates any brute-force approach which has a detrimental impact on the performance of the naive CE method.

\begin{figure}[h]
	\centering
	\includegraphics[scale=0.55]{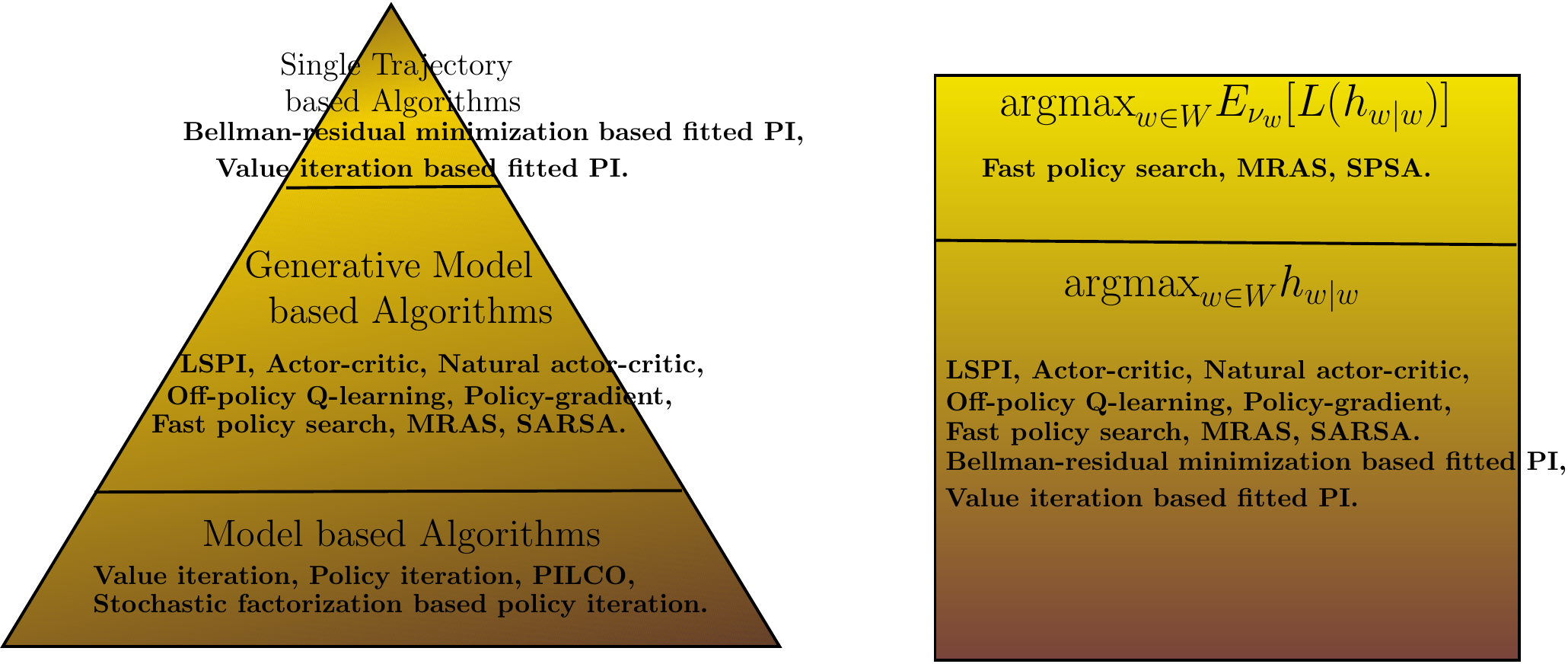}
	\caption{(a) Information pyramid\hspace*{25mm}(b) Optimization box}\label{fig:contentpyramid}
\end{figure}

The latter objective is achieved by employing the off-policy LSTD($\lambda$) for policy evaluation which is defined in Section 2.3. The advantage of this method lies in its ability to approximate the value function of an arbitrary policy (called the target policy) using the observations of the MDP under a possibly different policy (called the behaviour policy), with the only restriction being the absolute continuity between the target and behaviour policies. This implies that we optimize the approximate objective function given by 
$\mathbb{E}_{\nu_{w_b}}\left[L(\Phi x_{w \vert w_b})\right]$ (where $x_{w \vert w_b}$ is the solution generated by the off-policy LSTD($\lambda$)) instead of the true objective function $\mathbb{E}_{\nu_w}\left[L(h_{w \vert w})\right]$. Here, $\nu_{w_b}$ is
the steady state distribution of the Markov chain induced by the behaviour policy $\pi_{w_b}$. This is the best approximation possible under the absence of the generative model since $\nu_{w}$ is the long-run steady state marginal distribution of the Markov chain induced by the policy $\pi_{w}$ and one cannot correct the long-run discrepancies arising due to the restriction that the available sample trajectory is generated using the behaviour policy. However, hidden deep under the appealing characteristic of the single sample trajectory approach is the painful Achilles’ heel of choice, where one cannot forget that the quality of the solution contrived by the algorithm depends on the choice of the sample trajectory which is directly dependent on the behaviour policy that generates it. The additional approximation error incurred due to this particular information restrictive setting is indeed unavoidable. In order to choose the behaviour policy wisely, it is imperative to provide a quantitative analysis of the cost incurred in the choice of the behaviour policy. In this paper, we provide a bound on the approximation error of the off-policy LSTD($\lambda$) solution of an arbitrary target policy with respect to the deviation of the target policy from the behaviour policy. The practical aspect of the approach can be further improved by reconsidering the same sample trajectory for all value function evaluations. This implies that our algorithm just requires a single sample trajectory to solve the optimization problem defined in Equation (\ref{eq:modctrl}). Since the access to the generative model is forbidden, in order to reuse the trajectory, one has to find provisions in terms of memory to store the transition stream.\vspace*{3mm}\\
\textbf{Goal of the Paper: }
\textit{To solve the control problem defined in Equation (\ref{eq:modctrl}) without having access to any generative model. Formally stated, given an infinitely long sample trajectory $\{\mathbf{s}_0, \mathbf{a}_0, \mathbf{r}_0, \mathbf{s}_1, \mathbf{a}_1, \mathbf{r}_1, \mathbf{s}_2, \dots \}$ generated using the behaviour policy $\pi_{w_b}$ ($w_b \in \bbbr^{k_{2}}$), solve the control problem in (\ref{eq:modctrl})}.
\vspace*{2mm}\\
$\circledast$ \textbf{Assumption (A4)} \textit{The behaviour policy $\pi_{w_b}$, where $w_b \in \mathbb{W}$, satisfies the following condition: $\pi_{w_b}(a \vert s) > 0$, $\forall s \in \mathbb{S}, \forall a \in \mathbb{A}$.}\vspace*{2mm}\\
A few remarks are in order: We can classify the reinforcement learning algorithms based on the information made available to the algorithm in order to seek the optimal policy. We graphically illustrate this classification as a pyramid in Fig. \ref{fig:contentpyramid}. The bottom of the pyramid contains the classical methods, where the entire model information, \emph{i.e.}, both $P$ and $R$ are available, while in the middle, we have the  model free algorithms, where both $P$ and $R$ are assumed hidden, however an access to the generative model/simulator is presumed. In the top of the pyramid, we have the single trajectory approaches, where a single sample trajectory generated using a behaviour policy is made available,  however, the algorithms have no access to the model information or simulator. Observe that as one goes up the pyramid, the mass of the information vested upon the algorithm reduces considerably. The algorithm we propose in this paper belongs to the top of the information pyramid and to the upper half of the optimization box which makes it a unique combination.
\section{Proposed Algorithm}
In this section, we propose an algorithm to solve the control problem defined in Equation (\ref{eq:modctrl}). We employ a stochastic approximation variant of the Gaussian based cross entropy method to find the optimal policy. We delay the discussion of the algorithm until the next subsection. We now focus on the objective function estimation. The objective function values $\mathbb{E}_{\nu_{w}}\left[L(h_{w \vert w})\right]$ which are required to efficiently guide the search for $w^{*}$ are estimated using the off-policy LSTD($\lambda$) method. In LFA, given $w \in \mathbb{W}$, the best approximation of $V^{w}$ one can hope for is the projection $\Pi^{w}V^{w}$. Theorem $1$ of \cite{tsitsiklis1997analysis} shows that the on-policy LSTD($\lambda$) solution $\Phi x_{w \vert w}$ is indeed an approximation of the projection $\Pi^{w}V^{w}$. Using Babylonian-Pythagorean theorem and Theorem $1$ of \cite{tsitsiklis1997analysis} along with a little arithmetic, we obtain $\Vert \Phi x_{w \vert w} - \Pi^{w}V^{w} \Vert_{\nu_{w}} \leq \frac{\sqrt{(1-\lambda)\gamma(\gamma+\gamma\lambda+2)}}{1-\gamma}\Vert\Pi^{w}V^{w} - V^{w}\Vert_{\nu_{w}}$. Hence for $\lambda = 1$, we have $\Phi x_{w \vert w} = \Pi^{w}V^{w}$, \emph{i.e.}, the on-policy LSTD(1) provides the exact projection. However for $\lambda < 1$, only approximations to it are obtained. Now when off-policy LSTD($\lambda$) is applied, it adds one more level of approximation, \emph{i.e.}, $\Phi x_{w \vert w}$ is approximated by $\Phi x_{w \vert w_b}$. Hence to evaluate the performance of the off-policy approximation, we must quantify the errors incurred in the approximation procedure and we believe a  capacious analysis had been far overdue.
\subsection{Choice of the behaviour policy}
The behaviour policy is often an exploration policy which promotes the exploration of the state and action spaces of the MDP. Efficient exploration is a precondition for effective learning. In this paper, we operate in a minimalistic MDP setting, where the only information available for inference is the single stream of transitions and payoffs generated using the behaviour policy. So the choice of the behaviour policy is vital for a sound inductive reasoning.  The following theorem will provide a bound on the approximation error incurred in the off-policy LSTD($\lambda$) method. The provided bound can be beneficial in choosing a good behaviour policy and also supplements in understanding the stability and usefulness of the proposed algorithm.
\begin{theorem}\label{thm:solbd}
For a given $w \in \mathbb{W}$, the target policy vector, and $w_b \in \mathbb{W}$, the behaviour policy vector, let $x_{w \vert w}$ and $x_{w \vert w_b}$ be the solutions of the on-policy and off-policy versions of LSTD($\lambda$), respectively, with $\lambda \in [0,1]$.
\begin{flalign}
\textit{If }\hspace*{1mm}\sup_{s \in \mathbb{S}, a \in \mathbb{A}}&\Big\vert \frac{\pi_{w}(a \vert s)}{\pi_{w_b}(a \vert s)}-1\Big\vert < \epsilon_2, \textrm{ then } \frac{\big\Vert x_{w \vert w} - x_{w \vert w_b} \big\Vert_{\infty}}{\Vert x_{w \vert w}\Vert_{\infty}} \leq \nonumber\\&O\big((\vert \mathbb{S} \vert^{2}\epsilon^{2}_2 + \vert \mathbb{S} \vert \epsilon_2)\frac{(1+\gamma)(1+\gamma\lambda)}{(1-\gamma)(1-\gamma\lambda)}\Vert D^{\nu_{w_b}}\Vert_{\infty}\Vert(D^{\nu_{w_b}})^{-1}\Vert_{\infty}\big).\label{eqn:thmbd1}
\end{flalign}
\begin{flalign}
\textit{Also, }\hspace*{1mm}&\Vert \Phi x_{w \vert w_b} - V^{w} \Vert_{\nu_{w_b}} \leq 
\frac{\gamma-2\gamma\lambda+1}{1-\gamma}\Vert V^{w} - V^{w_b} \Vert_{\nu_{w_b}} + \hspace*{10mm}\nonumber\\&\hspace*{20mm}\frac{\epsilon_2(1-\gamma\lambda)\Vert R \Vert_{\infty}}{(1-\gamma)^{2}} + \frac{1-\gamma\lambda}{1-\gamma}\Vert \Pi^{w_b}V^{w} - V^{w} \Vert_{\nu_{w_b}},\label{eqn:thmbd2}
\end{flalign}
where $V^{w}$ and $V^{w_b}$ are the true value functions corresponding to the SRPs $\pi_w$ and $\pi_{w_b}$ respectively. Also, $\nu_{w_b}$ is the stationary distribution of the Markov chain defined by $P_{w_b}$ and $D^{\nu_{w_b}}$ is the diagonal matrix defined in Theorem \ref{thm:offplcy}.
\end{theorem}
\begin{Proof}
Given $w \in \mathbb{W}$, we have \\
\begin{gather*}
P_{w}(s, s^{\prime}) = \sum_{a \in \mathbb{A}}\pi_{w}(a \vert s)P(s, a, s^{'}), s, s^{'} \in \mathbb{S},\\
P_{w_b}(s, s^{\prime}) = \sum_{a \in \mathbb{A}}\pi_{w_b}(a \vert s)P(s, a, s^{'}), s, s^{'} \in \mathbb{S}.
\end{gather*}
Therefore,
\begin{gather*}
P_{w} = P_{w_b} + F, \hspace*{4mm} \textrm{where} \hspace*{2mm} F = P_{w} - P_{w_b}.
\end{gather*}
Hence, for $s, s^{\prime} \in \mathbb{S}$,
\begin{align}\label{eq:fbddrv}
\vert F(s, s^{\prime}) \vert &= \Big\vert \sum_{a \in \mathbb{A}}\left(\pi_{w}(a \vert s) - \pi_{w_b}(a \vert s)\right)P(s, a, s^{\prime}) \Big\vert,\nonumber\\
&= \Big\vert \sum_{a \in \mathbb{A}}\left(\frac{\pi_{w}(a \vert s)}{\pi_{w_b}(a \vert s)} - 1\right)\pi_{w_b}(a \vert s)P(s, a, s^{\prime}) \Big\vert,\nonumber\\
&\leq \sum_{a \in \mathbb{A}}\epsilon_2\pi_{w_b}(a \vert s)P(s, a, s^{\prime}),\nonumber\\
&= \epsilon_2 P_{w_b}(s, s^{\prime}).
\end{align}
The above bound of the deviation matrix $F$ in terms of $P_{w_b}$ compels us to apply the result from \cite{xue1997note}, which provides a sensitivity analysis of the stationary distribution of a Markov chain \emph{w.r.t.} its probability transition matrix. In particular, by appealing to Theorem $1$ of \cite{xue1997note} along with Equation (\ref{eq:fbddrv}), we obtain the following:
\begin{gather}
\Big\vert \frac{\nu_w(s)-\nu_{w_b}(s)}{\nu_{w_b}(s)}\Big\vert \leq 2(\vert \mathbb{S} \vert - 1) \epsilon_2 + O(\epsilon_2^{2}),\hspace*{4mm} s \in \mathbb{S}. \nonumber\\
\Longrightarrow \Big\vert \frac{\nu_w(s)-\nu_{w_b}(s)}{\nu_{w_b}(s)}\Big\vert \leq O(\vert \mathbb{S} \vert\epsilon_2),\hspace*{4mm} s \in \mathbb{S}.\label{eq:bdstd}
\end{gather}
Let $\epsilon_3 = O(\vert \mathbb{S} \vert \epsilon_2)$. Then from (\ref{eq:bdstd}), we get
\begin{gather}
\vert \nu_w(s)-\nu_{w_b}(s) \vert \leq \epsilon_3\vert \nu_{w_b}(s) \vert \leq \epsilon_3(\vert \nu_w(s)-\nu_{w_b}(s) \vert + \vert \nu_{w}(s) \vert)\nonumber\\
\Longrightarrow \frac{\vert \nu_w(s)-\nu_{w_b}(s)\vert}{\vert \nu_{w}(s) \vert} \leq \frac{\epsilon_3}{1-\epsilon_3} = O(\epsilon_3 + \epsilon^{2}_3) = O(\vert \mathbb{S} \vert\epsilon_2 + \vert \mathbb{S} \vert^{2}\epsilon_2^{2}).\label{eq:bdstd2}
\end{gather}
For the policy $\pi_w$, recall that the on-policy approximation is $\Phi x_{w \vert w}$, where $x_{w \vert w}$ is the unique solution to the linear system $A_{w \vert w}x = b_{w \vert w}$. Analogously, the off-policy approximation is given by $\Phi x_{w \vert w_b}$, where $x_{w \vert w_b}$ is the unique solution to the linear system $A_{w \vert w_b}x = b_{w \vert w_b}$. Now using the bound in (\ref{eq:bdstd2}) and the definitions of $A_{w \vert w}$, $A_{w \vert w_b}$, $b_{w \vert w}$ and $b_{w \vert w_b}$ in (\ref{eq:onls}) and (\ref{eq:offls}), it is easy to verify that
\begin{gather*}
\vert A_{w \vert w_b} - A_{w \vert w}\vert \leq O(\vert \mathbb{S} \vert^{2}\epsilon_2^{2} + \vert \mathbb{S} \vert \epsilon_2)\vert A_{w \vert w} \vert\hspace*{0mm}\textrm{ and }\hspace*{0mm}\\ \hspace*{20mm}\vert b_{w \vert w_b} - b_{w \vert w}\vert \leq O(\vert \mathbb{S} \vert^{2}\epsilon_2^{2} + \vert \mathbb{S} \vert \epsilon_2)\vert b_{w \vert w} \vert.
\end{gather*}
Hence the off-policy linear system $A_{w \vert w_b}x = b_{w \vert w_b}$ can be viewed as a perturbed version of the on-policy system $A_{w \vert w}x = b_{w \vert w}$. Let $\epsilon_4 = O(\vert \mathbb{S} \vert^{2}\epsilon_2^{2} + \vert \mathbb{S} \vert \epsilon_2)$. Now we make use of the norm bound on the solutions of perturbed linear system of equations provided in Theorem $2.2$ of \cite{higham1994survey}. In particular, using the remark following Theorem $2.2$ of \cite{higham1994survey},  we have
\begin{gather}\label{eq:normbd}
\frac{\big\Vert x_{w \vert w} - x_{w \vert w_b} \big\Vert_{\infty}}{\Vert  x_{w \vert w} \Vert_{\infty}} \leq \frac{2\epsilon_4\kappa(A_{w \vert w})}{1-\epsilon_4\kappa(A_{w \vert w})},
\end{gather}
where $\kappa(A_{w \vert w}) = \Vert A_{w \vert w} \Vert_{\infty} \Vert A^{-1}_{w \vert w}\Vert_{\infty}$ (condition number $\kappa(\cdot)$ is defined in Section \ref{sec:summary}). Using the definition of $A_{w \vert w}$ in (\ref{eq:onls}), we obtain $A^{-1}_{w \vert w} = \Phi^{-1}(I-\gamma P_{w})^{-1}(I-\gamma \lambda P_{w})(D^{\nu_{w}})^{-1}\Phi^{-\top}$, where $\Phi^{-1}$ is the left inverse of $\Phi$ and $\Phi^{-\top}$ is the right inverse of $\Phi^{\top}$. Therefore $\Vert A^{-1}_{w \vert w} \Vert_{\infty} \leq \Vert\Phi^{-1}\Vert_{\infty}\Vert(I-\gamma P_{w})^{-1}\Vert_{\infty}\Vert I-\gamma \lambda P_{w} \Vert_{\infty} \Vert(D^{\nu_{w}})^{-1}\Vert_{\infty}\Vert\Phi^{-\top}\Vert_{\infty}$. Now by arguing along the same lines as (\ref{eq:cr1_2}), one can show that $\Vert(I-\gamma P_{w})^{-1}\Vert_{\infty} \leq \frac{1}{1-\gamma}$. Also $\Vert I-\gamma \lambda P_{w} \Vert_{\infty} = 1+\gamma\lambda$. And the feature matrix $\Phi$ is presumed to be constant. A forteriori, $\Vert A^{-1}_{w \vert w} \Vert_{\infty} = O(\frac{1+\gamma\lambda}{1-\gamma}\Vert(D^{\nu_{w}})^{-1}\Vert_{\infty})$. Also from (\ref{eq:bdstd}), we have $\nu_{w}(s) \geq (1-\epsilon_3)\nu_{w_b}(s)$, $s \in \mathbb{S}$. Henceforth, $\Vert A^{-1}_{w \vert w} \Vert_{\infty} = O(\frac{1+\gamma\lambda}{(1-\gamma)(1-\epsilon_3)}\Vert(D^{\nu_{w_b}})^{-1}\Vert_{\infty})$. Similarly, one can show that $\Vert A_{w \vert w} \Vert_{\infty} = O(\frac{(1+\gamma)(1+\epsilon_3)}{(1-\gamma\lambda)}\Vert D^{\nu_{w_b}}\Vert_{\infty})$. Hence \vspace*{2mm}\\\hspace*{10mm}$\kappa(A_{w \vert w}) = O\big(\frac{(1+\epsilon_3)(1+\gamma)(1+\gamma\lambda)}{(1-\gamma)(1-\gamma\lambda)(1-\epsilon_3)}\Vert D^{\nu_{w_b}}\Vert_{\infty}\Vert(D^{\nu_{w_b}})^{-1}\Vert_{\infty}\big)$, \vspace*{2mm}\\\hspace*{24mm}= $O\big(\frac{(1+\epsilon_3)^{2}(1+\gamma)(1+\gamma\lambda)}{(1-\gamma)(1-\gamma\lambda)}\Vert D^{\nu_{w_b}}\Vert_{\infty}\Vert(D^{\nu_{w_b}})^{-1}\Vert_{\infty}\big)$.\vspace*{2mm}\\
Consequently from (\ref{eq:normbd}), we get
\begin{flalign*}
\frac{\big\Vert x_{w \vert w} - x_{w \vert w_b} \big\Vert_{\infty}}{\Vert x_{w \vert w}\Vert_{\infty}} &\leq O(\epsilon_4\kappa(A_{w \vert w}) +\epsilon^{2}_4\kappa^{2}(A_{w \vert w}))\\ = O&\big((\vert \mathbb{S} \vert^{2}\epsilon^{2}_2 + \vert \mathbb{S} \vert \epsilon_2)\frac{(1+\gamma)(1+\gamma\lambda)}{(1-\gamma)(1-\gamma\lambda)}\Vert D^{\nu_{w_b}}\Vert_{\infty}\Vert(D^{\nu_{w_b}})^{-1}\Vert_{\infty}\big).
\end{flalign*}
This completes the proof of (\ref{eqn:thmbd1}).\\

Now to prove (\ref{eqn:thmbd2}), here we define an operator $T_{w \vert w_b}^{(\lambda)}$ (referred to as the TD($\lambda$) operator in \cite{tsitsiklis1997analysis}) as follows: 
\begin{gather}\label{eq:Tlbd}
T_{w \vert w_b}^{(\lambda)} V = (1-\lambda)\sum_{i=0}^{\infty}\lambda^{i}\left(\sum_{j=0}^{i}(\gamma P_{w_b})^{j}R^{w}(s_{j}) + (\gamma P_{w_b})^{i+1}V\right)
\end{gather}
\begin{multline}
\textrm{ with }P_{w_b}(s, s^{\prime}) \triangleq \sum_{a \in \mathbb{A}}\pi_{w_b}(a \vert s)P(s, a, s^{\prime})\hspace*{6cm}\\\textrm{ and }R^{w}(s) \triangleq \sum_{s^{\prime} \in \mathbb{S}}\sum_{a \in \mathbb{A}}\pi_{w}(a \vert s)P(s, a, s^{\prime})R(s, a, s^{\prime}).\hspace*{35mm}\label{eq:rdef}
\end{multline}
Before we proceed any further, a few observations are in order:\vspace*{0mm}
\begin{gather}\label{eq:pronen}
\hspace*{-5mm}\textit{Observation 1: } \textrm{For }V \in \bbbr^{\vert \mathbb{S} \vert} \textrm{ and } w \in \mathbb{W} \textrm{, we have }\Vert \Pi^{w} V \Vert_{\nu_{w}} \leq \Vert V \Vert_{\nu_{w}}.
\end{gather}
\vspace*{-8mm}
\begin{flalign*}
&\textit{Proof: }\textrm{Using }<\Pi^{w} V - V, \Pi^{w} V>_{\nu_{w}} = 0 \textrm{ and by the Babylonian-}\hspace*{20mm} \\ 
&\textrm{Pythagorean theorem, we have }
\Vert V \Vert_{\nu_{w}}^{2} = \Vert \Pi^{w} V - V \Vert_{\nu_{w}}^{2} + \Vert \Pi^{w} V \Vert_{\nu_{w}}^{2},\\
&\Rightarrow \Vert \Pi^{w} V \Vert_{\nu_{w}} \leq \Vert V \Vert_{\nu_{w}}.\textrm{ This proves (\ref{eq:pronen})}.
\end{flalign*}
\textit{Observation 2:} For $w \in \mathbb{W}, s \in \mathbb{S}$,
\begin{flalign}\label{eq:rbnd}
\textrm{if }\sup_{a \in \mathbb{S}}&\Big\vert \frac{\pi_{w}(a \vert s)}{\pi_{w_b}(a \vert s)}-1\Big\vert < \epsilon_2 \textrm{ then } \vert R^{w}(s) - R^{w_b}(s) \vert \leq \epsilon_2\Vert R \Vert_{\infty}.\hspace*{0mm}
\end{flalign}
\begin{flalign}
\textit{Proof: } \textrm{From (\ref{eq:rdef}), }& \textrm{ we have},\nonumber\\
\vert R^{w}(s)  - R^{w_b}&(s) \vert = \big\vert \sum_{s^{\prime} \in \mathbb{S}}\sum_{a \in \mathbb{A}}\big(\pi_{w}(a \vert s) - \pi_{w_b}(a \vert s)\big)P(s, a, s^{\prime})R(s, a, s^{\prime})\big\vert,\nonumber\\
& \leq \sum_{s^{\prime} \in \mathbb{S}}\sum_{a \in \mathbb{A}}\big\vert\pi_{w}(a \vert s) - \pi_{w_b}(a \vert s))\big\vert P(s, a, s^{\prime})R(s, a, s^{\prime}),\nonumber\\
& \leq \sum_{s^{\prime} \in \mathbb{S}}\epsilon_2 P_{w_b}(s, s^{\prime})\Vert R \Vert_{\infty},\label{eq:rbnd2}\\
& \leq \epsilon_2 \Vert R \Vert_{\infty}. \hspace*{4mm}\textrm{ This proves (\ref{eq:rbnd})}. \nonumber
\end{flalign}
\begin{equation*}
\begin{aligned}
\textit{Observation 3: }\textrm{ For }V_1, V_2& \in \bbbr^{\vert \mathbb{S} \vert},\hspace*{24mm}\\
&\Vert T_{w \vert w_b}^{(\lambda)} V_1 - T_{w \vert w_b}^{(\lambda)} V_2 \Vert_{\nu_{w_b}} \leq \frac{\gamma(1-\lambda)}{1-\gamma\lambda}\Vert V_1 - V_2 \Vert_{\nu_{w_b}}.\hspace*{20mm}
\end{aligned}
\end{equation*}
\textit{Proof: } Refer Lemma 4 of \cite{tsitsiklis1997analysis}.\vspace*{1mm}\\
\begin{gather}\label{eq:Topbd}
\hspace*{-40mm}\textit{Observation 4: }\hspace*{3mm}\big\vert T_{w \vert w_b}^{(\lambda)} V(s) -  T_{w_b \vert w_b}^{(\lambda)} V(s) \big\vert  \leq \frac{\epsilon_2\Vert R \Vert_{\infty}}{1-\gamma}.
\end{gather}
\textit{Proof: } From (\ref{eq:Tlbd}) and observation 2, we have
\begin{flalign*}
\big\vert T_{w \vert w_b}^{(\lambda)} V(s&)  -  T_{w_b \vert w_b}^{(\lambda)}  V(s) \big\vert =\\
&\Big\vert (1-\lambda)\sum_{i=0}^{\infty}\lambda^{i}\sum_{j=0}^{i}\gamma^{j}\sum_{s^{'} \in \mathbb{S}}P_{w_b}^{j}(s,s^{'})\Big(R^{w}(s^{'})-R^{w_b}(s^{'})\Big) \Big\vert,\\
&\leq (1-\lambda)\sum_{i=0}^{\infty}\lambda^{i}\sum_{j=0}^{i}\gamma^{j}\sum_{s^{'} \in \mathbb{S}}P_{w_b}^{j}(s, s^{'})\Vert R \Vert_{\infty}\epsilon_2,\\
&= (1-\lambda)\sum_{i=0}^{\infty}\lambda^{i}\sum_{j=0}^{i}\gamma^{j}\epsilon_{2} \Vert R \Vert_{\infty},\\
&\leq \frac{\epsilon_2\Vert R \Vert_{\infty}}{1-\gamma}.\hspace*{4mm}\textrm{ This proves (\ref{eq:Topbd}).}
\end{flalign*}
\textit{Observation 5: }$\Phi x_{w \vert w_b} = \Pi^{w_b}T^{(\lambda)}_{w \vert w_b}\Phi x_{w \vert w_b}$. This is the \textit{off-policy projected  Bellman equation}. Detailed discussion is available in \cite{yu2012least}. For the on-policy case, similar equation exists which is as follows: $\Phi x_{w \vert w} = \Pi^{w}T^{(\lambda)}_{w \vert w}\Phi x_{w \vert w}$. For the proof of the above equation, refer Theorem $1$ of \cite{tsitsiklis1997analysis}. A few other relevant fixed point equations are $T^{(\lambda)}_{w \vert w}V^{w} = V^{w}$ and $T^{(\lambda)}_{w_b \vert w_b}V^{w_b} = V^{w_b}$. The proof of the above equations is provided in Lemma 5 of \cite{tsitsiklis1997analysis}.\vspace*{2mm}\\
This completes the observations. Now we will prove (\ref{eqn:thmbd2}). Using the triangle inequality and the above observations, we have
\begin{flalign*}
\Vert \Phi x_{w \vert w_b} & - V^{w} \Vert_{\nu_{w_b}} \leq  \Vert \Phi x_{w \vert w_b}  - \Pi^{w_b}V^{w_b} \Vert_{\nu_{w_b}} + \Vert \Pi^{w_b}V^{w_b}  - V^{w} \Vert_{\nu_{w_b}},\\
=_{1}\hspace*{2mm} \Vert & \Pi^{w_b}T^{(\lambda)}_{w \vert w_b}\Phi x_{w \vert w_b} - \Pi^{w_b}T^{(\lambda)}_{w_b \vert w_b}V^{w_b} \Vert_{\nu_{w_b}} + \Vert \Pi^{w_b}V^{w_b} - V^{w} \Vert_{\nu_{w_b}},\\
\leq_{2}\hspace*{2mm} \Vert & T^{(\lambda)}_{w \vert w_b}\Phi x_{w \vert w_b} - T^{(\lambda)}_{w_b \vert w_b}V^{w_b} \Vert_{\nu_{w_b}} + \Vert \Pi^{w_b}V^{w_b} - V^{w} \Vert_{\nu_{w_b}},
\end{flalign*}
\begin{flalign*}
\leq_{3}\hspace*{2mm} & \Vert  T^{(\lambda)}_{w \vert w_b}\Phi x_{w \vert w_b} - T^{(\lambda)}_{w \vert w_b}V^{w_b} \Vert_{\nu_{w_b}} + \Vert T^{(\lambda)}_{w \vert w_b}V^{w_b} - T^{(\lambda)}_{w_b \vert w_b}V^{w_b} \Vert_{\nu_{w_b}} + \\ &\hspace*{5cm} \Vert \Pi^{w_b}V^{w_b} - V^{w} \Vert_{\nu_{w_b}},\\
\leq_{4}\hspace*{2mm} & \frac{\gamma(1-\lambda)}{1-\gamma\lambda}\Vert  \Phi x_{w \vert w_b} - V^{w_b} \Vert_{\nu_{w_b}} + \Vert T^{(\lambda)}_{w \vert w_b}V^{w_b} - T^{(\lambda)}_{w_b \vert w_b}V^{w_b} \Vert_{\nu_{w_b}} + \\ & \hspace*{5cm} \Vert \Pi^{w_b}V^{w_b} - V^{w} \Vert_{\nu_{w_b}},\\
\leq_{5}\hspace*{2mm} & \frac{\gamma(1-\lambda)}{1-\gamma\lambda}\Vert \Phi x_{w \vert w_b} - V^{w} \Vert_{\nu_{w_b}} + \frac{\gamma(1-\lambda)}{1-\gamma\lambda}\Vert V^{w} - V^{w_b} \Vert_{\nu_{w_b}} + \\ &\hspace*{2cm} \frac{\epsilon_2\Vert R \Vert_{\infty}}{1-\gamma} + \Vert \Pi^{w_b}V^{w_b} - V^{w} \Vert_{\nu_{w_b}},
\end{flalign*}
Note that $=_{1}$ follows from Observation 5; $\leq_{2}$ follows from Observation 1; $\leq_{3}$ follows from the triangle inequality; $\leq_{4}$ follows from Observation 3; $\leq_{5}$ follows from Observation 4 and the triangle inequality.\vspace*{2mm}\\
This further implies
\begin{flalign*}
\frac{1-\gamma}{1-\gamma\lambda}\Vert & \Phi x_{w \vert w_b}  - V^{w} \Vert_{\nu_{w_b}} \\ \leq & \frac{\gamma(1-\lambda)}{1-\gamma\lambda}\Vert V^{w} - V^{w_b} \Vert_{\nu_{w_b}} + \frac{\epsilon_2\Vert R \Vert_{\infty}}{1-\gamma} + \Vert \Pi^{w_b}V^{w_b} - V^{w} \Vert_{\nu_{w_b}},\\
\leq & \frac{\gamma(1-\lambda)}{1-\gamma\lambda}\Vert V^{w} - V^{w_b} \Vert_{\nu_{w_b}} + \frac{\epsilon_2\Vert R \Vert_{\infty}}{1-\gamma} + \Vert \Pi^{w_b}V^{w_b} - \Pi^{w_b}V^{w} \Vert_{\nu_{w_b}} + \\ &\hspace*{4cm}\Vert \Pi^{w_b}V^{w} - V^{w} \Vert_{\nu_{w_b}},\\
\leq & \frac{\gamma(1-\lambda)}{1-\gamma\lambda}\Vert V^{w} - V^{w_b} \Vert_{\nu_{w_b}} + \frac{\epsilon_2\Vert R \Vert_{\infty}}{1-\gamma} + \Vert V^{w_b} - V^{w} \Vert_{\nu_{w_b}} + \\ & \hspace*{4cm}\Vert \Pi^{w_b}V^{w} - V^{w} \Vert_{\nu_{w_b}}.
\end{flalign*}
Therefore
\begin{flalign*}
\Vert \Phi x_{w \vert w_b} - V^{w} \Vert_{\nu_{w_b}} \leq 
\frac{\gamma-2\gamma\lambda+1}{1-\gamma}\Vert V^{w} - V^{w_b} \Vert_{\nu_{w_b}} + \frac{\epsilon_2(1-\gamma\lambda)\Vert R \Vert_{\infty}}{(1-\gamma)^{2}} + \\ \frac{1-\gamma\lambda}{1-\gamma}\Vert \Pi^{w_b}V^{w} - V^{w} \Vert_{\nu_{w_b}}.
\end{flalign*}
This completes the proof of (\ref{eqn:thmbd2}).
\end{Proof}
\vspace*{2mm}\\The implications of the bounds given in Theorem \ref{thm:solbd} are indeed significant. The quantity $\sup_{s \in \mathbb{S}, a \in \mathbb{A}}\Big\vert \frac{\pi_{w}(a \vert s)}{\pi_{w_b}(a \vert s)}-1\Big\vert$ given in the hypothesis of the theorem
can ostensibly be viewed as a measure of the closeness of the SRPs $\pi_w$ and $\pi_{w_b}$, with the minimum value of $0$ being achieved in the on-policy case. Under the hypothesis that $\sup_{s \in \mathbb{S}, a \in \mathbb{A}}\Big\vert \frac{\pi_{w}(a \vert s)}{\pi_{w_b}(a \vert s)}-1\Big\vert < \epsilon_2$, we obtain in (\ref{eqn:thmbd1}) an upper bound on the relative error of the on-policy and off-policy solutions. The bound is predominantly dominated by the hypothesis bound $\epsilon_2$, the eligibility factor $\lambda$, the discount factor $\gamma$ and $\Vert (D^{\nu_{w_b}})^{-1} \Vert_{\infty}\Vert D^{\nu_{w_b}} \Vert_{\infty}$. Note that $\Vert D^{\nu_{w_b}} \Vert_{\infty} = \max_{s} \nu_{w_b}(s)$ and $\Vert (D^{\nu_{w_b}})^{-1} \Vert_{\infty} = (\min_{s} \nu_{w_b}(s))^{-1}$. If the behaviour policy is chosen in such a way that all the states are equally likely under its stationary distribution, then $\Vert (D^{\nu_{w_b}})^{-1} \Vert_{\infty}\Vert D^{\nu_{w_b}} \Vert_{\infty} \approx 1$. Consequently, the upper bound can be reduced to $O\big((\vert \mathbb{S} \vert^{2}\epsilon^{2}_2 + \vert \mathbb{S} \vert \epsilon_2)\frac{(1+\gamma)(1+\gamma\lambda)}{(1-\gamma)(1-\gamma\lambda)}\big)$. 

Now regarding the latter bound provided in Equation (\ref{eqn:thmbd2}), given $w \in \mathbb{W}$, by using triangle inequality and Equation (\ref{eqn:thmbd2}), we obtain a proper quantification of the distance between the solution of the off-policy LSTD($\lambda$), \emph{i.e.}, $\Phi x_{w \vert w_b}$ and the projection $\Pi^{w_b}V^{w}$ in terms of $\Vert\cdot\Vert_{\nu_{w_b}}$ and $\epsilon_2$.  The above bound can be further improved by obtaining an expedient bound for  $\Vert V^{w} - V^{w_b} \Vert_{\nu_{w_b}}$ as follows:
\begin{corollary}
Let $w \in \mathbb{W}$, $\lambda \in [0,1]$ and $\gamma \in (0,1)$. Let the assumptions of Theorem \ref{thm:solbd} hold. Also, assume that $\epsilon_2$ which is defined in Theorem \ref{thm:solbd} satisfy $\epsilon_2\frac{1+\gamma}{1-\gamma} < 1$. Then $\exists K_1 > 0$, s.t.
\begin{equation*}
\begin{aligned}
&\Vert \Phi x_{w \vert w_b} - V^{w} \Vert_{\nu_{w_b}} \leq 
\frac{K_1(\gamma-2\gamma\lambda+1)(1+\gamma)\epsilon_2}{(1-\gamma)(1-\gamma-\epsilon_2(1+\gamma))} + \frac{\epsilon_2(1-\gamma\lambda)\Vert R \Vert_{\infty}}{(1-\gamma)^{2}} + \\&\hspace*{60mm}\frac{1-\gamma\lambda}{1-\gamma}\Vert \Pi^{w_b}V^{w} - V^{w} \Vert_{\nu_{w_b}},
\end{aligned}
\end{equation*}
\end{corollary}
\begin{Proof}
Given $w \in \mathbb{W}$, the value function $V^{w}$ satisfies the linear system given by the Bellman equation as shown in Equation (\ref{eq:bellman}), \emph{i.e.}, 
\begin{equation}\label{eq:lson}
(I-\gamma P_{w})V^{w} = R^{w}.
\end{equation}
Similarly, for the behaviour policy $w_b$, we have 
\begin{equation}\label{eq:lsoff}
(I-\gamma P_{w_b})V^{w_b} = R^{w_b}.
\end{equation}
Now, note that
\begin{flalign*}
&(I-\gamma P_{w}) = (I-\gamma P_{w_b}) + F, \textrm{ where } F = \gamma(P_{w_b} - P_{w}).\\
&R^{w} = R^{w_b} + b, \textrm{ where } b = R^{w} - R^{w_b}.
\end{flalign*}
By arguing along the same lines as (\ref{eq:rbnd2}), one can show that $\vert b(s) \vert \leq \epsilon_2 \vert  R^{w_b}(s) \vert$, $\forall s \in \mathbb{S}$. Similarly, $\vert F(s, s^{\prime}) \vert \leq \epsilon_2 \gamma \vert P_{w_b}(s, s^{\prime}) \vert \leq \epsilon_2 \vert (I-\gamma P_{w_b})(s, s^{\prime}) \vert$, $\forall s, s^{\prime} \in \mathbb{S}$. (The proof is similar to that of (\ref{eq:fbddrv})).
Hence the on-policy linear system given by (\ref{eq:lson}) can be viewed as a perturbed version of the linear system (\ref{eq:lsoff}) of the behaviour policy. So, using the remark following Theorem $2.2$ of \cite{higham1994survey}, we obtain the following:
\begin{flalign}\label{eq:cr1_1}
\frac{\Vert V^{w} - V^{w_b} \Vert_{\nu_{w_b}}}{\Vert V^{w_b} \Vert_{\nu_{w_b}}} \leq \frac{2\epsilon_2\kappa(I - \gamma P_{w_b})}{1-\epsilon_2\kappa(I - \gamma P_{w_b})}.
\end{flalign}
where $\kappa(I - \gamma P_{w_b}) = \Vert I - \gamma P_{w_b} \Vert_{\infty}\Vert (I - \gamma P_{w_b})^{-1} \Vert_{\infty}$ (condition number $\kappa(\cdot)$ is defined in Section \ref{sec:summary}). It is also easy to verify that $\Vert I - \gamma P_{w_b} \Vert_{\infty} = 1+\gamma$. Now to bound $\Vert (I - \gamma P_{w_b})^{-1} \Vert_{\infty}$, we use the Ahlberg-Nilson-Varah bound from \cite{varga1976diagonal}. In particular, by using Theorem A of \cite{varga1976diagonal}, we have
\begin{flalign}\label{eq:cr1_2}
\Vert (I - \gamma P_{w_b})^{-1} \Vert_{\infty} &\leq \frac{1}{\min_{1 \leq i \leq \vert \mathbb{S} \vert}\big\{\vert(I-\gamma P_{w_b})_{ii}\vert - \sum_{j=1, j \neq i }^{\vert \mathbb{S} \vert}\vert(I-\gamma P_{w_b})_{ij}\vert\big\}},\nonumber\\
& = \frac{1}{1-\gamma},
\end{flalign}
where $(\cdot)_{ij}$ is the $(i,j)$ entry of the matrix.\\
By putting together the above facts, we get $\kappa(I - \gamma P_{w_b}) \leq \frac{1+\gamma}{1-\gamma}$. Consequently from Equation (\ref{eq:cr1_1}) and the assumption that $\epsilon_2\frac{1+\gamma}{1-\gamma} < 1$, we obtain
\begin{flalign*}
\frac{\Vert V^{w} - V^{w_b} \Vert_{\nu_{w_b}}}{\Vert V^{w_b} \Vert_{\nu_{w_b}}} \leq \frac{2\epsilon_2(1+\gamma)}{1-\gamma-\epsilon_2(1+\gamma)}.
\end{flalign*}
Therefore $\Vert V^{w} - V^{w_b} \Vert_{\nu_{w_b}} \leq K_{1}\epsilon_2(1+\gamma)(1-\gamma-\epsilon_2(1+\gamma))^{-1}$, $K_1 > 0$. The corollary now easily follows from the above bound and from (\ref{eqn:thmbd2}) of Theorem \ref{thm:solbd}.
\end{Proof}\\

The note worthy result on the upper bound of the approximation error of the on-policy LSTD($\lambda$) provided in \cite{tsitsiklis1997analysis} can be easily derived from the above result as follows:
\begin{corollary}
For $w \in \mathbb{W}$, $\lambda \in [0,1]$ and $\gamma \in (0,1)$,
\begin{flalign*}
\Vert \Phi x_{w \vert w} - V^{w} \Vert_{\nu_{w}} \leq \frac{1-\gamma\lambda}{1-\gamma}\Vert \Pi^{w}V^{w} - V^{w} \Vert_{\nu_{w}}.
\end{flalign*}
\end{corollary}
\begin{Proof}
In the on-policy case, $w_b = w$. Hence $\epsilon_2 = 0$. The corollary directly follows from direct substitution of these values in (\ref{eqn:thmbd2}).
\end{Proof}
\subsection{Estimation of the Objective Function: } 
The objective function of the control problem defined in Equation (\ref{eq:modctrl}) is 
\begin{equation}\label{eq:trobj}
J(w) = \mathbb{E}_{\nu_{w}}\left[L(h_{w \vert w})\right].
\end{equation}
In this paper, we employ off-policy LSTD($\lambda$) to approximate $h_{w \vert w}$ for a given policy parameter $w \in \mathbb{W}$. A sample trajectory $\{\mathbf{s}_{0}, \mathbf{a}_0, \mathbf{r}_{0}, \mathbf{s}_{1}, \mathbf{a}_1, \mathbf{r}_{1}, \mathbf{s}_{2}, \dots \}$ (fixed for the algorithm) generated using the behaviour policy $\pi_{w_b}$  is provided. 

The procedure to estimate the objective function $J$ is formally defined in Algorithm \ref{algo:predict}. The \textit{Predict} procedure in Algorithm \ref{algo:predict} is almost the same as the off-policy LSTD algorithm. The additional recursion (step $10$)  estimates the objective function defined in Equation (\ref{eq:trobj}) as follows:
\begin{equation}\label{eq:lweqval}
\ell_{k+1}^{w} = \ell_{k}^{w} + \alpha_{k+1}\Big(L(\mathbf{x}_{k}^{\top}\phi(\mathbf{s}_{k+1})) - \ell_{k}^{w}\Big),
\end{equation}
where $\alpha_{k} = 1/k$. The above choice of $\alpha_k$ is merely a recommendation and not a strict requirement. This, however, alleviates the extra burden of deciding $\alpha_k$ during implementation.  

For a given $w \in \mathbb{W}$, $\ell_{k}^{w}$ attempts to find an approximate value of the objective function $J(w)$. The following lemma formally characterizes the limiting behaviour of the iterates $\ell_{k}^{w}$.
\begin{lemma} \label{lmn:jw}
For a given $w \in \mathbb{W}$,
\begin{equation}\label{eq:jwconv}
\ell_{k}^{w} \rightarrow \ell^{w}_{*} = \mathbb{E}_{\nu_{w_b}}\left[L(x_{w \vert w_b}^{\top}\phi(\mathbf{s}))\right] \textrm{ as } k \rightarrow \infty \textrm{ w.p. 1.}
\end{equation}
\end{lemma}
\begin{Proof}
We begin the proof by defining the filtration $\{\mathcal{F}_{k}\}_{k \in \mathbb{N}}$, where the $\sigma$-field $\mathcal{F}_{k} \triangleq \sigma(\{\mathbf{x}_{i}, \ell^{w}_{i}, \mathbf{s}_{i}, \mathbf{a}_{i}, \mathbf{r}_{i}, 0 \leq i \leq k \})$.\\\\
Now recalling the recursion (\ref{eq:lweqval}),
\begin{equation*}
\begin{aligned}
\ell_{k+1}^{w} &:= \ell_{k}^{w} + \alpha_{k+1}\Big(L(\mathbf{x}_{k}^{\top}\phi(\mathbf{s}_{k+1})) - \ell_{k}^{w}\Big)\hspace*{20mm}\\
&:= \ell_{k}^{w} + \alpha_{k+1}\Big(h(\ell_{k}^{w}) + \mathbb{M}_{k+1} + c_k\Big),
\end{aligned}
\end{equation*}
where $\mathbb{M}_{k+1} \triangleq L(x_{w \vert w_b}^{\top}\phi(\mathbf{s}_{k+1})) - \mathbb{E}\left[L(x_{w \vert w_b}^{\top}\phi(\mathbf{s}_{k+1})) \big\vert \mathcal{F}_{k}\right]$, \\$h(z) \triangleq  \mathbb{E}_{\nu_{w_b}}\left[L(x_{w \vert w_b}^{\top}\phi(\mathbf{s}_{k+1}))\right] - z$ and $c_k \triangleq L(\mathbf{x}_{k}^{\top}\phi(\mathbf{s}_{k+1})) - L(x_{w \vert w_b}^{\top}\phi(\mathbf{s}_{k+1})) + $\\ $\mathbb{E}\left[L(x_{w \vert w_b}^{\top}\phi(\mathbf{s}_{k+1})) \big\vert \mathcal{F}_{k}\right] - \mathbb{E}_{\nu_{w_b}}\left[L(x_{w \vert w_b}^{\top}\phi(\mathbf{s}_{k+1}))\right]$.\\\\
We state here few observations:
\begin{enumerate}
\item $\{\mathbb{M}_{k}, k \geq 1\}$ is a martingale difference noise sequence \emph{w.r.t.} $\{\mathcal{F}_{k}\}$, \emph{i.e.}, $\mathbb{M}_{k}$ is $\mathcal{F}_{k}$-measurable, integrable and $\mathbb{E}[\mathbb{M}_{k+1} \vert \mathcal{F}_{k}] = 0$ \emph{a.s.}, $\forall k \geq 0$.
\item $h(\cdot)$ is a Lipschitz continuous function.
\item $\exists K > 0$ \emph{s.t.} $\mathbb{E}[\vert \mathbb{M}_{k+1} \vert^{2} \vert \mathcal{F}_{k}] \leq K(1+\vert \ell_{k} \vert^{2})$ \emph{a.s.}, $\forall k \geq 0$.
\item 
By Theorem \ref{thm:offplcy}, $c_k \rightarrow 0$ as $k \rightarrow \infty$ \emph{w.p. 1}. This directly follows by considering the following facts: (a) by Equation (\ref{thm:offplcy}), the off-policy LSTD($\lambda$) iterates $\{\mathbf{x}_{k}\}$ converges almost surely to the off-policy solution $x_{w \vert w_b}$ (b) by assumption (A2), $P_{w_b}(\mathbf{s}_{k} = s) \rightarrow \nu_{w_b}(s)$ as $k \rightarrow \infty$ and (c) $L(\cdot)$ and $\phi(\cdot)$ are bounded.
\item For a given $w \in \mathbb{W}$, the iterates $\{l_{k}^{w}\}_{k \in \mathbb{N}}$ are stable, \emph{i.e.}, $\sup_{k} \vert \ell_{k}^{w} \vert < \infty$ \emph{a.s.} A brief proof is provided here: For $c > 0$, we define 
\begin{equation}
h_{c}(z) \triangleq \frac{h(cz)}{c} = \frac{\mathbb{E}_{\nu_{w_b}}\left[L(x_{w \vert w_b}^{\top}\phi(\mathbf{s}))\right]}{c} - z. 
\end{equation}
Now consider the following ODE corresponding to the following $\infty$-system: 
\begin{equation}
\dot{z}(t) = h_{\infty}(z(t)) \triangleq \lim_{c \rightarrow \infty}h_{c}(z(t)). 
\end{equation}
Note that $h_{\infty}(z) = -z$. It can be easily verified that the above ODE is globally asymptotically stable to the origin. This further implies the stability of the iterates $\{\ell_{k}^{w}\}$ using Theorem $2$, Chapter $3$ of \cite{borkar2008stochastic}.
\end{enumerate}
Now by appealing to the third extension of Theorem $2$, Section $2.2$, Chapter $2$ of \cite{borkar2008stochastic} and from the above observations, we can henceforth conclude almost surely that the iterates $\{\ell_{k}^{w}\}$ asymptotically track the ODE given by:
\begin{equation}\label{eq:odeJk}
\dot{z}(t) = h(z(t)).
\end{equation}
This further implies that the limit points of the iterates $\{\ell_{k}^{w}\}$ are indeed contained in the limit set of the ODE (\ref{eq:odeJk}) almost surely. However, it is easy to verify that $\mathbb{E}_{\nu_{w_b}}\left[L(x^{\top}_{w \vert w_b}\phi(\mathbf{s}))\right]$ is the unique globally asymptotically stable equilibrium of the ODE (\ref{eq:odeJk}). Hence $\lim_{k \rightarrow \infty}\ell_{k}^{w} = \mathbb{E}_{\nu_{w_b}}\left[L(x^{\top}_{w \vert w_b}\phi(\mathbf{s}))\right]$ \emph{a.s.} This completes the proof of (\ref{eq:jwconv}). 
\end{Proof}\\
\begin{remark}
By the above lemma, for a given $w \in \mathbb{W}$, the quantity $\ell_{k}^{w}$ tracks $\mathbb{E}_{\nu_{w_b}}\left[L(x^{\top}_{w \vert w_b}\phi(\mathbf{s}))\right]$. This is however different from the true objective function value $J(w) = \mathbb{E}_{\nu_{w}}\left[L(h_{w \vert w})\right]$, when $w \neq w_b$. This additional approximation error incurred is the extra cost one has to pay for the dearth in information (in the form of generative model) about the underlying MDP. Nevertheless, from Equations (\ref{eqn:thmbd1}) and (\ref{eq:bdstd}), we know that the relative errors in the solutions $x_{w\vert w}$ and $x_{w \vert w_b}$ as well as in the stationary distributions $\nu_{w}$ and $\nu_{w_b}$ are small. We also know that $\Phi x_{w \vert w} \approx h_{w \vert w}$. Further, if we can restrict the smoothness of the performance function $L$, then we can contain the deviation of $L(y)$  when the input variable $y$ is perturbed slightly. All these factors further affirm the fact that the approximation proposed in (\ref{eq:lweqval}) is  well-conditioned. This is indeed significant, considering the restricted setting we operate in, \emph{i.e.}, non-availability of the generative model.
\end{remark}
\scalebox{1.05}{
\begin{minipage}{0.94001\linewidth}
\begin{algorithm}[H]
\caption{Predict Procedure}\label{algo:predict}
\begin{algorithmic}[1]
\State \textbf{Input parameters: } $w \in \mathbb{W}$, $N \in \mathbb{N}$ \hspace*{1mm}$\blacktriangleright \textit{Input policy vector, Trajectory length}$
\State \textbf{Data: } \textit{A priori chosen sample trajectory $\{s_{0}, a_0, r_{0}, s_{1}, a_1, r_{1}, s_{2}, \dots \}$ generated using the behaviour policy $\pi_{w_b}$}
\Function{Predict($w$, $N$)}{}
\State $k := 0$;\hspace*{10mm}$\blacktriangleright \textit{Iteration count initialized to } 0$
\While{$k < N$} 
\State $\mathbf{e}_{k+1} := \gamma\lambda \rho_{k} \mathbf{e}_k + \phi(s_k)$;\hspace*{5mm}$\blacktriangleright \textit{ The sampling ratio }$ $\rho_k = \frac{\pi_{w}(a_{k} \vert s_{k})}{\pi_{w_b}(a_{k} \vert s_{k})}$
\State $\mathbf{A}_{k+1} := \mathbf{A}_{k} + \frac{1}{k+1}\left(\mathbf{e}_k(\phi(s_{k})-\gamma\rho_{k}\phi(s_{k+1}))^{\top} - \mathbf{A}_k\right)$;
\State $\mathbf{b}_{k+1} := \mathbf{b}_{k} + \frac{1}{k+1}(\rho_{k}r_{k}\mathbf{e}_k -   \mathbf{b}_k)$;
\State $\mathbf{x}_{k+1} := \mathbf{A}^{-1}_{k+1}\mathbf{b}_{k+1}$;\hspace*{10mm}$\blacktriangleright \textit{Prediction vector}$
\State $\ell_{k+1}^{w} := \ell_{k}^{w} + \alpha_{k+1}\Big(L(\mathbf{x}_{k}^{\top}\phi(\mathbf{s}_{k+1})) - \ell_{k}^{w}\Big)$;\hspace*{2mm}$\blacktriangleright \textit{Objective func estimation}$\label{line:jw}
\State $k := k + 1$;
\EndWhile
\State \textbf{return }$\ell_{N}^{w}$;\hspace*{10mm}$\blacktriangleright \textit{Outputs after } N \textit{ iterations}$
\EndFunction
\end{algorithmic}
\end{algorithm}
\end{minipage}}
\subsection{Stochastic Approximation Version of Gaussian Cross Entropy Method and its Application to the Control Problem}
Cross entropy method \cite{rubinstein2013cross,kroese2006cross} solves optimization problems where the objective function does not possess good structural properties, such as possibly discontinuous, non-differentiable, \emph{i.e.}, those of the kind:
\begin{equation}
\textrm{Find } x^{*} \in \argmax_{x \in \mathbb{X} \subset \bbbr^{d}} J(x),
\end{equation}
where $J:\mathbb{X} \rightarrow \bbbr$ is a bounded Borel measurable function.\vspace*{2mm}\\
CE is a \textit{model based search method} \cite{zlochin2004model} used to solve the global optimization problem. CE is a zero-order method (\emph{a.k.a.} gradient-free method) which implies the algorithm does not require gradient or higher-order derivatives of the objective function. This remarkable feature of the algorithm makes it a suitable choice for the ``black-box'' optimization setting,  where neither a closed form expression  nor structural properties of the objective function $J$ are available. CE method has found successful application in diverse domains which include continuous multi-extremal optimization \cite{rubinstein1999cross}, buffer allocation \cite{alon2005application}, queueing models \cite{de2000analysis}, DNA sequence alignment \cite{keith2002rare}, control and navigation \cite{helvik2001using}, reinforcement learning \cite{mannor2003cross,menache2005basis} and several NP-hard problems \cite{rubinstein2002cross,rubinstein1999cross}. We would also like to mention that there are other model based search methods in the literature, a few pertinent ones include the gradient-based adaptive stochastic search for simulation optimization (GASSO) \cite{zhou2014simulation}, estimation of distribution algorithm (EDA) \cite{muhlenbein1996recombination} and model reference adaptive search (MRAS) \cite{hu2007model}. However, in this paper, we do not explore the possibility of employing the above algorithms in a MDP setting.

The Gaussian based cross entropy generates a sequence of Gaussian distributions ${\{\theta_{j} = (\mu_{j}, \Sigma_{j})^{\top} \in \Theta \subset \bbbr^{d(d+1)}\}}_{j \in \mathbb{N}}$ parametrized by its mean vector $\mu_{j} \in \bbbr^{d}$ and the covariance matrix $\Sigma_{j} \in \bbbr^{d \times d}$, with the property that the support of the multivariate Gaussian probability density function given by \[f_{\theta_{j+1}}(x) = (2\pi|\Sigma_{j+1}|)^{-d/2}\exp{(-\frac{1}{2}(x-\mu_{j+1})^{\top}\Sigma_{j+1}^{-1}(x-\mu_{j+1}))}\] satisfies (P1) below.\vspace*{2mm}\\
\textbf{Property (P1) }\normalsize$supp(f_{\theta_{j+1}}) \subseteq \{x \vert J(x) \geq \gamma_{\rho}(J, \theta_{j})\}$,\vspace*{2mm}\\
where $\rho \in (0,1)$ is fixed \emph{a priori}. Note that $\gamma_{\rho}(J, \theta_{j})$ is the $(1-\rho)$-quantile of $J$ \emph{w.r.t.} the distribution $f_{\theta_{j}}$. Hence it is easy to verify that the \textit{threshold sequence } $\{\gamma_{\rho}(J, \theta_{j})\}_{j \in \mathbb{N}}$ is a monotonically non-decreasing sequence. The intuition behind this recursive generation of the model sequence is that by assigning greater weight to the higher values of $J$ at each iteration, the expected behaviour of the model sequence should improve. We make the following assumption on the model parameter space $\Theta$:\vspace*{2mm}\\
$\circledast$ \textbf{Assumption (A5) } \normalsize\textit{The parameter space $\Theta$ is a compact subset of $\bbbr^{d(d+1)}$}.\vspace*{3mm}\\
The invariant in each iteration of the CE method is property (P1). The CE method maintains this invariant by solving at each instant $j+1$, the following optimization problem:
\begin{equation}\label{eq:opt1}
\theta_{j+1} = \argmax_{\theta \in \Theta}\Gamma_{j}(\theta, \gamma_{\rho}(J, \theta_{j})),
\end{equation}
where $\Gamma_{j}(\theta, \gamma) \triangleq \mathbb{E}_{\theta_{j}}\left[\varphi(J(\mathbf{x}))I_{\bm{\{}J(\mathbf{x}) \geq \gamma\bm{\}}}\log{f_\theta(\mathbf{x})}\right]$ and $\varphi:\bbbr \rightarrow \bbbr_{+}$ is a positive, strictly monotonically increasing function. This recursive equation forms the basis of the cross entropy method and is referred to as the \textit{model update procedure}.\\
Note that the solution to Equation (\ref{eq:opt1}) is obtained by equating $\nabla \Gamma_{j}$ to $0$:
\begin{gather} 
\nabla_{\vartheta^{\theta}_1}\Gamma_{j}(\theta, \gamma) = 0 \hspace*{3mm}\bm{\Rightarrow}\hspace*{3mm} \mu = \frac{\mathbb{E}_{\theta_j}\left[\mathbf{g_{1}}\bm{\big{(}}J(\mathbf{x}), \mathbf{x}, \gamma\bm{\big{)}}\right]}{\mathbb{E}_{\theta_{j}}\left[\mathbf{g_{0}}(J(\mathbf{x}), \gamma)\right]} \hspace*{3mm}\triangleq \hspace*{3mm} \Upsilon_{1}(\theta_{j}, \gamma),\label{eq:sigmaideal1}\hspace*{5mm}\\
\nabla_{\vartheta^{\theta}_2}\Gamma_{j}(\theta, \gamma) = 0 \hspace*{3mm}\bm{\Rightarrow}\hspace*{3mm} \Sigma =  \frac{\mathbb{E}_{\theta_{j}}\left[\mathbf{g_{2}}\bm{\big{(}}J(\mathbf{x}), \mathbf{x}, \gamma, \mu\bm{\big{)}}\right]}{\mathbb{E}_{\theta_{j}}\left[\mathbf{g_{0}}\bm{\big{(}}J(\mathbf{x}), \gamma\bm{\big{)}}\right]} \hspace*{3mm} \triangleq \hspace*{3mm} \Upsilon_{2}(\theta_{j}, \gamma),\label{eq:sigmaideal2}
\end{gather} 
\begin{subequations}
\begin{empheq}{align}
&\hspace*{-17mm}\textrm{ where }\hspace*{4mm}\mathbf{g_{0}}\bm{(}y, \gamma\bm{)} \triangleq \varphi(y)I_{\bm{\{}y \geq \gamma\bm{\}}}, \hspace*{40mm}\\ 
&\mathbf{g_{1}}\bm{(}y, x, \gamma\bm{)} \triangleq x\varphi(y)I_{\bm{\{}y \geq \gamma\bm{\}}}, \hspace*{31mm}\\
&\hspace*{0mm}\mathbf{g_{2}}\bm{(}y, x, \gamma, \mu\bm{)} \triangleq \varphi(y)(x-\mu)(x-\mu)^{\top}I_{\bm{\{}y \geq \gamma\bm{\}}}\\ 
&(\vartheta^{\theta}_1, \vartheta^{\theta}_2)^{\top} = (\Sigma^{-1}\mu, -\frac{1}{2}\Sigma^{-1})^{\top}.\hspace*{27mm} 
\end{empheq}
\end{subequations}
The mapping of $(\mu, \Sigma)^{\top} \mapsto (\Sigma^{-1}\mu, \frac{-1}{2}\Sigma^{-1})^{\top}$ is a bijective transformation and it makes the algebra a lot simpler. Also it is not hard to verify that $\Upsilon_1$ and $\Upsilon_2$ are well defined. 

Now from (\ref{eq:sigmaideal1}) and (\ref{eq:sigmaideal2}), we can rewrite the recursion (\ref{eq:opt1}) as 
\begin{equation}
\theta_{j+1} = \big(\Upsilon_{1}\left(\theta_{j}, \gamma_{\rho}(J, \theta_{j})\right), \Upsilon_{2}\left(\theta_{j}, \gamma_{\rho}\left(J, \theta_{j}\right)\right)\big)^{\top}.
 \end{equation}
From the definition of $\Upsilon_1$ and $\Upsilon_2$ , note that $f_{\theta_{j+1}}$ also satisfies property (P1) and hence the invariance is maintained. The above update rule for recursively generating model sequence $\{\theta_{j}\}$ is commonly referred to as the \textit{ideal version of the standard CE method}. However, in this paper, we employ an extended version of the CE method proposed in \cite{predictsce2016,genstochce2016,ajincedet} whose update rule is slightly different. In the extended version, a mixture PDF $\widehat{f}_{\theta_j} = (1-\zeta)f_{\theta_{j}} + \zeta f_{\theta_0}$ (with $\zeta \in (0,1)$ and $\theta_0$ is the initial distribution parameter) is employed to compute $\gamma_{\rho}$, $\Upsilon_1$ and $\Upsilon_2$ instead of the original PDF $f_{\theta_j}$. In this case, the update rule is defined as follows:
\begin{equation}\label{eq:mixcemethod}
\begin{aligned}
\theta_{j+1} = \left(\Upsilon_{1}\left(\widehat{\theta}_{j}, \gamma_{\rho}(J, \widehat{\theta}_{j})\right), \Upsilon_{2}\left(\widehat{\theta}_{j}, \gamma_{\rho}\left(J, \widehat{\theta}_{j}\right)\right)\right)^{\top}.\\
\end{aligned}
\end{equation}
Here $\gamma_{\rho}(J, \widehat{\theta})$ is defined as the $(1-\rho)$-quantile of $J$ \emph{w.r.t.} the mixture distribution $\widehat{f}_{\theta}$. Similarly we define  $\Upsilon_1(\widehat{\theta}, \cdot)$ and $\Upsilon_2(\widehat{\theta}, \cdot)$ respectively. This extended version is shown to exhibit global optimum convergence \cite{predictsce2016,genstochce2016,ajincedet}.

However, there are certain tractability concerns. The quantities $\gamma_{\rho}(J, \widehat{\theta}_{j})$, $\Upsilon_{1}(\widehat{\theta}_{j}, \cdot)$ and $\Upsilon_{2}(\widehat{\theta}_{j}, \cdot)$ involved in the update rule are intractable, \emph{i.e.} computationally hard to compute (and hence  the tag name `\textit{ideal}').  To overcome this, a naive approach usually found in the literature is to employ sample averaging, with sample size increasing to infinity. However, this approach suffers from hefty storage and computational complexity which is primarily attributed to the accumulation and processing of huge number of samples. In \cite{predictsce2016,genstochce2016,ajincedet}, a stochastic approximation variant of the extended cross entropy method has been proposed. The proposed approach is efficient both computationally and storage wise, when compared to the rest of the state-of-the-art CE tracking methods \cite{hu2012stochastic,wang2013parameter,kroese2006cross}. It also integrates the mixture approach (\ref{eq:mixcemethod}) and henceforth exhibits global optimum convergence.

The goal of the stochastic approximation (SA) version of Gaussian CE method is to find a sequence of Gaussian model parameters $\{\theta_j = (\mu_j, \Sigma_j)^{\top}\}$ (where $\mu_j$ is the mean vector and $\Sigma_j$ is the covariance matrix) which tracks the ideal CE method. The algorithm efficiently accomplishes the goal by employing multiple stochastic approximation recursions. The algorithm is shown 
\begin{wrapfigure}{r}{0.4\textwidth}
\vspace*{-5mm}
  \begin{center}
   \fbox{\includegraphics[width=0.35\textwidth, height=30mm]{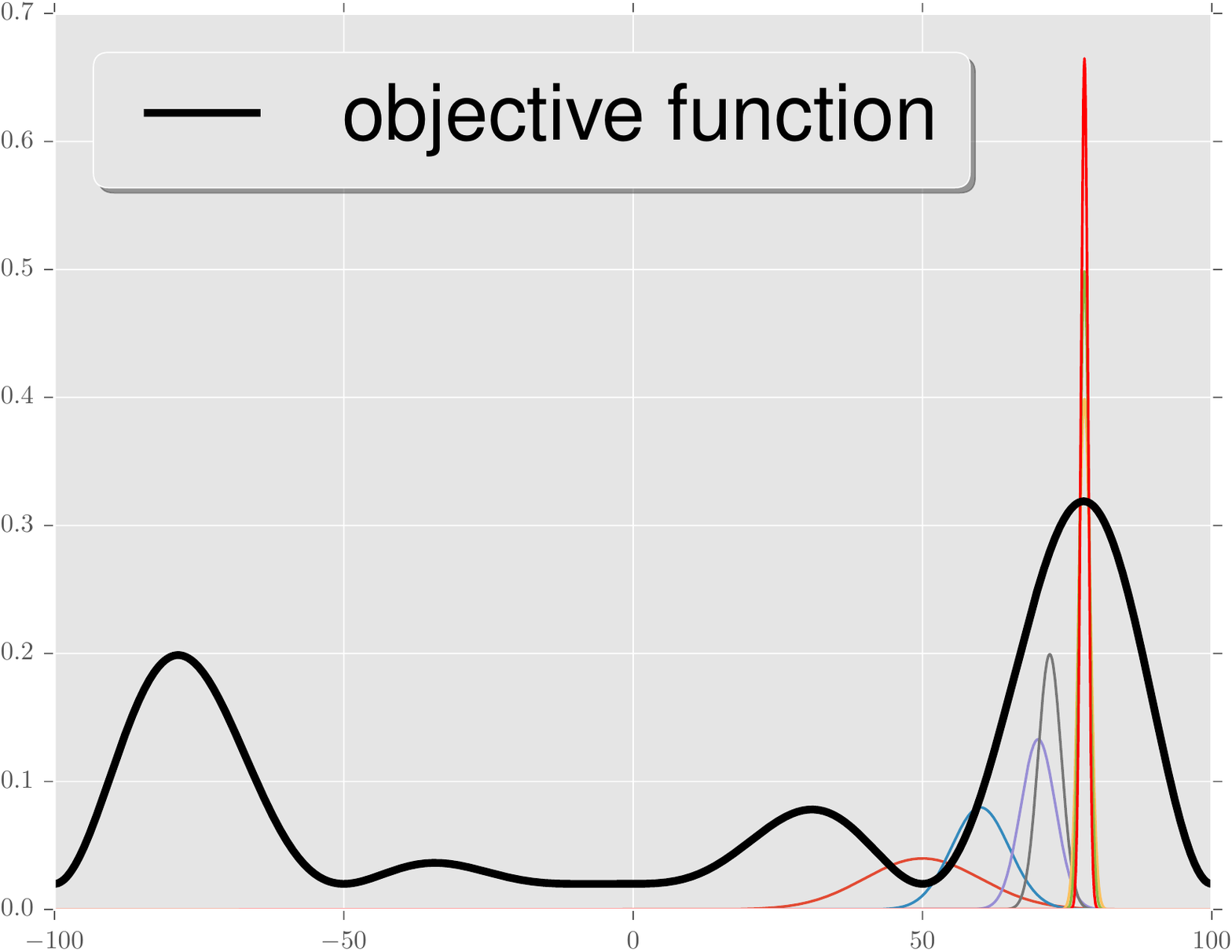}}
  \end{center}
  \vspace*{-2mm}
  \caption{Illustration of the  sequence $\{\theta_j\}$ generated by the CE method}
  \vspace*{-5mm}
\end{wrapfigure}
to exhibit global optimum convergence, \emph{i.e.}, the model sequence $\{\theta_{j}\}$ converges to the degenerate distribution concentrated on any of the global optima of the objective function, in both deterministic (when the objective function is deterministic) and stochastic settings, \emph{i.e.}, when noisy versions of the objective function are available. Successful application of the stochastic approximation version of CE in stochastic settings is appealing to the control problem we consider in this paper, since the off-policy LSTD($\lambda$) method only provides estimates of the value function.
The SA version of CE is a discrete evolutionary procedure where the model sequence $\{\theta_j\}$ is adapted to the degenerate distribution concentrated at global optima, where at each discrete step of the evolution a single sample from the solution space is used. This unique nature of the SA version is appealing to settings where the objective function values are hard to obtain, especially to the MDP control problem we consider in this paper. The single sample requirement attribute which is unique to the SA version implies that one does not need to scale the computing machine for unnecessary value function evaluations.

Our algorithm which attempts to solve the control problem defined in Equation (\ref{eq:modctrl}) is formally illustrated in Algorithm \ref{alg:ce2-nd}.\vspace*{2mm}\\
A few remarks about the algorithm are in order:\vspace*{2mm}\\
1. The learning rates $\{\overbar{\beta}_{j}\}$, $\{\beta_{j}\}$ and the mixing weight $\zeta$ are deterministic, non-increasing and satisfy the following:
\begin{equation}\label{eqn:lrnrate}
\begin{aligned}
\zeta \in (0, 1), \hspace*{2mm}&\beta_{j} > 0, \hspace*{2mm}\overbar{\beta}_{j} > 0, 
\\ &\sum_{j=1}^{\infty}\beta_{j}  = \infty, \hspace*{5mm} \sum_{j=1}^{\infty}\overbar{\beta}_{j}  = \infty, \hspace*{4mm}\sum_{j=1}^{\infty}\left(\beta^{2}_{j}+\bar{\beta}^{2}_{j}\right) < \infty.
\end{aligned}
\end{equation}
2. In our algorithm, the objective function is estimated in (\ref{eqn:ce2lstd}) using the \textit{Predict} procedure which is defined in Algorithm \ref{algo:predict}. Even though an infinitely long sample trajectory is assumed  to be available, the \textit{Predict} procedure has to practically terminate after processing a finite number of transitions from the trajectory. Hence a user configured trajectory length rule $\{N_{j} \in \mathbb{N}\setminus\{0\}\}_{j \in \mathbb{N}}$ with $N_{j} \uparrow \infty$ is used. At each iteration $j$ of the cross entropy method, when \textit{Predict} procedure is invoked to estimate the objective function $L(h_{w_j \vert w_j})$, the procedure terminates after processing the first $N_{j}$ transitions in the trajectory. It is also important to note that the same sample trajectory is reused for all invocations of \textit{Predict}. This eliminates the need for any further observations of the MDP.\vspace*{2mm}\\
3. Recall that we employ the stochastic approximation (SA) version of the extended CE method to solve our control problem (\ref{eq:modctrl}).  The SA version (hence Algorithm \ref{alg:ce2-nd}) maintains three variables: $\gamma_j, \xi^{(0)}_{j}$ and $\xi^{(1)}_{j}$, with $\gamma_j$ tracking $\gamma_{\rho}(\cdot, \theta_j)$, while $\xi^{(0)}_j$ and $\xi^{(1)}_j$ track $\Upsilon_1(\theta_j, \cdot)$ and $\Upsilon_2(\theta_j, \cdot)$ respectively. Their stochastic recursions are defined in Equations (\ref{eq:gammaeq}), (\ref{eq:etaeq}) and (\ref{eq:xieq}) of Algorithm \ref{alg:ce2-nd}. The increment terms for their respective stochastic recursions are defined recursively as follows:
\begin{align}
&\Delta \gamma_{j}(y) \triangleq -(1-\rho)I_{\{y \geq \gamma_j\}}+\rho I_{\{y \leq \gamma_j\}}.\\
&\Delta \xi^{(0)}_{j}(x, y) \triangleq \mathbf{g_{1}}(y, x, \gamma_j) - \xi^{(0)}_j \mathbf{g_{0}}(y, \gamma_j).\\
&\Delta \xi^{(1)}_{j}(x, y) \triangleq \mathbf{g_{2}}(y, x, \gamma_j, \xi^{(0)}_j) - \xi^{(1)}_j \mathbf{g_{0}}(y, \gamma_j).
\end{align}
4. The initial distribution parameter $\theta_0$ is chosen by hand such that probability density function $f_{\theta_0}$ has strictly positive values for every point in the solution space $\mathbb{W}$, \emph{i.e.}, $f_{\theta_0}(w) > 0, \forall w \in \mathbb{W}$.\vspace*{2mm}\\
5. The stopping rule we adopt here for the control problem is to terminate the algorithm when the model sequence $\{\theta_j\}$ is sufficiently close consequently for a finitely long time, \emph{i.e.}, $\exists \bar{j} \geq 0$ \emph{s.t.} $\Vert \theta_j - \theta_{j+1} \Vert < \delta_1$, $\bar{j} \leq \forall j \leq \bar{j}+N(\delta_1)$, where $\delta_1 \in \bbbr_{+}$, $N(\delta_1) \in \mathbb{N}$ are decided \emph{a priori}.\vspace*{2mm}\\
6. The quantile factor $\rho$ is also a relevant parameter of the CE method. An empirical analysis in \cite{predictsce2016} has revealed that the convergence rate of the algorithm is sensitive to the choice of $\rho$. The paper also recommends that $[0.01, 0.3]$ is the most suitable choice of $\rho$.\vspace*{2mm}\\
7. We also extended the algorithm to include Polyak averaging of the model sequence $\{\theta_j\}$. The sequence $\{\overbar{\theta}_{j}\}$ maintains the Polyak averages of the sequence $\{\theta_j\}$ and its update step is given in (\ref{eq:algbarth}). Note that the Polyak averaging \cite{polyak1992acceleration} is a double averaging technique which does not cripple the convergence of the original sequence $\{\theta_j\}$, however it reduces the variance of the iterates and accelerates the convergence of the sequence.\\
\noindent
\scalebox{0.99}{
\hspace*{0mm}\begin{minipage}{1.001\linewidth}
\begin{algorithm}[H]
   \caption{}\label{alg:ce2-nd}
\begin{algorithmic}[1]\vspace*{1mm}
	\State \textbf{Input parameters: } {\setlength{\abovedisplayskip}{-8pt}\setlength{\belowdisplayskip}{4pt}\begin{flalign*}
\hspace*{30mm}\epsilon, \rho \in (0,1), \bar{\beta}_j, \beta_j, \zeta, c_j \in (0,1), c_j \rightarrow 0, \theta_{0} := (\mu_0, \Sigma_0)^{\top},\hspace*{9mm}\vspace*{4mm}\\
	\hspace*{10mm}\{N_j, j \in \mathbb{N}\} \hspace*{4mm}\blacktriangleright \textit{Trajectory length rule chosen a priori}
\end{flalign*}}
	\State \textbf{Initialization:} $j := 0, \gamma_0 := 0$, $\xi^{(0)}_0 := 0_{k_{2} \times 1}$, $\xi^{(1)}_{0} := 0_{k_{2} \times k_{2}}$, $T_{0} := 0$, $\theta^{p} = NULL$, $\gamma^{p}_{0} := -\infty$. \vspace*{3mm}
	\While{stopping criteria not satisfied}
	\vspace*{3mm}
	\State {\setlength{\abovedisplayskip}{-11pt}\begin{flalign}
	\hspace*{-10mm}\textbf{Mixtrure distribution generation: }\widehat{f}_{\theta_j} := (1-\zeta)f_{\theta_{j}} + \zeta f_{\theta_0}; \hspace*{12mm}\end{flalign}}
	\State {\setlength{\abovedisplayskip}{-11pt}\begin{flalign*}
	\hspace*{-12mm}\textbf{Sample generation: }\mathbf{w}_{j+1} \sim \widehat{f}_{\theta_{j}}(\cdot); \hspace*{45mm}
	\end{flalign*}}
	\State {\setlength{\abovedisplayskip}{-11pt}\begin{flalign}
	\textbf{Objective function estimation: }\hat{J}(\mathbf{w}_{j+1}) := Predict(\mathbf{w}_{j+1}, N_{j+1});\hspace*{3mm}\label{eqn:ce2lstd}\hspace*{6mm}
	\end{flalign}}
	\State {\setlength{\abovedisplayskip}{-11pt}\begin{flalign}\label{eq:gammaeq}
	\hspace*{-22mm}\textbf{Tracking $\gamma_{\rho}(J, \theta_j)$: }\hspace*{4mm}\gamma_{j+1} := \gamma_{j} - \beta_{j} \Delta \gamma_{j}(\hat{J}(\mathbf{w}_{j+1}));\hspace*{13mm}
	\end{flalign}}
	\State {\setlength{\abovedisplayskip}{-11pt}\begin{flalign}\label{eq:etaeq}
	\hspace*{-12mm}\textbf{Tracking $\Upsilon_1(\theta_j, \gamma_{\rho}(J, \theta_j))$: }\hspace*{4mm}\xi^{(0)}_{j+1} := \xi^{(0)}_{j}+\beta_{j}\Delta \xi^{(0)}_{j}(\mathbf{w}_{j+1}, \hat{J}(\mathbf{w}_{j+1}));
	\end{flalign}}
	\State {\setlength{\abovedisplayskip}{-11pt}\begin{flalign}\label{eq:xieq}
	\hspace*{-12mm}\textbf{Tracking $\Upsilon_2(\theta_j, \gamma_{\rho}(J, \theta_j))$: }\hspace*{4mm}\xi^{(1)}_{j+1} := \xi^{(1)}_{j} + \beta_{j} \Delta \xi^{(1)}_{j}(\mathbf{w}_{j+1}, \hat{J}(\mathbf{w}_{j+1}));\hspace*{0mm}
	\end{flalign}}
	\If{$\theta^{p} \neq \textit{ NULL }$}\vspace{2mm}
	\hspace*{20mm}\State $\hspace*{2mm}\mathbf{w}^{p}_{j+1} \sim \widehat{f}_{\theta^{p}} := (1-\zeta)f_{\theta^{p}} + \zeta f_{\theta_0};\hspace*{10mm}$\vspace*{2mm}
	\hspace*{10mm}\State $\hspace*{2mm}\gamma^{p}_{j+1} := \gamma^{p}_{j} - \beta_{j} \Delta \gamma^{p}_{j}(\hat{J}(\mathbf{w}^{p}_{j+1}));$\vspace{2mm}
	\EndIf
	\vspace*{4mm}
\State {\setlength{\abovedisplayskip}{-14pt}\begin{flalign}\label{eq:Teq}
\hspace*{-10mm}\textbf{Threshold comparison: }T_{j+1} := T_{j} + c \left(I_{\{\gamma_{j} > \gamma^{p}_{j}\}} - I_{\{\gamma_{j} \leq \gamma^{p}_{j}\}} - T_{j}\right);\hspace*{1mm}
	\end{flalign}}
	\If{$T_{j+1} > \epsilon$}
	\vspace*{2mm}
		\State {\setlength{\abovedisplayskip}{-11pt}\begin{flalign*}
		\textbf{Save old model and old threshold:}\hspace*{4mm} \theta^{p}_{j+1} := \theta_{j}; \hspace*{4mm}\gamma^{p}_{j+1} := \gamma_{j}; 
		\end{flalign*}}
	\State {\setlength{\abovedisplayskip}{-11pt}\begin{flalign}
		\hspace*{8mm}\textbf{Model parameter update:}\hspace*{4mm} \theta_{j+1} := \theta_{j} + \beta_{j}\left((\xi^{(0)}_{j}, \xi^{(1)}_{j})^{\top} - \theta_{j}\right);\label{eq:algth}
		\end{flalign}}
		\State {\setlength{\abovedisplayskip}{-13pt}\begin{flalign}
		\hspace*{5mm}\textbf{Reset parameters:}\hspace*{4mm} 		
		T_{j} := 0;\hspace*{4mm} c := c_{j};\hspace*{32mm}
		\end{flalign}}
		\State {\setlength{\abovedisplayskip}{-11pt}\begin{flalign}
		\hspace*{9mm}\textbf{Weighted Polyak averaging: }\hspace*{4mm}\overbar{\theta}_{j+1} := \overbar{\theta}_{j} + \overbar{\beta}_{j+1}\left(\theta_{j+1} - \overbar{\theta}_{j}\right);\hspace*{4mm} \label{eq:algbarth}
	\end{flalign}}
	\Else
		\vspace*{2mm}
		\State $\hspace*{5mm}\gamma^{p}_{j+1} = \gamma^{p}_{j}; \hspace{4mm} \theta_{j+1} = \theta_{j};\hspace*{10mm}$
		\vspace*{4mm}
	\EndIf
	\vspace*{2mm}
	\State $j := j+1$;\vspace*{3mm}
	\EndWhile
\end{algorithmic}
\end{algorithm}
\end{minipage}}\vspace*{1mm}\\

\subsection{Convergence Analysis of Algorithm \ref{alg:ce2-nd}}
The convergence analysis of the generalized variant of Algorithm \ref{alg:ce2-nd} is already addressed in \cite{predictsce2016} and its application to the prediction problem is given in \cite{predictsce2016}. However, for completeness, we will restate the results here. We do not give proof of those results, however, provide references for the same. The additional Polyak averaging (step 18 of Algorithm \ref{alg:ce2-nd}) requires analysis, which is covered below.

Note that Algorithm \ref{alg:ce2-nd} employs the off-policy prediction method for estimating the objective function. In particular, in step 6 of Algorithm \ref{alg:ce2-nd}, we have $\hat{J}(\mathbf{w}_{j+1}) := Predict(\mathbf{w}_{j+1}, N_{j+1})$, which converges to $\mathbb{E}_{\nu_{w_b}}\left[L(x^{\top}_{w \vert w_b}\phi(\mathbf{s}))\right]$ almost surely as $N_j \rightarrow \infty$ ( by Lemma \ref{lmn:jw}). Hence the objective function optimized by Algorithm \ref{alg:ce2-nd} is $J_b(w) \triangleq \mathbb{E}_{\nu_{w_b}}\left[L(x^{\top}_{w \vert w_b}\phi(\mathbf{s}))\right]$, where $w_b \in \mathbb{W}$ is the chosen behaviour policy vector.

Also note that the model parameter $\theta_{j}$ in Algorithm \ref{alg:ce2-nd} is not updated at each iteration $j$. Rather it is updated whenever $T_{j}$ hits the $\epsilon$ threshold (step 15 of Algorithm \ref{alg:ce2-nd}), where $\epsilon \in (0, 1)$ is a constant. So the update of $\theta_{j}$ only happens along a sub-sequence $\{j_{(n)}\}_{n \in \mathbb{N}}$ of $\{j\}_{j \in \mathbb{N}}$. Between $j = j_{(n)}$ and $j = j_{(n+1)}$, the model parameter $\theta_j$ remains constant and the variable $\gamma_{j}$ estimates $(1-\rho)$-quantile of  $J_b$  \emph{w.r.t.}  $\widehat{f}_{\theta_{j_{(n)}}}$.\\\\
\textbf{Notation: } We denote by $\gamma_{\rho}(J_b, \widehat{\theta}))$, the $(1-\rho)$-quantile of $J_b$ \emph{w.r.t.} the mixture distribution $\widehat{f}_{\theta}$ and let $E_{\widehat{\theta}}[\cdot]$ be the expectation \emph{w.r.t} $\widehat{f}_{\theta}$.\\

Since the model parameter $\theta_j$ remains constant between $j = j_{(n)}$ and $j = j_{(n+1)}$, the convergence behaviour of $\gamma_j$, $\xi^{(0)}_j$ and $\xi^{(1)}_j$ can be studied by keeping $\theta_j$ constant.
\begin{lemma}
	Let $\theta_{j} \equiv \theta, \forall j$. Also assume $sup_{j} \vert \gamma_{j} \vert < \infty$. Then the stochastic sequence $\{\gamma_{j}\}$ defined in Equation (\ref{eq:gammaeq}) satisfies $\lim_{j \rightarrow \infty}\gamma_{j}  = \gamma_{\rho}(J_b, \widehat{\theta}) \hspace*{2mm}\textrm{ w.p. 1 }$.
\end{lemma}
\begin{Proof}
Refer Lemma $3$ of \cite{predictsce2016}.
\end{Proof}

\begin{lemma}\label{lmn:xilimit}
Assume $\theta_{j} \equiv \theta,\forall j$ . Then almost surely,\\
\begin{enumerate}[leftmargin=4mm]
\item
{\setlength{\abovedisplayskip}{-22pt}\setlength{\belowdisplayskip}{20pt}\begin{flalign*}
\hspace*{-42mm}\lim_{j \rightarrow \infty} \xi^{(0)}_{j} = \xi^{(0)}_{*} = \frac{\mathbb{E}_{\widehat{\theta}}\left[\mathbf{g_{1}}\big{(}J_b(\mathbf{x}), \mathbf{x}, \gamma_{\rho}(J_b, \widehat{\theta})\bm{\big{)}}\right]}{\mathbb{E}_{\widehat{\theta}}\left[\mathbf{g_{0}}\big{(}J_b(\mathbf{x}), \gamma_{\rho}(J_b, \widehat{\theta})\big{)}\right]}. 
\end{flalign*}}
\item
{\setlength{\abovedisplayskip}{-23pt}\setlength{\belowdisplayskip}{10pt}\begin{flalign*}
\hspace*{-35mm}\lim_{j \rightarrow \infty} \xi^{(1)}_{j} = \xi^{(1)}_{*} = \frac{\mathbb{E}_{\widehat{\theta}}\left[\mathbf{g_{2}}\big{(}J_b(\mathbf{x}), \mathbf{x}, \gamma_{\rho}(J_b, \widehat{\theta}), \xi^{(0)}_{*}\big{)}\right]}{\mathbb{E}_{\widehat{\theta}}\left[\mathbf{g_{0}}\big{(}J_b(\mathbf{x}), \gamma_{\rho}(J_b, \widehat{\theta})\big{)}\right]}.
\end{flalign*}}
\item
$T_j$ defined in Equation (\ref{eq:Teq}) satisfies $-1 < T_j < 1, \forall j$.\vspace*{2mm}\\
\item
If $\gamma_{\rho}(J_b, \widehat{\theta}) > \gamma_{\rho}(J_b, \widehat{\theta}^{p})$, then $T_{j}$,$j \geq 1$ in (\ref{eq:Teq}) satisfy $\lim_{j \rightarrow \infty} T_{j} = 1$ a.s.
\end{enumerate}
\end{lemma}
\begin{Proof}
For $(i)$, $(ii)$ and $(iv)$, refer Lemma $4$ of \cite{predictsce2016}. For $(iii)$ refer Proposition $1$ of \cite{predictsce2016}.
\end{Proof}\\\\
\textbf{Notation: }For the subsequence $\{j_{(n)}\}_{n > 0}$ of $\{j\}_{j \in \mathbb{N}}$, we denote $j^{-}_{(n)} \triangleq j_{(n)}-1$ for $n > 0$.
\vspace*{4mm}\\
Along the subsequence $\{j_{(n)}\}_{n \geq 0}$ with $j_{0} = 0$ the updating of $\theta_{j}$ can be expressed as follows:  
\begin{equation}\label{eqn:thetareal}
\theta_{j_{(n+1)}} := \theta_{j_{(n)}} + \beta_{j_{(n+1)}}\Delta\theta_{j_{(n+1)}},
\end{equation}  
where  $\Delta\theta_{j_{(n+1)}}$ = $(\xi^{(0)}_{j^{-}_{(n+1)}}, \xi^{(1)}_{j^{-}_{(n+1)}})^{\top} - \theta_{j_{(n)}}$.\\

We now present our main result. The following theorem shows that the model sequence $\{\theta_{j}\}$ and the averaged sequence $\{\overbar{\theta}_{j}\}$ generated by Algorithm \ref{alg:ce2-nd} converge to the degenerate distribution concentrated on the global maximum of the objective function $J_b$.
\begin{theorem}\label{thm:main}
	Let $\varphi(x) = exp(rx), r \in \bbbr$.  Let $\rho, \zeta \in (0,1)$. Let the learning rates $\{\overbar{\beta}_{j}\}$ and $\{\beta_{j}\}$ satisfy Equation (\ref{eqn:lrnrate}). Assume $J_b \in \mathcal{C}^{2}$. Let $\{\theta_{j} = (\mu_{j}, \Sigma_{j})\}_{j \in \mathbb{N}}$ and $\{\overbar{\theta}_{j} = (\overbar{\mu}_{j}, \overbar{\Sigma}_{j})\}_{j \in \mathbb{N}}$ be the sequences generated by Algorithm \ref{alg:ce2-nd} and also assume $\theta_{j} \in \Theta$, $\forall j \in \mathbb{N}$. Let $\overbar{\beta}_{j} = \mathit{o}(\beta_{j})$. Let $w_b \in \mathbb{W}$ be the chosen behaviour policy vector. Also, let the assumptions (A1-A5) hold. Then 
\begin{equation}\label{eq:thconv}
\theta_{j} \rightarrow (w^{b*}, 0_{k_2 \times k_2})^{\top} \hspace*{5mm}\textrm{ as } \hspace*{2mm}j \rightarrow \infty \hspace*{5mm} w.p.1,
\end{equation}
\begin{equation}\label{eq:thbarconv}
\overbar{\theta}_{j} \rightarrow (w^{b*}, 0_{k_2 \times k_2})^{\top} \hspace*{5mm}\textrm{ as } \hspace*{2mm}j \rightarrow \infty \hspace*{5mm} w.p.1,
\end{equation}
$\textrm{where } w^{b*} \in \argmax_{w \in \mathbb{W}} J_b(w)$ with $J_b(w) \triangleq \mathbb{E}_{\nu_{w_b}}\left[L(x^{\top}_{w \vert w_b}\phi(\mathbf{s}))\right]$.
\end{theorem}
\begin{Proof}
Since $\overbar{\beta}_{j} = o(\beta_{j})$, $\overbar{\beta}_{j} \rightarrow 0$ faster than $\beta_{j} \rightarrow 0$. This implies that the updates of $\theta_j$ in (\ref{eq:algth}) are larger than those of $\overbar{\theta}_j$ in (\ref{eq:algbarth}). Hence the sequence $\{\theta_{j}\}$ appears quasi-convergent when viewed from the  timescale of $\{\overbar{\theta}_{j}\}$ sequence.

	Theorem $2$ of \cite{predictsce2016} analyses the limiting behaviour of the stochastic recursion (\ref{eq:algth}) of Algorithm \ref{alg:ce2-nd} in great detail. The analysis discloses the global optimum convergence of the algorithm under limited regularity conditions. It is shown that the model sequence $\{\theta_j\}$ converges almost surely to the degenerate distribution concentrated on the global optimum. The proposed regularity conditions for the global optimum convergence are that the objective function belongs to $\mathcal{C}^{2}$ and the existence of a Lyapunov function on the neighbourhood of the degenerate distribution concentrated on the global optimum. This justifies the hypothesis $J_{b} \in \mathcal{C}^{2}$ in the statement of the theorem and we further assume the existence of a Lyapunov function on the neighbourhood of the degenerate distribution $(w^{b*}, 0_{k_2 \times k_2})^{\top}$. Then by Theorem $2$ of \cite{predictsce2016}, we deduce that $\{\theta_j\}$ converges to $(w^{b*}, 0_{k_2 \times k_2})^{\top}$. This completes the proof of (\ref{eq:thconv}).\\\\
For brevity, lets define $\theta^{*} \triangleq (w^{b*}, 0_{k_2 \times k_2})^{\top}$. We also define the filtration $\{\overbar{\mathcal{F}}_{j}\}_{j \in \mathbb{N}}$, where the $\sigma$-field $\overbar{\mathcal{F}}_{j} \triangleq \sigma(\theta_{i}, \overbar{\theta}_i, 0 \leq i \leq j \})$. Now recalling recursion (\ref{eq:algbarth}),
\begin{align*}
\overbar{\theta}_{j+1} &:= \overbar{\theta}_{j} + \overbar{\beta}_{j+1}\left(\theta_{j+1} - \overbar{\theta}_{j}\right),\\
&:= \overbar{\theta}_{j} + \overbar{\beta}_{j+1}\left(\theta_{j} - \mathbb{E}\left[\theta_{j+1} \vert \overbar{\mathcal{F}}_{j}\right] + \mathbb{E}\left[\theta_{j+1} \vert \overbar{\mathcal{F}}_{j}\right] - \theta^{*} + \theta^{*} - \overbar{\theta}_{j}\right),\\
&:= \overbar{\theta}_{j} + \overbar{\beta}_{j+1}\left(\overbar{\mathbb{M}}_{j+1} + \overbar{b}_{j} + \overbar{h}(\overbar{\theta}_{j})\right),
\end{align*}
where $\overbar{\mathbb{M}}_{j+1} \triangleq \theta_{j+1} - \mathbb{E}\left[\theta_{j+1} \vert \overbar{\mathcal{F}}_{j}\right]$, $\overbar{b}_{j} \triangleq \mathbb{E}\left[\theta_{j+1} \vert \overbar{\mathcal{F}}_{j}\right] - \theta^{*}$ and $\overbar{h}(x) \triangleq \theta^{*} - x$.\\\\
Here we make the following observations:
\begin{enumerate}
\item $\overbar{b}_{j} \rightarrow  0$ almost surely as $j \rightarrow \infty$. This follows from the hypothesis $\overbar{\beta}_{j} = o(\beta_j)$ and by considering the fact that $\theta_j \rightarrow \theta^{*}$ almost surely.
\item $\overbar{h}$ is Lipschitz continuous.
\item $\{\overbar{\mathbb{M}}_{j}\}$ is a martingale difference sequence.
\item $\{\overbar{\theta}_{j}\}$ is stable, \emph{i.e.}, $\sup_{j}\Vert \overbar{\theta}_{j} \Vert < \infty$.
\item The ODE defined by $\dot{\overbar{\theta}}(t) = \overbar{h}(\overbar{\theta}(t))$ is globally asymptotically stable at $\theta^{*}$.
\end{enumerate}
All the above facts are easy to verify. Now by appealing to the third extension of Theorem $2$, Section $2.2$, Chapter $2$ of \cite{borkar2008stochastic} and from the above observations, we can henceforth conclude that $\overbar{\theta}_{j} \rightarrow \theta^{*}$ almost surely as $j \rightarrow \infty$. This completes the proof of (\ref{eq:thbarconv}).
\end{Proof}
\section{Experimental Illustrations}
The performance of our algorithm is evaluated on four different MDP settings:
\begin{enumerate}
	\item
	Chain walk MDP.
	\item
	Linearized cart-pole balancing.
	\item
	$5$-link actuated pendulum balancing.
	\item
	Random MDP.
\end{enumerate} 
Our algorithm is compared against the state-of-the-art algorithms such as least squares policy iteration (LSPI), fast policy search method, model reference adaptive search (MRAS) and simultaneous perturbation stochastic approximation (SPSA). In each setting, the results shown are averages over $10$ independent sample sequences generated by the algorithms  with different initial conditions. The function $\varphi(\cdot)$ used here is $\varphi(x) = \exp(rx)$, where $r \in \bbbr_{+}$. 
\subsection{Experiment 1: Chain Walk}
This particular setting which has been proposed in \cite{koller2000policy} demonstrates the unique scenario where policy iteration is non-convergent when approximate value functions are employed instead of true ones. This particular example is also utilized to empirically evaluate the performance of LSPI in \cite{lagoudakis2003least}. Here, we compare the performance of our algorithm against LSPI and also against the stable Q-learning algorithm with linear function approximation (called Greedy-GQ) proposed in \cite{maei2010toward}. This particular demonstration is pertinent in two ways: (1) when LSPI was evaluated on this setting, the maximum state space cardinality considered was $50$. We consider here a larger MDP with $450$ states and $(2)$ the stable Greedy-GQ algorithm is only evaluated over a small experimental setting in \cite{maei2010toward}. Here, by applying it on a relatively harder setting, we attempt to assess its applicability and robustness.
\begin{figure}
	\centering
	\includegraphics[scale=0.34]{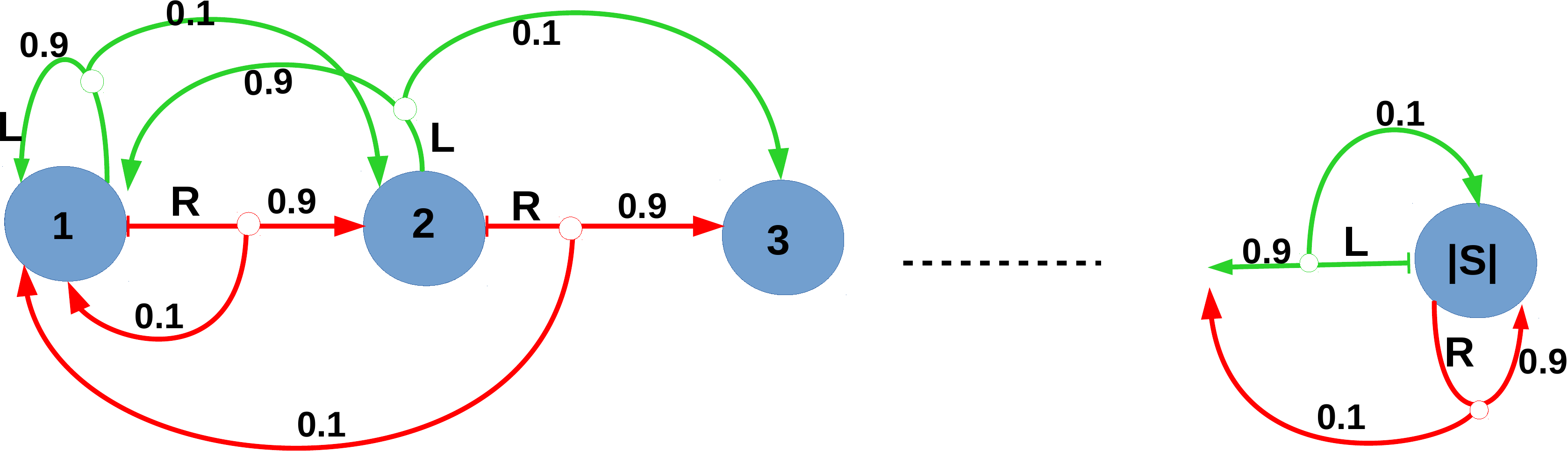}
	\caption{Chain walk  MDP}
\end{figure}   

\vspace*{2mm}
\noindent
\textbf{Setup: }We consider a  Markov decision process with $\vert \mathbb{S} \vert = 450$, $\mathbb{A} = \{L, R\}$, $k_1=5$, $k_2=10$ and the discount factor $\gamma = 0.99$.\vspace*{2mm}\\
\textbf{Reward function: }
$R(\cdot, \cdot, 150) = R(\cdot, \cdot, 300) = 1.0$ and zero for all other transitions. This implies that only the transitions to states $150$ and $300$ will acquire a positive payoff, while the rest are nugatory transitions. \vspace*{2mm}\\
\textbf{Transition dynamics: }The transition probability kernel is defined as follows:\vspace*{3mm}\\
{\setlength{\abovedisplayskip}{-11pt}\setlength{\belowdisplayskip}{2pt}\begin{flalign*}
\textrm{For }1 < s < \vert \mathbb{S} \vert\hspace*{5mm}
\begin{cases}
P(s, L, s+1) =  0.1, \hspace*{4mm} P(s, L, s-1) = 0.9,\\ 
P(s, R, s+1) = 0.9, \hspace*{4mm}  P(s, R, s-1) = 0.1.
\end{cases}
\end{flalign*}}
{\setlength{\abovedisplayskip}{-1pt}\setlength{\belowdisplayskip}{-5pt}\begin{flalign*}
&P(1, L, 2) =  0.1, \hspace*{4mm} P(1, L, 1) = 0.9,\\
&P(1, R, 2) =  0.9, \hspace*{4mm} P(1, R, 1) = 0.1,\\
&P(\vert \mathbb{S} \vert, L, \vert \mathbb{S} \vert) =  0.1, \hspace*{4mm} P(\vert \mathbb{S} \vert, L, \vert \mathbb{S} \vert-1) = 0.9,\\
&P(\vert \mathbb{S} \vert, R, \vert \mathbb{S} \vert) =  0.9, \hspace*{4mm} P(\vert \mathbb{S} \vert, R, \vert \mathbb{S} \vert-1) = 0.1,
\end{flalign*}}
\vspace{2mm}\\
\textbf{Feature set: } We employ radial basis functions (RBF) as both policy and prediction features. We utilize $5$ RBFs for prediction and $10$ for policy features, \emph{i.e.}, $k_1 = 5$ and $k_2 = 10$. Note that RBFs are Gaussian kernels which are parametrized by the centroid $m \in \bbbr$ and spread $v \in \bbbr_{+}$ and are expressed as:
\begin{equation}
b(s) = e^{-\frac{(s-m)^{2}}{2.0v^{2}}}.
\end{equation}
In our experiments, we initially tried to employ polynomials for features and found that the approximations they produced were quite poor. However, with RBFs one can indeed obtain decent performance by uniformly distributing the centroids in the state or state-action space and by considering the spread to be the half of the distance between subsequent centroids. In this way, one can indeed cover the respective spaces reasonably well. 
The policy features and the prediction features are defined as follows:\vspace*{3mm}\\
\noindent
\begin{minipage}{0.27\textwidth}
	\vspace*{0mm}
	\noindent\rule{12cm}{1.0pt}\vspace*{2mm}\\
	\hspace*{10mm}\textbf{\underline{Policy features}}\vspace*{0mm}\\
	\noindent
	\small
	\begin{eqnarray*}
	\psi(s,a) = \begin{pmatrix}
				I_{\{a = L\}}e^{-\frac{(s-m_1)^{2}}{2.0v_{1}^{2}}}\\ 
				\vdots\\
				I_{\{a = L\}}e^{-\frac{(s-m_5)^{2}}{2.0v_5^{2}}}\\
	I_{\{a = R\}}e^{-\frac{(s-m_1)^{2}}{2.0v_1^{2}}}\\ 
\vdots\\
I_{\{a = R\}}e^{-\frac{(s-m_5)^{2}}{2.0v_5^{2}}}			
	\end{pmatrix}.
	\end{eqnarray*}
	\null
	\normalsize
	\par\xdef\tpd{\the\prevdepth}
	\vspace*{0mm}
\end{minipage}
\hspace*{2cm}
\begin{minipage}{0.48\textwidth}
	\vspace*{0mm}
	\vspace*{-12mm}\vline height 100pt depth 50pt width 1pt\vspace*{-54mm}\hspace*{10mm}
	\begin{align*}
	&\textbf{\underline{Prediction features}}\vspace*{10mm}\\
	&\phi_{i}(s) = e^{-\frac{(s-m_{i})^{2}}{2.0v_{i}^{2}}}, \\\\\\\\\\\
	\end{align*}
\end{minipage}\\
where $m_i = 5+10(i-1), v_i = 5$, $1 \leq i \leq 5$.
\normalsize
\vspace*{2mm}\\
\textbf{Behaviour policy: } This is the most important choice and one has to be discreet while choosing the behaviour policy. For this setting, we prefer a policy which is unbiased and which uniformly covers the action space to provide sufficient exploration. Henceforth, by choosing $w_{b} = (0,0,\dots,0)^{\top}$ we obtain a uniform distribution over action space for every state in $\mathbb{S}$.\vspace*{2mm}\\
\textbf{Performance function:} Note that both LSPI and Q-learning seek in the policy parameter space to find the optimal or sub-optimal policy by recalibrating the parameter vector at each iteration in the direction of the improved value function. But the objective function  that we consider in this paper is a more generalized version involving the performance function $L$ and scalarization using $\mathbb{E}_{\nu_w}[\cdot]$. So the predicament, the above algorithms attempt to resolve becomes a special instance of our generalized version and hence to compare our algorithm against them, we consider the objective function to be the weighted Euclidean norm of the approximate value function (with weight being the stationary distribution $\nu_w$). Therefore, the performance function $L$ is defined as
$L(h_{w \vert w}) = h^{2}_{w \vert w}$ (where squaring of the vector is defined as squaring of each of its components). Note that, in our algorithm, we approximate $h_{w \vert w}$ using the behaviour policy and the true approximation and the stationary distribution involved are $\Phi x_{w \vert w_b}$ and $\nu_{w_b}$ respectively. However, since the behaviour policy chosen is the uniform distribution over the action space for each state in $\mathbb{S}$, one can easily deduce that the underlying Markov chain of the behaviour policy is a uniform random walk and its stationary distribution is the uniform distribution over the state space $\mathbb{S}$. 
\clearpage
\begin{figure}
	\centering
	\includegraphics[scale=0.32]{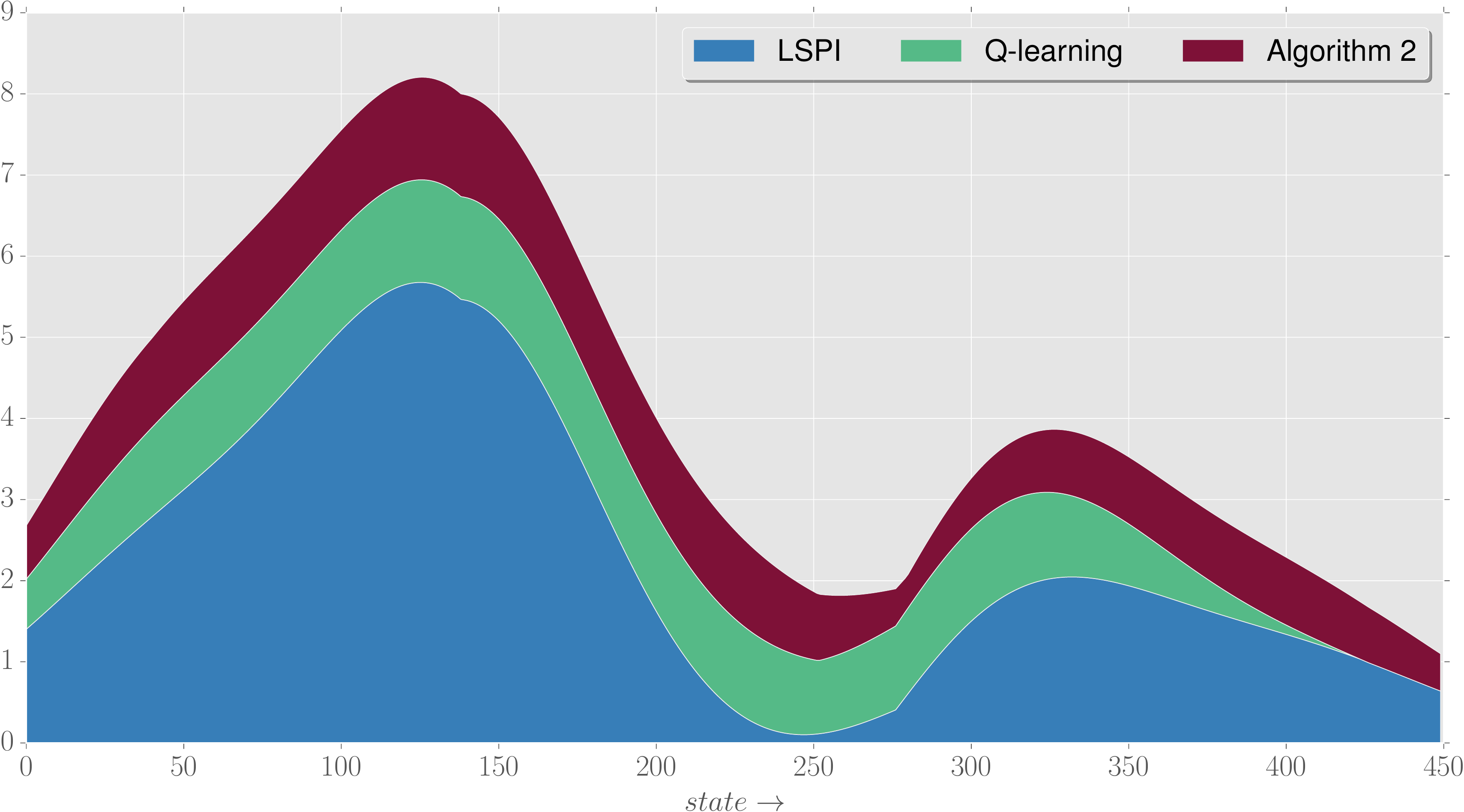}
	\caption{The plot of the respective optimal value functions contrived by LSPI, Q-learning and Algorithm \ref{alg:ce2-nd} for the chain walk MDP setting. The optimal solutions of various algorithms are being developed by averaging over $10$ independent trials. For Algorithm $2$, we averaged the various optimal solutions obtained for different sample trajectories generated using the same behaviour policy, but with different initial states which are chosen randomly. Our approach (Algorithm 2) literally surpassed other algorithms in terms of its quality. The random choice of the initial state effectively favoured sufficient exploration of the state space which directly assisted in generating high quality solutions.}
\end{figure}
\begin{table}[h]
\begin{center}
\caption{Algorithm parameter values used in the chain walk experiment}
\hspace*{0mm} \begin{tabular}{ | l | l |}
	\specialrule{.2em}{.04em}{.04em} 
	$\beta_{j}$\hspace*{10mm} & $0.2$\hspace*{23mm}\\[3pt] \hline
	\small$\overbar{\beta}_{j}$\normalsize\hspace*{10mm} & $0$\hspace*{22mm}\\[3pt] \hline
	$\zeta$\hspace*{10mm} & $0$\hspace*{22mm}\\[3pt] \hline
	$c_j$\hspace*{10mm} & $0.08$ \hspace*{23mm}\\[3pt] \hline
	$\rho$\hspace*{10mm} & $0.05$ \hspace*{22mm}\\[3pt]  \hline
	$\epsilon$\hspace*{10mm} & $0.9$ \hspace*{24mm}\\[3pt]  \hline
	$\tau$\hspace*{10mm} & $1.0$ \hspace*{24mm}\\[3pt]  \hline
	$r$\hspace*{10mm} & $0.01$ \hspace*{20mm}\\[3pt]  \hline
	\specialrule{.2em}{.02em}{.02em} 
\end{tabular}
\end{center}
\end{table}
\clearpage
\subsection{Experiment $2$: Linearized Cart-pole Balancing \cite{dann2014policy}}
\noindent
\textbf{Setup: } A pole with mass $m$ and length $l$ is connected to a cart of mass $M$. It can rotate in the interval $[-\pi, \pi]$ with negative angle representing the rotation in the counter clockwise direction. The cart is free to move in either direction within the bounds of a linear track and the distance lies in the region $[-4.0, 4.0]$ with negative distance representing the movement to the left of the origin.In our experiment, we have $m = 0.5$, $M = 0.5$, $l = 20.5$ and the discount factor $\gamma = 0.1$.\vspace*{2mm}\\
\textbf{Goal: } To bring the cart to the equilibrium position, \emph{i.e.}, to balance the pole upright and the cart at the centre of the track.\vspace*{1mm}\\
\textbf{State space: }The state is the 4-tuple $(x, \dot{x}, \psi, \dot{\psi})^{\top}$ where $\psi$ is  the angle of the pendulum \emph{w.r.t.} the vertical axis, $\dot{\psi}$ is the angular velocity, $x$ the relative cart position from the centre of the track and $\dot{x}$ is its velocity. For better tractability, we restrict $\dot{x} \in [-5.0, 5.0]$ and $\dot{\psi} \in [-5.0, 5.0]$, respectively.\vspace*{2mm}\\
\textbf{Control (Policy) space: }The controller applies a horizontal force $a$ on the cart parallel to the track. The stochastic policy used in this setting corresponds to $\pi(a|s) = \mathcal{N}(a | \vartheta^{\top}s, \sigma^{2})$ (normal distribution with mean ${\vartheta}^{\top}s$ and standard deviation $\sigma$). Here the policy is parametrized by $\vartheta \in \bbbr^{4}$ and $\sigma \in \bbbr$.\vspace*{2mm}\\
\textbf{System dynamics: }
The dynamical equations of the system are given by
\begin{equation}
\ddot{\psi} = \frac{-3ml\dot{\psi}^{2}\sin{\psi}\cos{\psi}+(6M+m)g\sin{\psi}-6(a-b\dot{\psi})\cos{\psi}}{4l(M+m)-3ml\cos{\psi}},
\end{equation}
\begin{equation}
\ddot{x} = \frac{-2ml\dot{\psi}^{2}\sin{\psi}+3mg\sin{\psi}\cos{\psi}+4a-4b\dot{\psi}}{4(M+m)-3m\cos{\psi}}.
\end{equation}
By making further assumptions on the initial conditions, the system dynamics can be approximated accurately by the linear system 
\begin{equation}
\begin{bmatrix} 
x_{t+1}\\
\dot{x}_{t+1}\\
\psi_{t+1}\\
\dot{\psi}_{t+1}
\end{bmatrix} = \begin{bmatrix} 
x_{t}\\
\dot{x}_{t}\\
\psi_{t}\\
\dot{\psi}_{t}
\end{bmatrix} + \Delta t \begin{bmatrix}
\dot{\psi}_{t} \\
\frac{3(M+m)\psi_t-3a+3b\dot{\psi_t}}{4Ml-ml} \\
\dot{x}_{t} \\
\frac{3mg\psi_t + 4a - 4b\dot{\psi_t}}{4M-m}
\end{bmatrix} + 
\begin{bmatrix}
0 \\
0 \\
0 \\
\mathbf{z}
\end{bmatrix},
\end{equation}
where $b$ is the friction coefficient of the cart on the floor, $g = 9.81\frac{m}{sec^{2}}$ is the gravitational constant, $\Delta t$ is the integration time step, \emph{i.e.}, the time difference between two transitions and $\mathbf{z}$ is a standard Gaussian noise on the velocity of the cart. In our experiment, we set $b = 0.1Newton(msec)^{-1}$ and $\Delta t = 0.1sec$, respectively.\vspace*{2mm}\\
\textbf{Reward function: }
$R(s, a) = R(\psi, \dot{\psi}, x, \dot{x}, a) = -4\psi^2 - x^2 - 0.1a^2$. The reward function can be viewed as assigning penalty which is directly proportional to the deviation from the equilibrium state.\vspace*{2mm}\\
\clearpage
\noindent
\textbf{Prediction features: } $\phi(s \in \bbbr^{4}) = (1, s_{1}^{2}, s_{2}^{2} \dots, s_{1}s_{2}, s_{1}s_{3}, \dots, s_{3}s_{4})^{\top} \in \bbbr^{11}$.\vspace*{2mm}\\
\textbf{Behaviour policy: } $\pi_{b}(a|s) = \mathcal{N}(a | \vartheta_{b}^{\top}s, \sigma_{b}^{2})$,
where $\vartheta_{b} = (3.684, 3.193, 4.252,$ $3.401)^{\top}$ and $\sigma_{b} = 5.01$. The behaviour policy is determined by vaguely solving the problem using true value functions and then choosing the behaviour policy vector $\vartheta_b$ by perturbing each component of the vague solution so obtained. The margin of perturbation we considered is chosen randomly from the interval $[-5.0, 5.0]$.\vspace*{2mm}\\
\textbf{Performance function:} The performance function $L$ is defined as under: We randomly select (from the given intervals described in the definition of the state space), $s_{0} = (0.235, 3.581, 2.276, 1.069)^{\top}$. Now, define
\begin{equation}\label{eq:exp2perffunc}
\begin{aligned}
	L(h_{w \vert w})(s) = 
	\begin{cases}
		0.1h_{w \vert w}(s_0), \textrm{ for } s = s_{0}\vspace*{2mm}\\
	0, \forall s \in \mathbb{S}\setminus\{s_{0}\}.
	\end{cases}
\end{aligned}
\end{equation}
Here $s_{0}$ is the initial state of the cart-pole system which implies that the cart is initially stationed at a distance of 0.235 from the centre and the pendulum is at an angle of 2.276 ($=\frac{\pi}{1.38}$) from the vertical position. The initial velocity
of the cart and the angular velocity of the pendulum are 3.581 and 1.069 respectively. The goal is to find the optimal policy (which corresponds to the parameters of the horizontal force) to bring the cart to the equilibrium position, \emph{i.e.}, cart at the centre of the track
and the pendulum in the vertical position. The nature of the performance function $L$ in Equation (\ref{eq:exp2perffunc}) is to explicitly capture this aspect of the problem, \emph{i.e.}, to find the optimal policy that takes the cart from $s_{0}$ to the equilibrium 
position and hence, only the cumulative cost incurred starting from $s_{0}$ is considered. Note that $s_{0}$ is chosen arbitrarily for the experiment and thus does not render any particular advantage to any of the algorithms.
\clearpage
\begin{figure}[h]
	\begin{subfigure}[h]{0.45\textwidth}
		\fbox{\includegraphics[width=52mm, height=44mm]{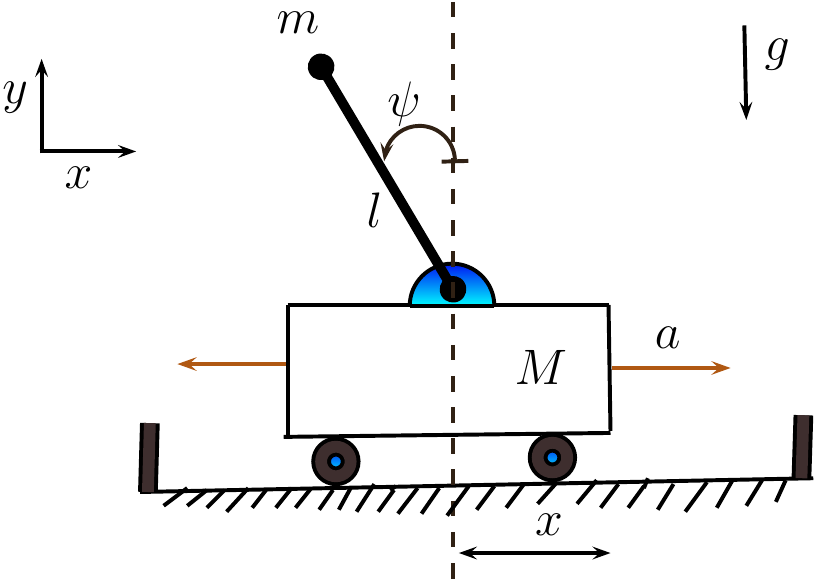}}
		\subcaption{Cart-pole setup}
	\end{subfigure}%
	\begin{subfigure}[h]{0.55\textwidth}
		\vspace*{0mm}\hspace*{0mm}
		\fbox{\includegraphics[width=62mm, height=44mm]{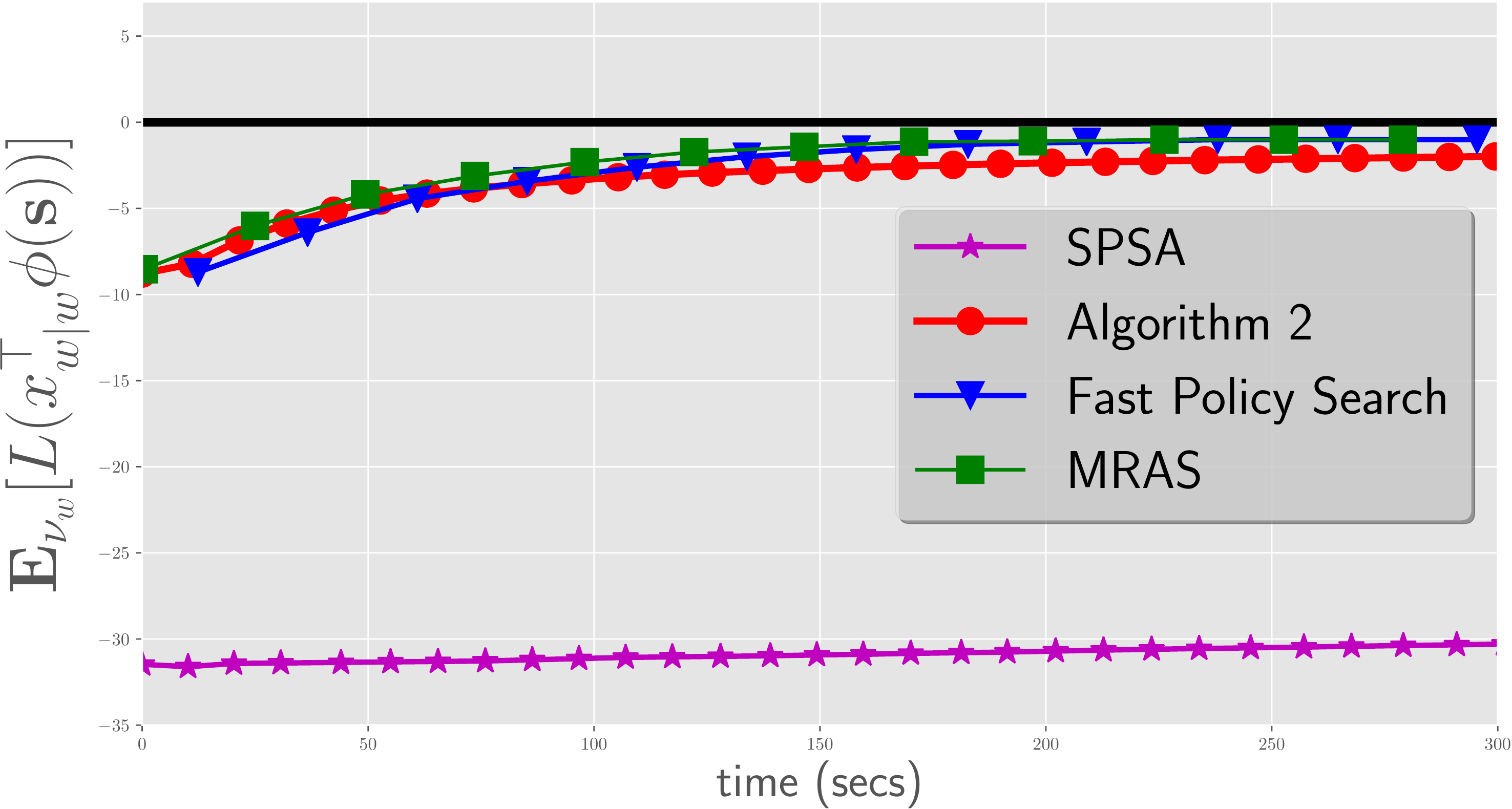}}
		\subcaption{Cart-pole results}
	\end{subfigure}
	\caption{(a) The cart-pole system: The goal is to keep the pole in the  upright position and the cart at the centre of the track by applying a force $a$ either to the right or to the left. The system is parametrized by the position $x$ of the cart, the angle of the pole $\psi$, the velocity $\dot{x}$ and the angular velocity $\dot{\psi}$. 
	(b) Here, for Algorithm \ref{alg:ce2-nd}, we plot $\mathbb{E}_{\nu_{w_b}}\left[L\left(x_{\overbar{\mu}_{j} \vert w_b}^{\top}\phi(\mathbf{s})\right)\right]$, where $\overbar{\mu}_j$ is the mean vector of the Polyak averaged model sequence $\{\overbar{\theta}_j\}$, \emph{i.e.}, $\overbar{\theta}_{j} = (\overbar{\mu}_{j}, \overbar{\Sigma}_{j})^{\top}$. For the other algorithms, \emph{i.e.}, SPSA, MRAS and fast policy search, we plot $\mathbb{E}_{\nu_{w_j}}\left[L\left(x_{w_{j} \vert w_j}^{\top}\phi(\mathbf{s})\right)\right]$, where $\{w_j \in \mathbb{W}\}$ is the iterative sequence generated by the respective algorithms. This implies that Algorithm \ref{alg:ce2-nd} operates in the off-policy setting, while the rest of the algorithms utilize on-policy value function approximations to generate the optimal policy vector. With this advantage, the algorithms SPSA, MRAS and fast policy search are expected to perform better as they have complete access to the generative model unlike Algorithm \ref{alg:ce2-nd} which has access only to the sample trajectory generated by the behaviour policy. Also, note that $x$-axis is time in seconds relative to the start of the algorithm since MRAS and fast policy search are batch based approaches, while Algorithm \ref{alg:ce2-nd} and SPSA are incremental schemes.
	\textit{Regarding the accuracy of the solution obtained by our algorithm, note that the global optimum is indeed zero, since the reward function is defined as the negative penalty with respect to the deviation from the equilibrium position and the goal is to bring the cart to the equilibrium position.}}\label{fig:pendres1}
\end{figure}
\subsection{Experiment $3$: $5$-Link Actuated Pendulum Balancing \cite{dann2014policy}}
\noindent
\textbf{Setup: } $5$ independent poles each with mass $m$ and length $l$ with the top pole being a pendulum connected using $5$ rotational joints. In our experiment, we take $m = 1.5$, $l = 10.0$ and the discount factor $\gamma = 0.1$.\vspace*{2mm}\\
\textbf{Goal: } To keep all the poles in the horizontal position by applying independent torques at each joint.\vspace*{2mm}\\
\textbf{State space: }The state $s = (q, \dot{q})^{\top} \in \bbbr^{10}$ where $q = (\psi_{1}, \psi_{2}, \psi_{3}, \psi_{4}, \psi_{5}) \in \bbbr^{5}$ and $\dot{q} = (\dot{\psi}_{1}, \dot{\psi}_{2}, \dot{\psi}_{3}, \dot{\psi}_{4}, \dot{\psi}_{5})  \in \bbbr^{5}$ with $\psi_{i}$ being the angle of the pole $i$ \emph{w.r.t.} the horizontal axis and $\dot{\psi}_{i}$ is the angular velocity. In our experiment, we consider the following bounds on the state space: $\psi_i \in [-\pi, \pi]$, $\forall 1 \leq i \leq 5$ and $\dot{\psi}_i \in [-5.0, 5.0]$, $\forall 1 \leq i \leq 5$.\vspace*{2mm}\\
\textbf{Control space: }The action $a = (a_{1}, a_{2}, \dots, a_{5})^{\top} \in \bbbr^{5}$ where $a_{i}$ is the torque applied to the joint $i$. The stochastic policy used in this setting corresponds to 
\begin{equation}\label{eq:exp2ctrlspace}
\pi(a|s) = \mathcal{N}_{5}(a | A s, B) \hspace*{3mm} \textrm{where} \hspace*{3mm} A \in \bbbr^{5 \times 10}, B \in \bbbr^{5 \times 5}.
\end{equation}
We assume that the torques $a_i$ applied at each joint are independent and hence $B$ is a diagonal matrix. The policy parameter space $\mathbb{W}$ is defined as $\mathbb{W} = \{w \in \bbbr^{55} \vert w = (A_{00}, A_{01}, A_{02}, \dots, A_{48}, A_{49}, B_{00}, B_{11}, \dots, B_{44})^{\top}\}$.\vspace*{2mm}\\
\textbf{System dynamics: } The state equations representing the approximate linear system dynamics are given by
\begin{equation} 
\begin{bmatrix} 
q_{t+1}\\
\dot{q}_{t+1}
\end{bmatrix} = 
\begin{bmatrix} 
I && \Delta t\hspace*{1mm} I\\
-\Delta t \hspace*{1mm}M^{-1}U && I
\end{bmatrix}\begin{bmatrix}q_{t}\\ \dot{q}_{t}\end{bmatrix} + \Delta t \begin{bmatrix}
0 \\
M^{-1}
\end{bmatrix}a + 
\mathbf{z}
\end{equation}
where $\Delta t$ is the integration time step, \emph{i.e.}, the time difference between two transitions and $M$ is the mass matrix in the horizontal position with $M_{ij} = l^{2}(6-max(i,j))m$. $U$ is a diagonal matrix with $U_{ii} = -gl(6-i)m$, where $g$ is the gravitational constant.  Each component of $\mathbf{z}$ is a standard Gaussian noise. In our experiment, we take $\Delta t = 0.1$ and $g=9.8$.\vspace*{2mm}\\
\textbf{Reward function: } $R(q, \dot{q}, a) = -q^{\top}q$. The reward function can be viewed as assigning penalty (negative reward) with respect to the deviation from the optimal pole position (the unique position with zero deviation from the horizontal position and hence attracts no penalty, \emph{i.e.}, highest reward).\vspace*{2mm}\\
\textbf{Feature vectors: } $\phi(s \in \bbbr^{10}) = (1, s_{1}^{2}, s_{2}^{2} \dots, s_{1}s_{2}, s_{1}s_{3}, \dots, s_{9}s_{10})^{\top} \in \bbbr^{46}$.\vspace*{2mm}\\
\textbf{Behaviour policy: }The behaviour policy considered in the experiment is given by $\pi_{b}(a | s) = \mathcal{N}_{5}(a | A_{b}s, B_{b})$, where 
\begin{eqnarray*}
{A^{\top}_b = \begin{pmatrix}
        5.794 &  2.000 &  6.230 &  4.500 &  6.145 \\
        4.843 &  5.014 &  2.306 &  2.796 &  7.000 \\
        6.031 &  6.500 &  6.600 &  8.379 &  4.252 \\
        6.640 &  3.424 &  5.937 &  5.045 &  3.617 \\
        8.661 &  3.463 &  4.430 &  3.000 &  4.233 \\
        5.660 &  3.437 &  7.275 &  7.417 &  5.755 \\
        3.781 &  2.989 &  4.756 &  6.417 &  6.760 \\
        3.391 &  3.696 &  4.153 &  5.761 &  3.196 \\
        5.725 &  2.929 &  3.205 &  3.631 &  8.651 \\
        1.337 &  4.677 &  8.009 &  3.609 &  5.602
\end{pmatrix}}
\textrm{ and }
{B_b = \begin{pmatrix}
         5.0 &   & & \textbf{O} &   \\
         & 5.0 &  &  &   \\
         &  &  5.0 &  &  \\
         &  &  &  5.0 &  \\
         & \textbf{O}  &  &  &  5.0
\end{pmatrix}.}
\end{eqnarray*} 
\normalsize
The methodology employed to induce the behaviour policy in this case is similar to that of the cart-pole setting.\vspace*{2mm}\\
\textbf{Performance function:} The performance function $L$ is defined as under:
We randomly select (from the given intervals described in the definition of the state space), $s_{0} =$  $(  -1.515, -2.437, -1.386, -3.041, 0.001, 4.510, 0.691, 1.450, 3.241,$ $3.535)^{\top}$. Now define
\begin{equation}
	L(h_{w \vert w})(s) = 
	\begin{cases}
		0.1h_{w \vert w}(s_{0}), \textrm{ for } s = s_{0}\vspace*{2mm}\\
		0, \forall s \in \mathbb{S} \setminus \{s_{0}\}.
	\end{cases}
\end{equation}
The rationale behind the choice of the above particular performance function is similar to that of Experiment $2$. Also, note that $s_{0}$ is chosen arbitrarily for the experiment and thus does not accord any unfounded predisposition to any of the algorithms.
\begin{table}[h]
	\begin{center}
		\caption{Algorithm parameter values used in the experiments. Note that $\{j_{(n)}\}$ is the subsequence of $\{j\}$ when recursion (\ref{eq:algbarth}) is executed.}
		\vspace*{2mm}
		\hspace*{0mm} \begin{tabular}{ | l | l | l |}
			\specialrule{.2em}{.04em}{.04em} 
			& \textbf{Cart-pole experiment} \hspace*{5mm}& \hspace*{1mm}\textbf{Actuated pendulum balancing}\hspace*{3mm}\\[3pt]  \hline
			$\beta_{j}$\hspace*{10mm} & $0.7$\hspace*{34mm} & $0.7$\hspace*{44mm} \\[3pt]  \hline
			\small$\overbar{\beta}_{j}$\normalsize\hspace*{10mm} & $j^{-1}_{(n)}$\hspace*{34mm} & $j^{-1}_{(n)}$\hspace*{44mm} \\[3pt]  \hline
			$\zeta$\hspace*{10mm} & $j^{-1}_{(n)}$\hspace*{34mm} & $j^{-1}_{(n)}$ \hspace*{44mm} \\[3pt]  \hline
			$\lambda$\hspace*{10mm} & $0.1$ \hspace*{33mm}& $0.1$ \hspace*{44mm} \\[3pt]  \hline
			$c_j$\hspace*{10mm} & $0.1$ \hspace*{33mm}& $0.1$ \hspace*{44mm} \\[3pt]  \hline
			$\rho$\hspace*{10mm} & $0.01$ \hspace*{32mm}& $0.01$ \hspace*{42mm} \\[3pt]  \hline
			$\epsilon$\hspace*{10mm} & $0.9$ \hspace*{33mm}& $0.9$\hspace*{45mm} \\[3pt]  \hline
			$r$\hspace*{10mm} & $0.01$ \hspace*{32mm}& $0.01$\hspace*{43mm} \\[3pt]  \hline
			$N_{j}$\hspace*{10mm} & $4000, \forall j$ \hspace*{28mm} & $4000, \forall j$\hspace*{40mm} \\[3pt]  \hline
			\specialrule{.2em}{.02em}{.02em} 
		\end{tabular}
	\end{center}
\end{table}
\normalsize
\clearpage
\begin{figure}[h]
	        \begin{subfigure}[h]{0.45\textwidth}
            \fbox{\includegraphics[width=52mm, height=44mm]{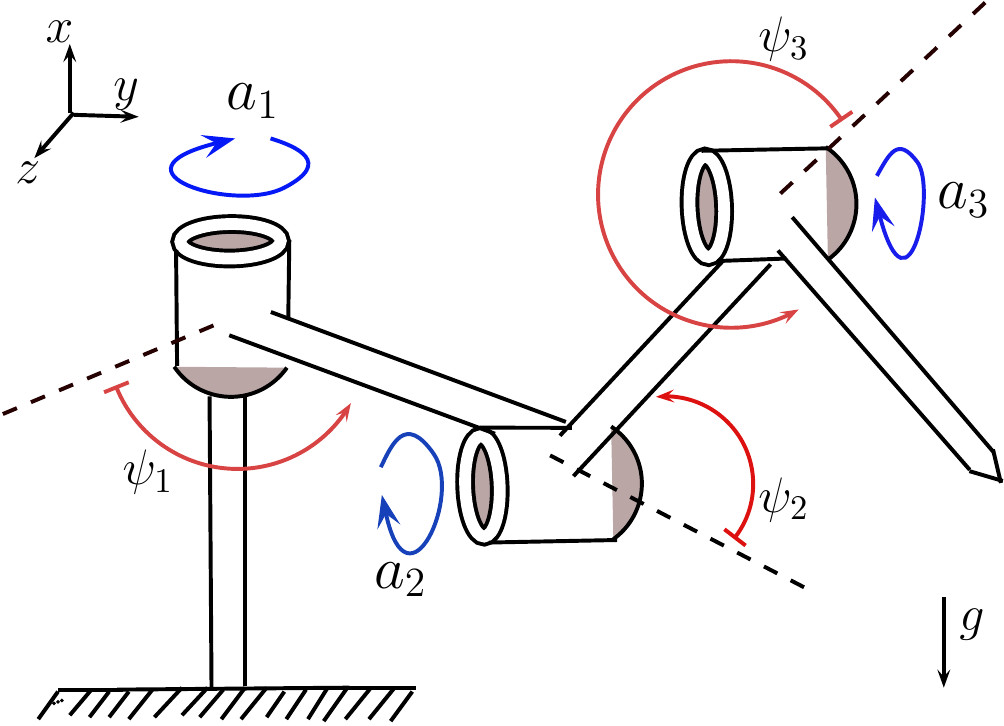}}
            \subcaption{$3$-link actuated pendulum setting}\label{fig:armpic}
        \end{subfigure}
        \begin{subfigure}[h]{0.55\textwidth}
            \vspace*{0mm}\fbox{\hspace*{0mm}\includegraphics[width=62mm, height=44mm]{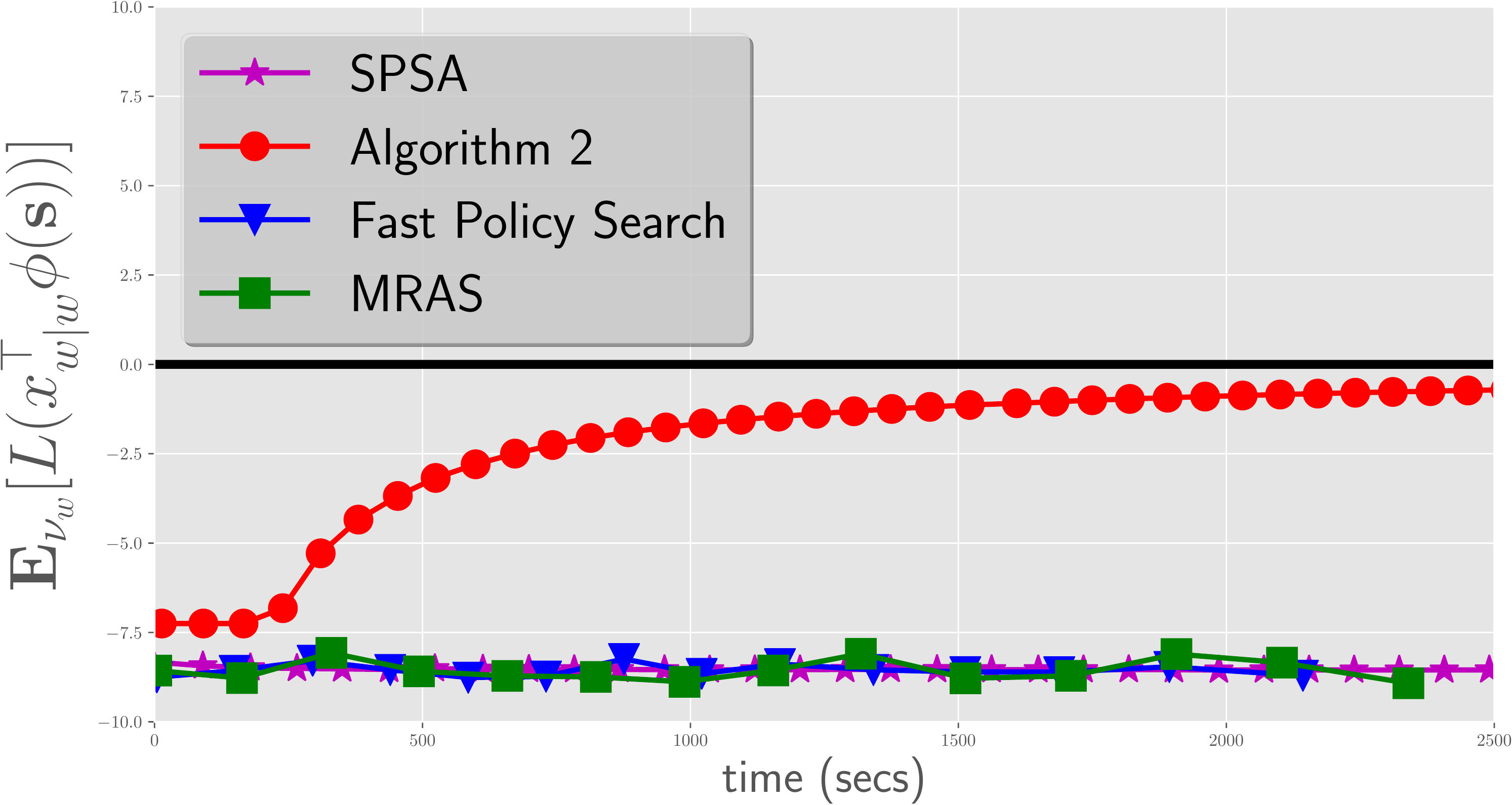}\hspace*{1mm}}
		\subcaption{$5$-link actuated pendulum results}
        \end{subfigure}
	\caption{(a) Each rotational joint $i$, $1 \leq i \leq 3$ is independently actuated by a torque $a_{i}$. The system is parametrized by the angle $\psi_{i}$ against the horizontal direction and the angular velocity $\dot{\psi}_{i}$. The goal is to balance the pole in the horizontal direction, \emph{i.e.}, all $\psi_{i}$ should be as close to $0$ as possible by actuating Gaussian torques $a_i$ (Equation (\ref{eq:exp2ctrlspace})). 
	(b) Here, for Algorithm \ref{alg:ce2-nd}, we plot $\mathbb{E}_{\nu_{w_b}}\left[L\left(x_{\overbar{\mu}_{j} \vert w_b}^{\top}\phi(\mathbf{s})\right)\right]$, where $\overbar{\mu}_j$ is the mean vector of the Polyak averaged model sequence $\{\overbar{\theta}_j\}$, \emph{i.e.}, $\overbar{\theta}_{j} = (\overbar{\mu}_{j}, \overbar{\Sigma}_{j})^{\top}$. For the other algorithms, \emph{i.e.}, SPSA, MRAS and fast policy search, we plot $\mathbb{E}_{\nu_{w_j}}\left[L\left(x_{w_{j} \vert w_j}^{\top}\phi(\mathbf{s})\right)\right]$, where $\{w_j \in \mathbb{W}\}$ is the iterative sequence generated by the respective algorithms. This implies that Algorithm \ref{alg:ce2-nd} operates in the off-policy setting, while the rest of the algorithms utilize on-policy value function approximations to generate the optimal policy vector. With this advantage, the algorithms MRAS, SPSA and fast policy search are expected to perform better as they have unrestricted access to the generative model unlike Algorithm \ref{alg:ce2-nd} which has access only to a sample trajectory generated by the behaviour policy. Also, note that $x$-axis is time in seconds relative to the start of the algorithm since MRAS and fast policy search are batch based approaches, while Algorithm \ref{alg:ce2-nd} and SPSA are incremental schemes.
	\textit{Again, regarding the accuracy of the solution obtained by our algorithm, note that the global optimum is indeed zero, since the reward function is defined as the negative penalty with respect to the deviation from the equilibrium position and the goal is to bring the system to the equilibrium position}.}\label{fig:armres1}
\end{figure}%
\clearpage
\subsection{Experiment $4$: Random MDP}
\noindent
\textbf{Setup: }We consider a randomly generated Markov decision process with $\vert \mathbb{S} \vert = 500$, $\vert \mathbb{A} \vert = 30$, $k_1=5$, $k_2=5$ and $\gamma = 0.8$.\vspace*{3mm}\\
\textbf{Reward function: }The reward function $R$ is defined as follows:
\begin{flalign}
R(s, a, s^{'}) = \omega_{1}(s)\omega_{1}(s^{'})\left(\frac{\sin{(a)}+2.0}{(1.0+s{'})^{0.25}}\right), \hspace*{5mm}s, s^{'} \in \mathbb{S}, a \in \mathbb{A}.
\end{flalign}
Here $\omega_{1} \in [3,5]^{\vert \mathbb{S} \vert}$ is initialized for the algorithm with $\omega_{1}(s) \sim U(1,4)$.\vspace*{3mm}\\
\textbf{Transition dynamics: }The transition probability kernel $P$ is defined as follows:
\begin{flalign}\label{eqn:prnd}
P(s, a, s^{'}) =  {n \choose s^{'}}\omega_{2}(s, a)^{s^{'}}(1.0-\omega_{2}(s, a))^{n - s^{'}}, \hspace*{5mm} s, s^{'} \in \mathbb{S}, a \in \mathbb{A}.
\end{flalign}
Here the matrix $\omega_{2} \in [0,1]^{\mathbb{S} \times \mathbb{A}}$ is initialized for the algorithm with $\omega_{2}(s, a) \sim U(0,1)$.\vspace*{3mm}\\
\textbf{Feature set: }The policy features and the prediction features are as follows:\vspace*{2mm}\\
\noindent
\begin{minipage}{0.27\textwidth}
\vspace*{-8mm}
\noindent\rule{12cm}{0.4pt}\vspace*{1mm}\\
\hspace*{5mm}\textbf{\underline{Policy features}}\vspace*{2mm}\\
\noindent
\hspace*{0mm}$\psi(s, a) = B[s\vert \mathbb{A} \vert + a]$
\small
\begin{eqnarray*}
\hspace*{5mm}\textrm{where }{B = \begin{pmatrix}
 1 & 0 & 0 & 0 & 0\\ 
 0 & 1 & 0 & 0 & 0\\
 0 & 0 & 1 & 0 & 0\\
 0 & 0 & 0 & 1 & 0\\
 0 & 0 & 0 & 0 & 1\\
 1 & 0 & 0 & 0 & 0\\
 0 & 1 & 0 & 0 & 0\\
 \vdots & & \ddots & &\vdots\\
\end{pmatrix}.}_{15000 \times 5}
\end{eqnarray*}
\null
\normalsize
\par\xdef\tpd{\the\prevdepth}
\vspace*{-18mm}
\end{minipage}
\hspace*{2cm}
\begin{minipage}{0.48\textwidth}
\begin{equation*}
\hspace*{0mm}\vline\hspace*{10mm}\vspace*{-6mm}
\begin{aligned}
&\textbf{\underline{Prediction features}}\vspace*{2mm}\\
&\hspace*{-4mm}\phi_{i}(s) = e^{-\frac{(s-m_{i})^{2}}{2.0v_{i}^{2}}}, \\
&\textrm{ where } m_i = 5+10(i-1), v_i = 5.\\\\\\\\\\\
\end{aligned}
\end{equation*}
\end{minipage}\vspace*{4mm}\\\\\\
\normalsize
In this experimental setting, we employ the Gibbs ``softmax'' policies defined in Equation (\ref{eq:stplcy}).\vspace*{3mm}\\
\textbf{Behaviour policy: } The behaviour policy vector $w_b$ considered for the experiment is $w_{b} = (12.774, 15.615, 20.626, 25.877, 11.945)^{\top}$.\vspace*{2mm}\\
\textbf{Performance function:} The performance function $L$ is defined as follows: \\
$L(h_{w \vert w}) = 0.1 h^{2}_{w \vert w}$ (Note that squaring the vector here corresponds to co-ordinate wise squaring).\vspace*{2mm}\\
\clearpage
\noindent
\begin{minipage}{0.4\textwidth}
\begin{table}[H]
\begin{center}
\caption{Algorithm parameter values used in the random MDP experiment. Note that $\{j_{(n)}\}$ is the sub-sequence of $\{j\}$ when recursion (\ref{eq:algbarth}) is executed.}
\hspace*{0mm} \begin{tabular}{ | c | c |}
  \specialrule{.2em}{.04em}{.04em} 
	$\beta_{j}$\hspace*{10mm} & $0.7$\hspace*{23mm}\\[2pt] \hline
    \small$\overbar{\beta}_{j}$\normalsize\hspace*{10mm} & $j^{-1}_{(n)}$\hspace*{22mm}\\[2pt]  \hline
    $\zeta$\hspace*{10mm} & $j^{-1}_{(n)}$\hspace*{22mm}\\[2pt]  \hline
    $c_j$\hspace*{10mm} & $0.1$ \hspace*{23mm}\\[2pt]  \hline
    $\rho$\hspace*{10mm} & $0.01$ \hspace*{22mm}\\[2pt]  \hline
    $\epsilon$\hspace*{10mm} & $0.9$ \hspace*{24mm}\\[2pt]  \hline
    $\tau$\hspace*{10mm} & $10^{3}$ \hspace*{24mm}\\[2pt]  \hline
    $r$\hspace*{10mm} & $0.001$ \hspace*{20mm}\\[2pt]  \hline
    $N_{j}$\hspace*{10mm} & $1000, \forall j$ \hspace*{18mm}\\[2pt]  \hline
   \specialrule{.2em}{.02em}{.02em} 
  \end{tabular}
\end{center}
\end{table}
\null
\normalsize
\par\xdef\tpd{\the\prevdepth}
\small
\end{minipage}
\hspace*{5mm}
\begin{minipage}{0.5\textwidth}
\begin{figure}[H]
\centering
\fbox{\includegraphics[width=0.94\linewidth, height=49mm]{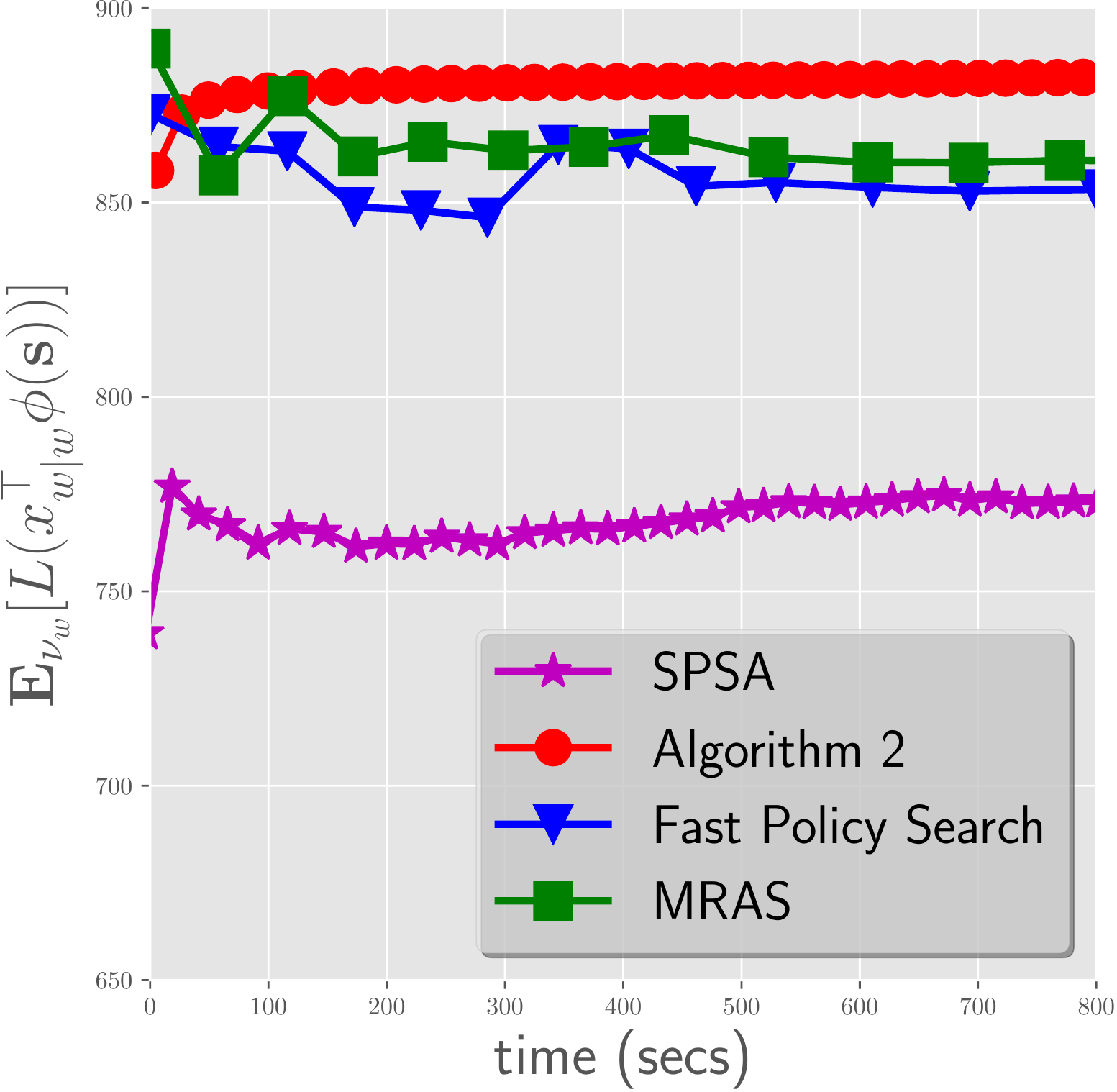}}
\caption{Plot of the results obtained in the random MDP experiment. Here also, $x$-axis is time in secs relative to the start of the algorithm.}
\end{figure}
\end{minipage}
\normalsize

As with the previous two experiments, Algorithm \ref{alg:ce2-nd} was run for the off-policy case while SPSA, MRAS and fast policy search were run for the on-policy setting.
\subsection{Exegesis of the experiments}\label{sec:exegesis-of-the-experiments}
In this section, we summarize the inferences drawn from the above experiments:\vspace*{3mm}\\
($1$) The proposed algorithm performed better than the state-of-the-art methods without compromising on the rate of convergence. The choice of the underlying behaviour policy indeed influenced this improved performance. Note that to labour high quality solutions, the choice of the behaviour policy is pivotal. In Experiment $1$, we considered a uniform policy, where every action is equally likely to be chosen for each state in $\mathbb{S}$. 
The results obtained in that experiment are quite promising, since, by only utilizing a uniform behaviour policy, we were able to grind out superior quality solutions. One has to justify the results to add credibility, considering the fact that LSPI is shown to produce optimal policy given a generative model.
Note that in the original LSPI paper, we find that the LSPI method utilizes a sample trajectory provided in the form of tuples $\{(s_i, a_i, r_i, s_i^{\prime})\}_{i \in \mathbb{N}}$, where $s_i$ and $a_i$ are drawn uniformly randomly from $\mathbb{S}$ and $\mathbb{A}$ respectively, while $s_{i}^{\prime}$ is the transitioned state given $s_i$ and $a_i$ by following the underlying transition dynamics of the MDP and  $r_i$ is the immediate reward for that transition.
One can immediately see that the information content required to generate such a trajectory is equivalent to that of maintaining a generative model. 
\begin{wrapfigure}{r}{0.45\textwidth}
	\vspace{-30pt}
		\hspace*{-17mm}\includegraphics[width=0.7\textwidth, height=55mm]{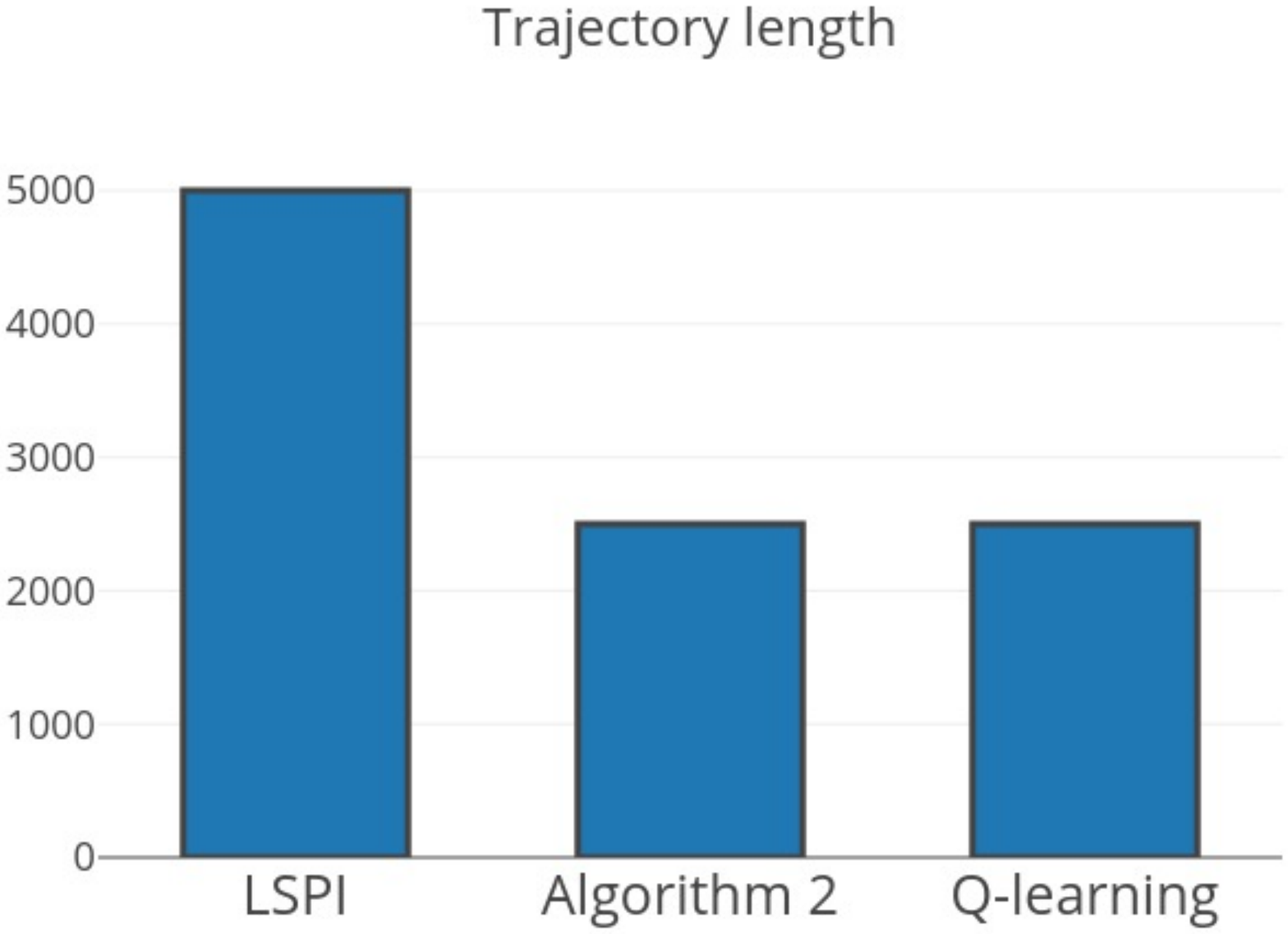}
	\vspace{-40pt}
\end{wrapfigure}   
Further, in \cite{lagoudakis2003least}, where LSPI is being empirically evaluated, we find that a trajectory length of $5000$ is being used in the $20$-state chain walk to obtain optimal performance. However, in our experiment (Experiment $1$) with $450$ states, we only consider a trajectory length of $5000$ for LSPI and hence obtain the sub-optimal performance. But, one should also consider the fact that the behaviour policy utilized by our algorithm in the same experiment is uniform (no prior information about the MDP is being  availed) and the trajectory length is only half of that of LSPI. Now, regarding the performance of Q-learning, we know (from Theorem $1$ of \cite{maei2010toward}) that the method can only provide sub-optimal solutions.  

\begin{figure}[h]
	\centering
	\includegraphics[scale=0.27]{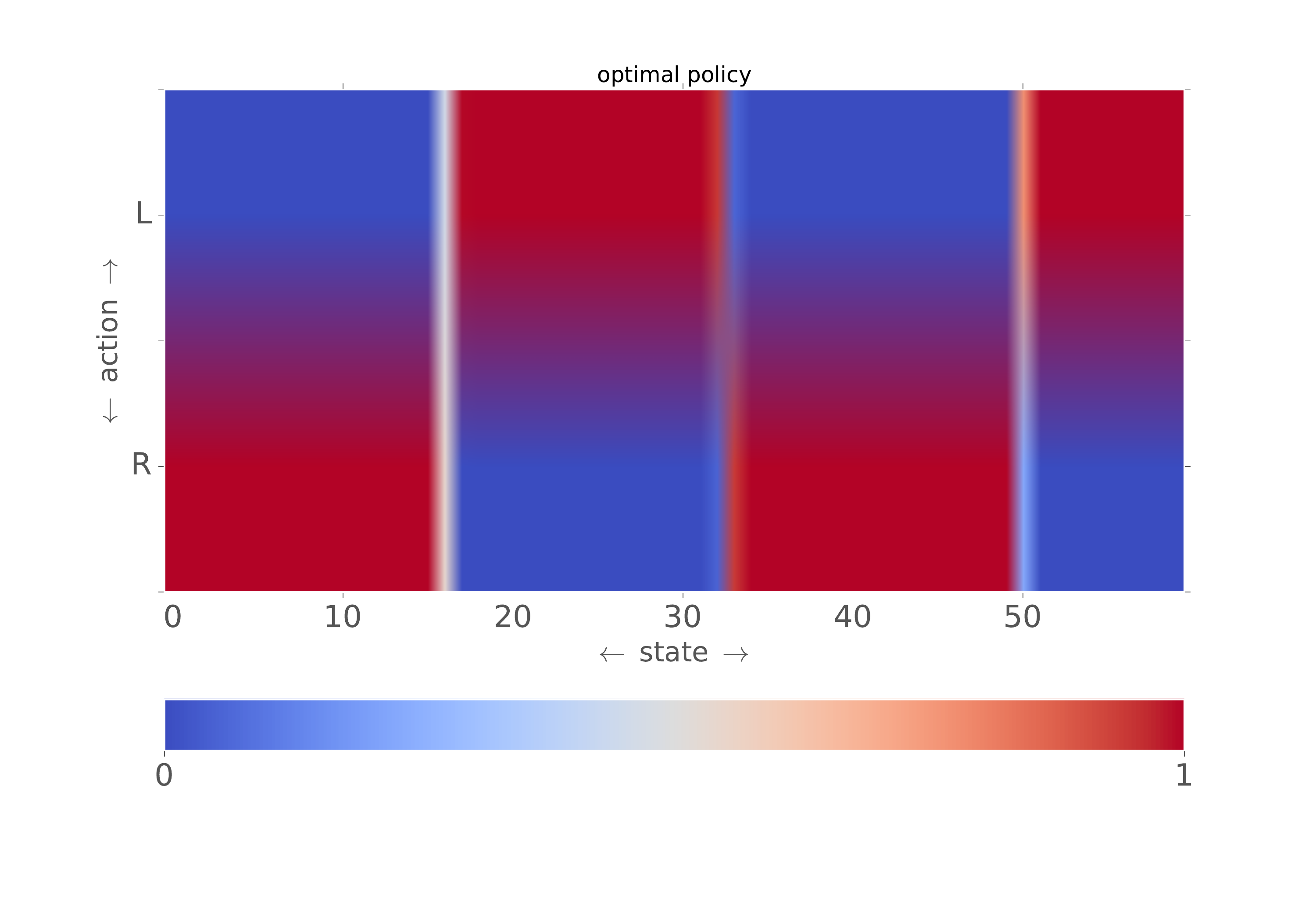}
	\caption{The schematic diagram of the optimal policy generated by Algorithm $2$ for the chain walk MDP with $\vert \mathbb{S} \vert = 60$, $\mathbb{A} = \{L, R\}$ and the discount factor $\gamma = 0.01$.}\label{fig:resgamma01}
\end{figure}

\begin{figure}[h]
	\centering
	\includegraphics[scale=0.27]{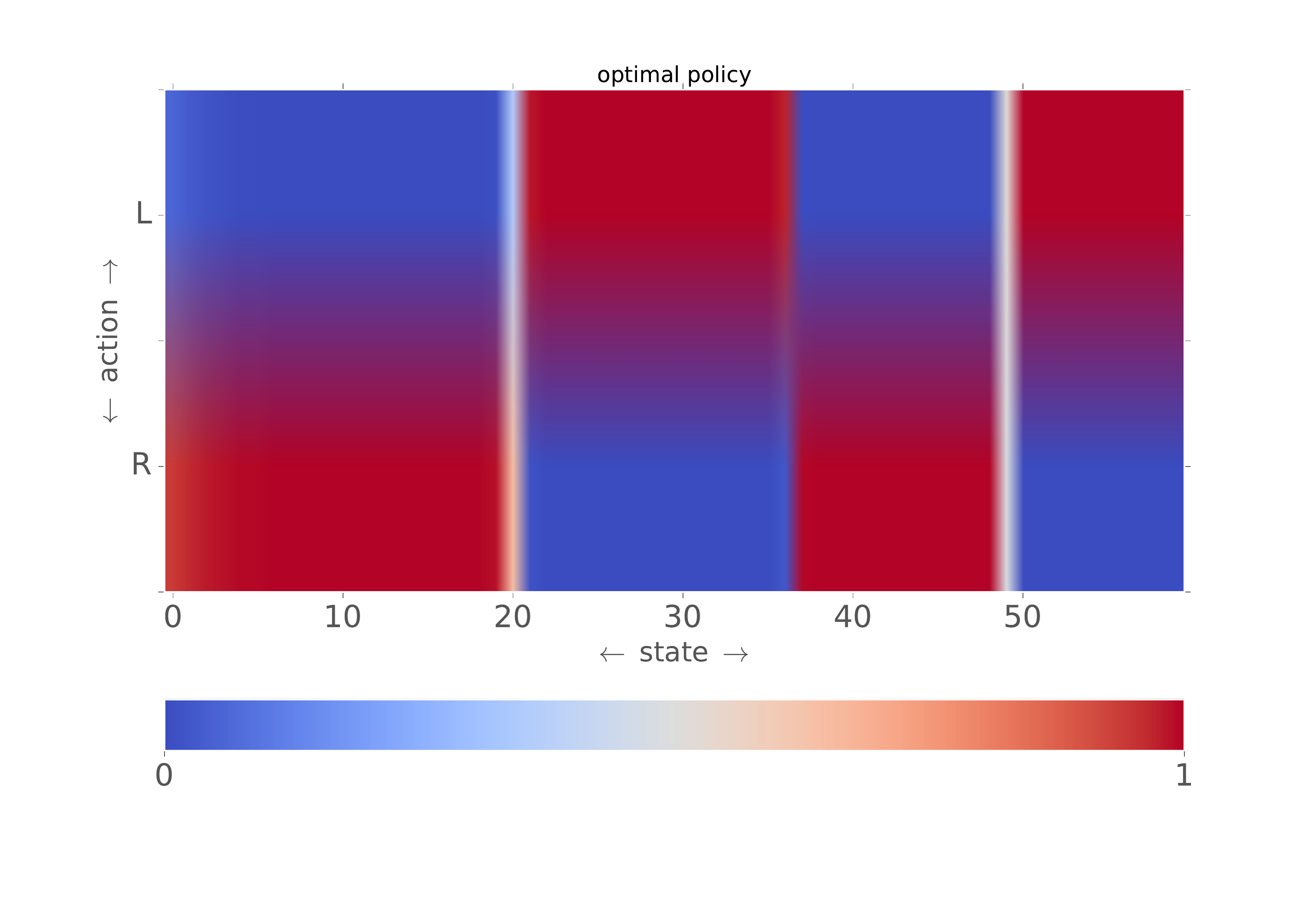}
	\caption{The schematic diagram of the optimal policy generated by Algorithm $2$ for the chain walk MDP with $\vert \mathbb{S} \vert = 60$, $\mathbb{A} = \{L, R\}$ and the discount factor $\gamma = 0.99$.}\label{fig:resgamma99}
\end{figure}

\begin{figure}[!h]
	\begin{subfigure}{0.4\textwidth}
	\vspace*{5mm}\includegraphics[height=45mm, width=44mm]{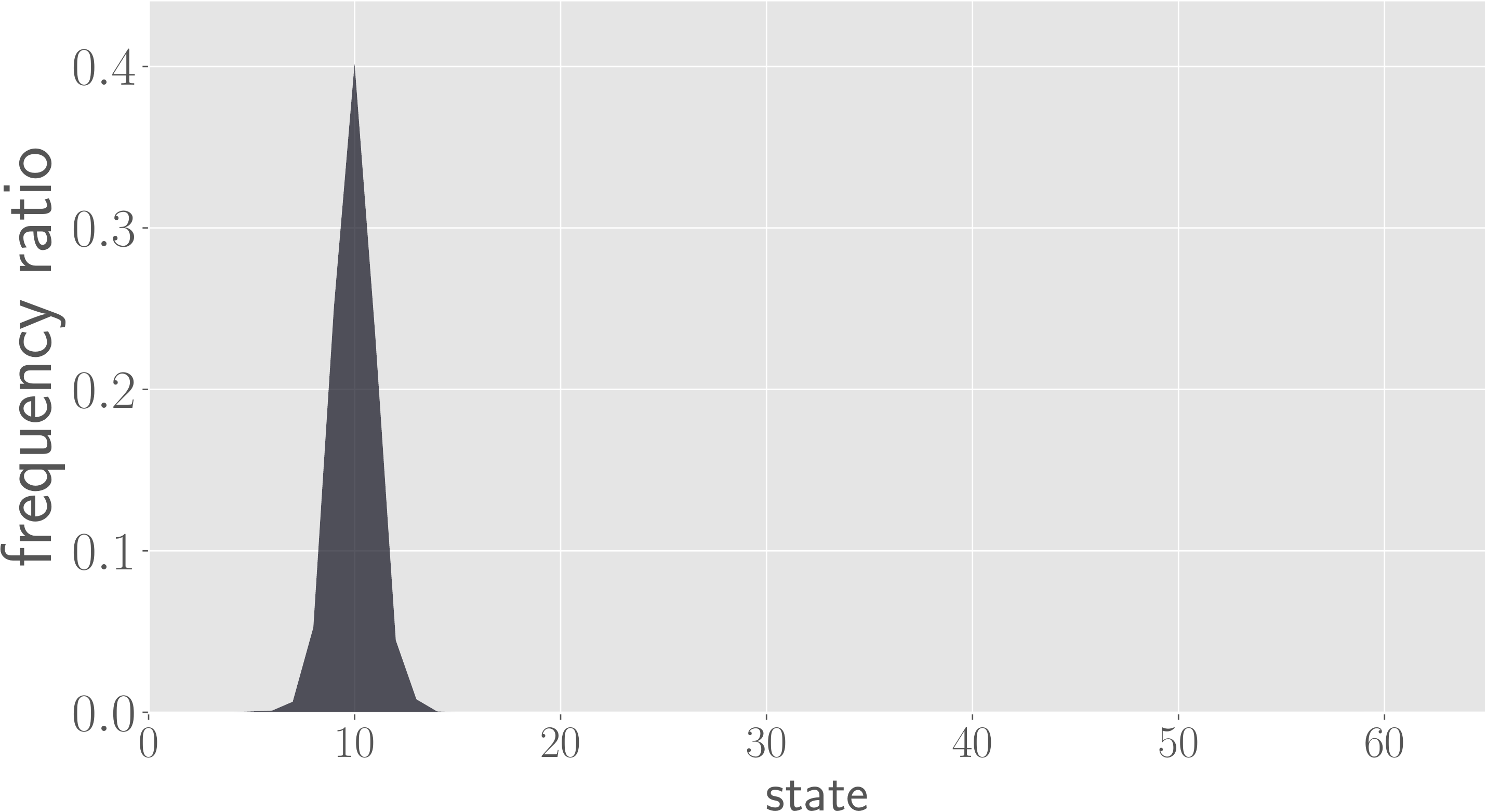}
	\subcaption{Frequency ratio of the states in the sample trajectory}
\end{subfigure}\hspace*{1mm}
	\begin{subfigure}{0.6\textwidth}
	\includegraphics[height=45mm, width=70mm]{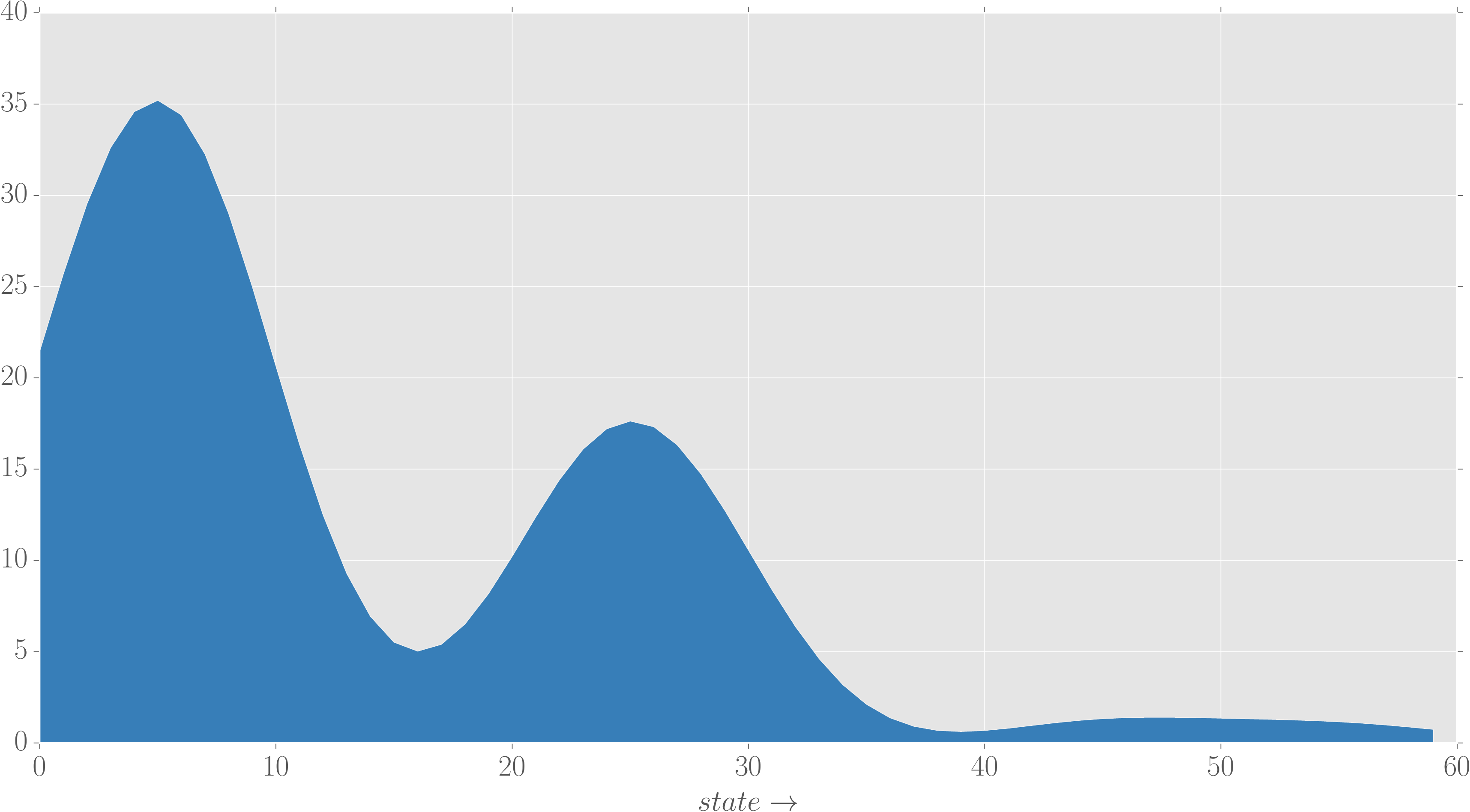}
	\subcaption{Optimal value function generated by Algorithm $2$}
\end{subfigure}
\caption{The frequency ratio of a particular state in the sample trajectory is defined as the ratio of the number of occurrences of that state in the sample trajectory to the total number of state transitions in the sample trajectory. For an ergodic Markov chain, this ratio will eventually converge to is stationary distribution. In this particular example, observe that \textit{the accuracy of the value function is better for states whose relative frequency is good.}}\label{fig:rfreqleft}
\end{figure}

\begin{figure}[!h]
	\begin{subfigure}{0.4\textwidth}
		\vspace*{5mm}\includegraphics[height=40mm, width=44mm]{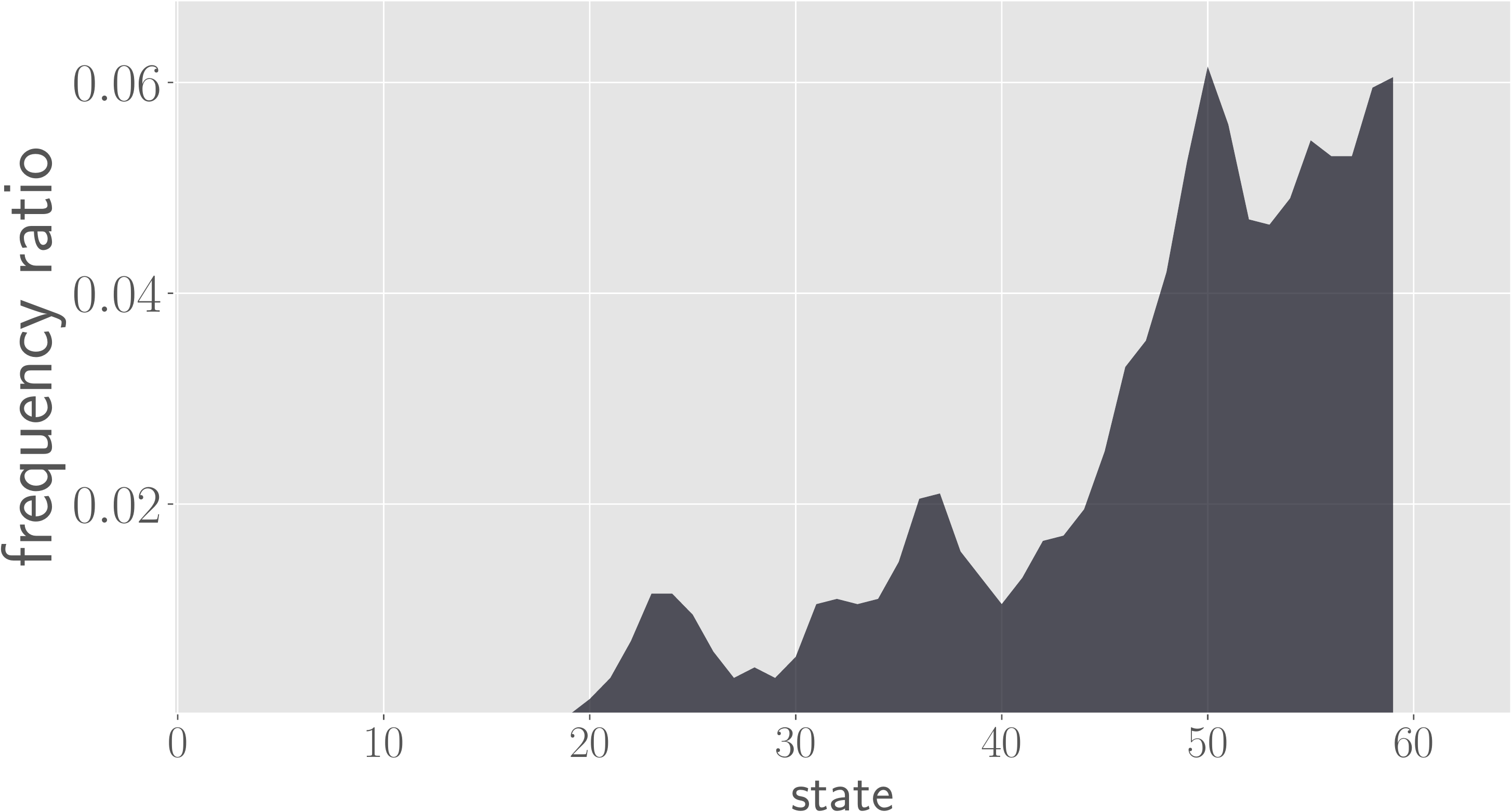}
		\subcaption{Frequency ratio of the states in the sample trajectory}
	\end{subfigure}\hspace*{1mm}
	\begin{subfigure}{0.6\textwidth}
		\includegraphics[height=40mm, width=70mm]{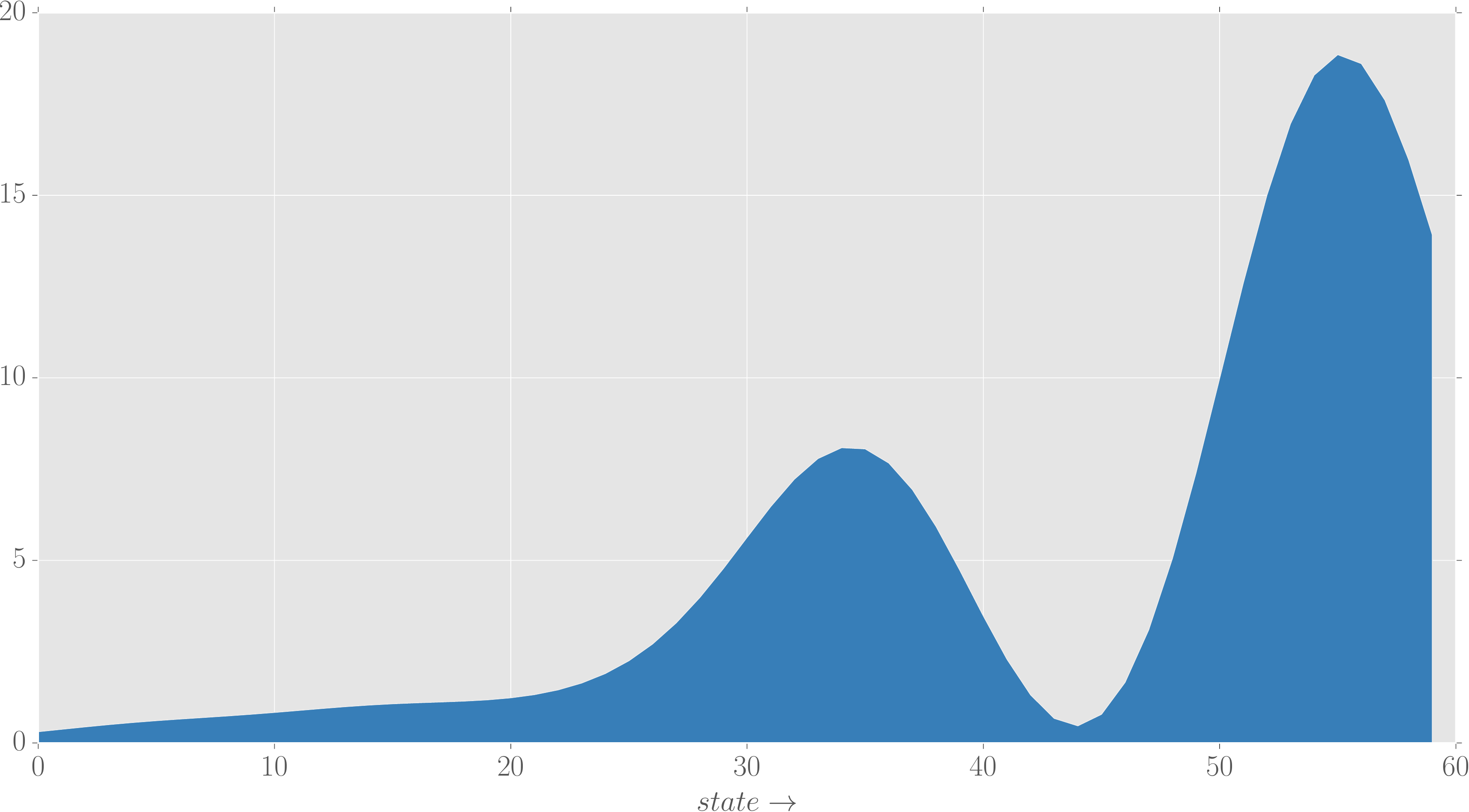}
		\subcaption{Optimal value function generated by Algorithm $2$}
	\end{subfigure}
\caption{In this setting, the relative frequency is better on the right half of the state space and the value function also seems to be more accurate in that region.}\label{fig:rfreqright}
\end{figure}

 In Experiments $2$, $3$ and $4$, we surmised the behaviour policy based on more than a passable knowledge of the MDP.  To make the comparison unbiased (since our algorithm utilized prior information about the MDP to induce the behaviour policy), in the algorithms (MRAS, fast policy search and SPSA) to which our method is being compared, we employed the more accurate on-policy approximation which requires the generative model. This is contrary to our method, where off-policy approximation is tried. Our algorithm  exhibited as good a performance as the state-of-the-art methods in the cart-pole experiment and noticeably the finest performance in the actuated pendulum experiment. This is regardless of the fact that our algorithm is primarily designed for the discrete, finite MDP setting, while the cart-pole experiment and the actuated pendulum experiment are MDPs with continuous state and action spaces. The suboptimal performance of the fast policy search and MRAS is primarily attributed to the insufficient sample size. But the underlying computing machine which we consider for the experiments is a $64$-bit Intel i$3$ processor with $4$GB of memory. Because of these limited resources, there is a finite limit to which the sample size can be scaled. This illustrates the effectiveness of our approach on a resource restricted setting. Now regarding the random MDP experiment, the performance of our algorithm is on par (in fact superior) to the state-of-the-art schemes.\vspace*{4mm}\\ 
($2$) The significance of these results is further strengthened by the fact that all the baseline algorithms considered in the experiments have access to the generative-model and the outcome depicted above is obtained after processing a bevy of sample trajectories. This is contrary to our method where such a privilege is not conferred.\vspace*{5mm}\\
($3$) The algorithm does not seem to be heavily dependent on the discount factor $\gamma$. To corroborate the claim, we show here the performance of the algorithm for two different, yet extreme values of $\gamma$, \emph{i.e.}, for $\gamma \in \{0.01, 0.99\}$ on the chain walk MDP with $60$ states. Here, only the transitions to states $20$ and $40$ incur a positive cost, while the rest are null transitions. The optimal policies generated by our algorithm in the two cases are shown in Figs. \ref{fig:resgamma01} and  \ref{fig:resgamma99} respectively. As one can observe, for $\gamma = 0.99$, the window around state $20$ is wider than that for $\gamma = 0.01$. This is the expected behaviour since the discount factor controls the relative weights of future transitions while evaluating the discounted value function. However, note that this is not the case with regards to state $40$. This lack of accuracy in the final third primarily due to the fact that the behaviour policy we consider in this setting has its stationary distribution heavily concentrated on the first half of the state space. This particular scenario thus also illustrates the dependency of behaviour policy on the accuracy of the solution generated by our algorithm. This is indeed revealed in Theorem \ref{thm:main}. To exemplify it further, we show here how the relative frequency of the states in the given trajectory generated using the behaviour policy determines the accuracy of the solution of our algorithm. Remember that the relative frequency of the states in the sample trajectory is indeed decided by the stationary distribution of the Markov chain induced by the behaviour policy. The results are shown in Figs. \ref{fig:rfreqleft} and \ref{fig:rfreqright}.

\noindent
\vspace*{2mm}\\
($4$) Finally, in the experiments, we found that the parameter which required the highest tuning is $\beta_j$ which is also intuitive since $\beta_{j}$ controls most of the stochastic recursions. The other parameters required minimum tuning with almost all of them taking common values.
\subsection{Data efficiency}
\begin{figure}[!h]
	\begin{subfigure}[h]{0.5\textwidth}
	    \fbox{\includegraphics[width=0.95\textwidth, height=44mm]{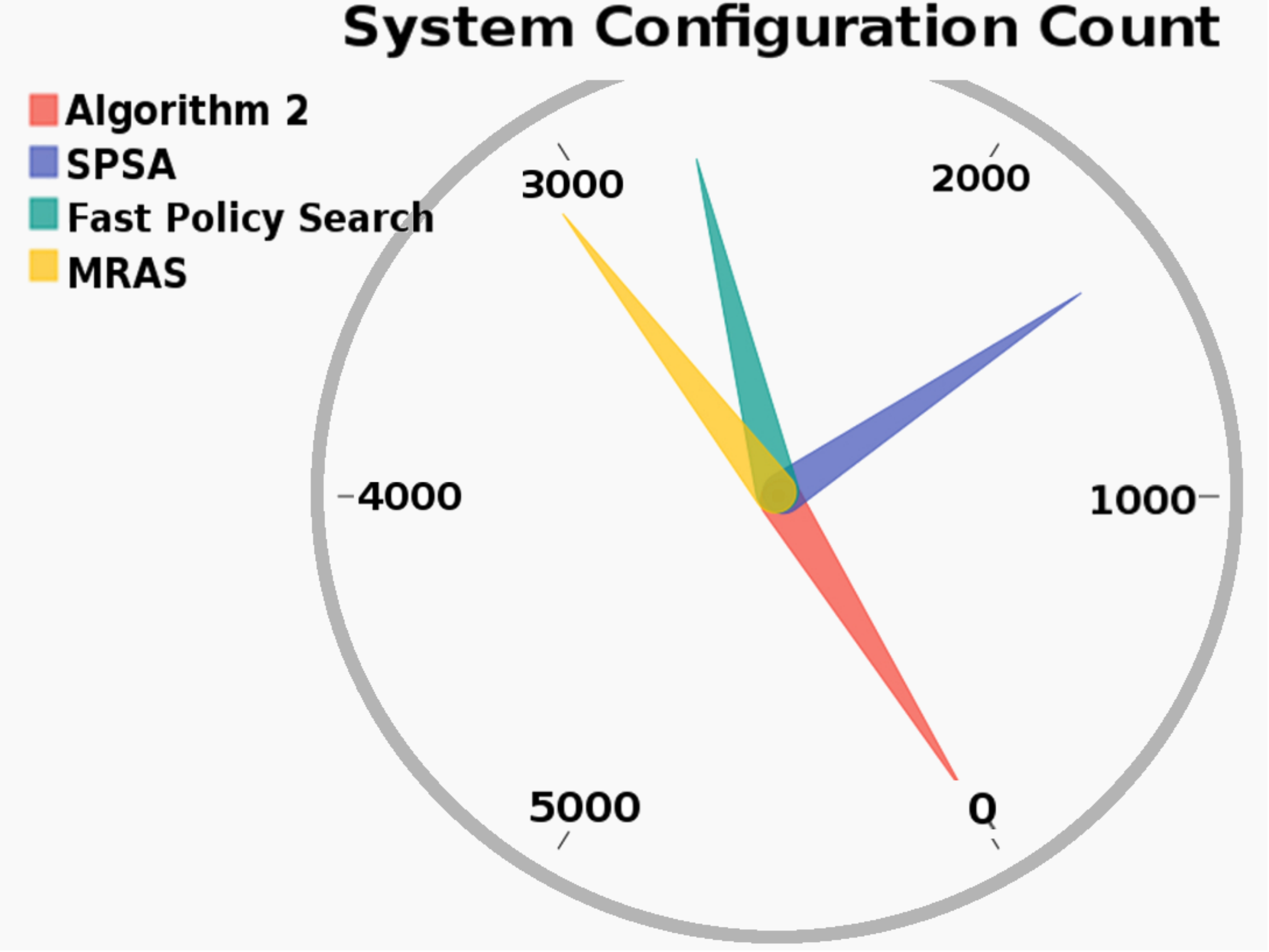}}
    \end{subfigure}
    \begin{subfigure}[h]{0.5\textwidth}
        \fbox{\includegraphics[width=0.93\textwidth, height=44mm]{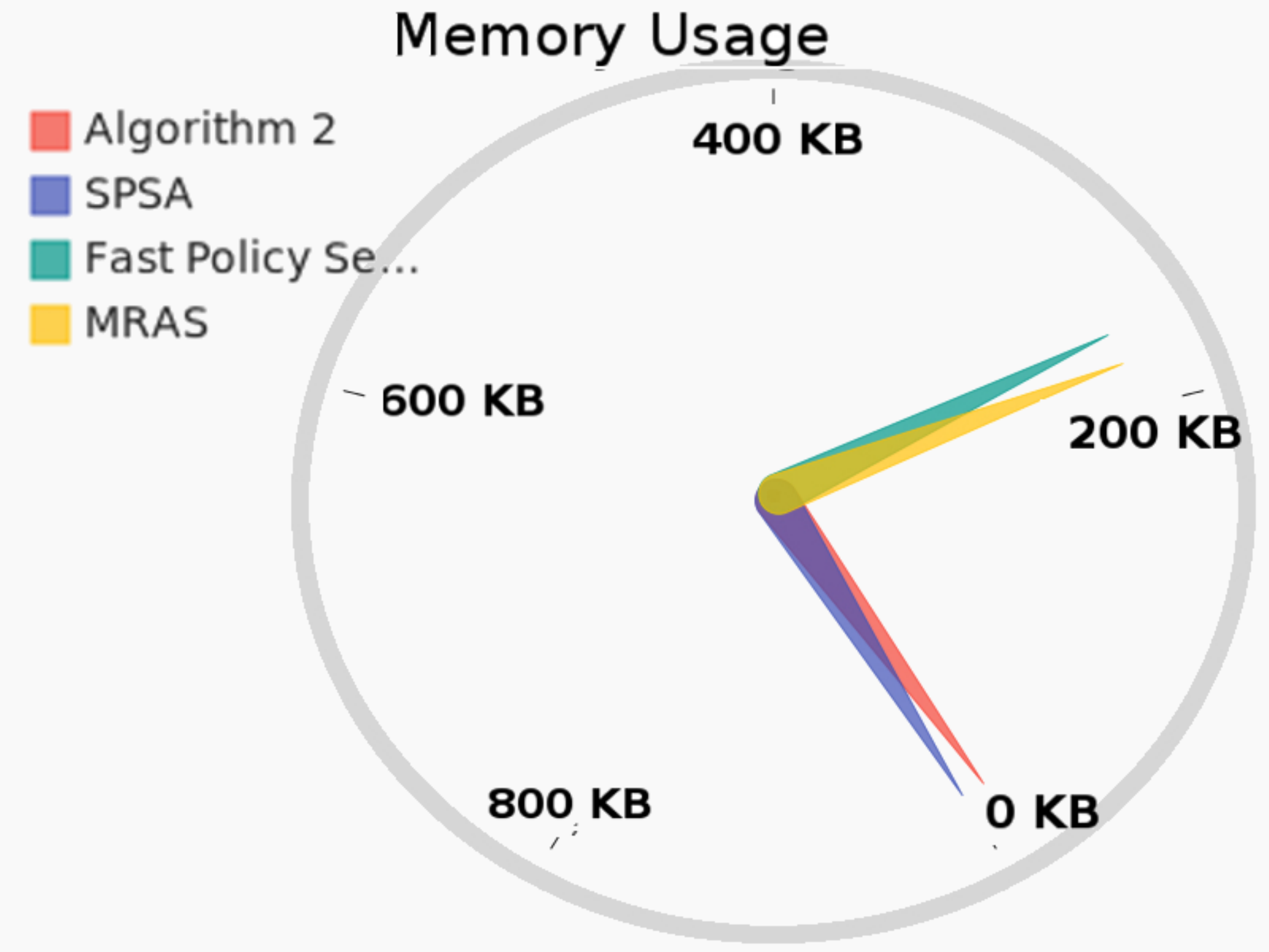}}
    \end{subfigure}
    \caption{Efficiency comparison of Algorithm $2$ \emph{w.r.t.} the state-of-the-art methods.}\label{fig:effcomp}
\end{figure}

\begin{wrapfigure}{r}{0.6\textwidth}
	\vspace{-20pt}
	\begin{center}
		\includegraphics[width=0.6\textwidth, height=55mm]{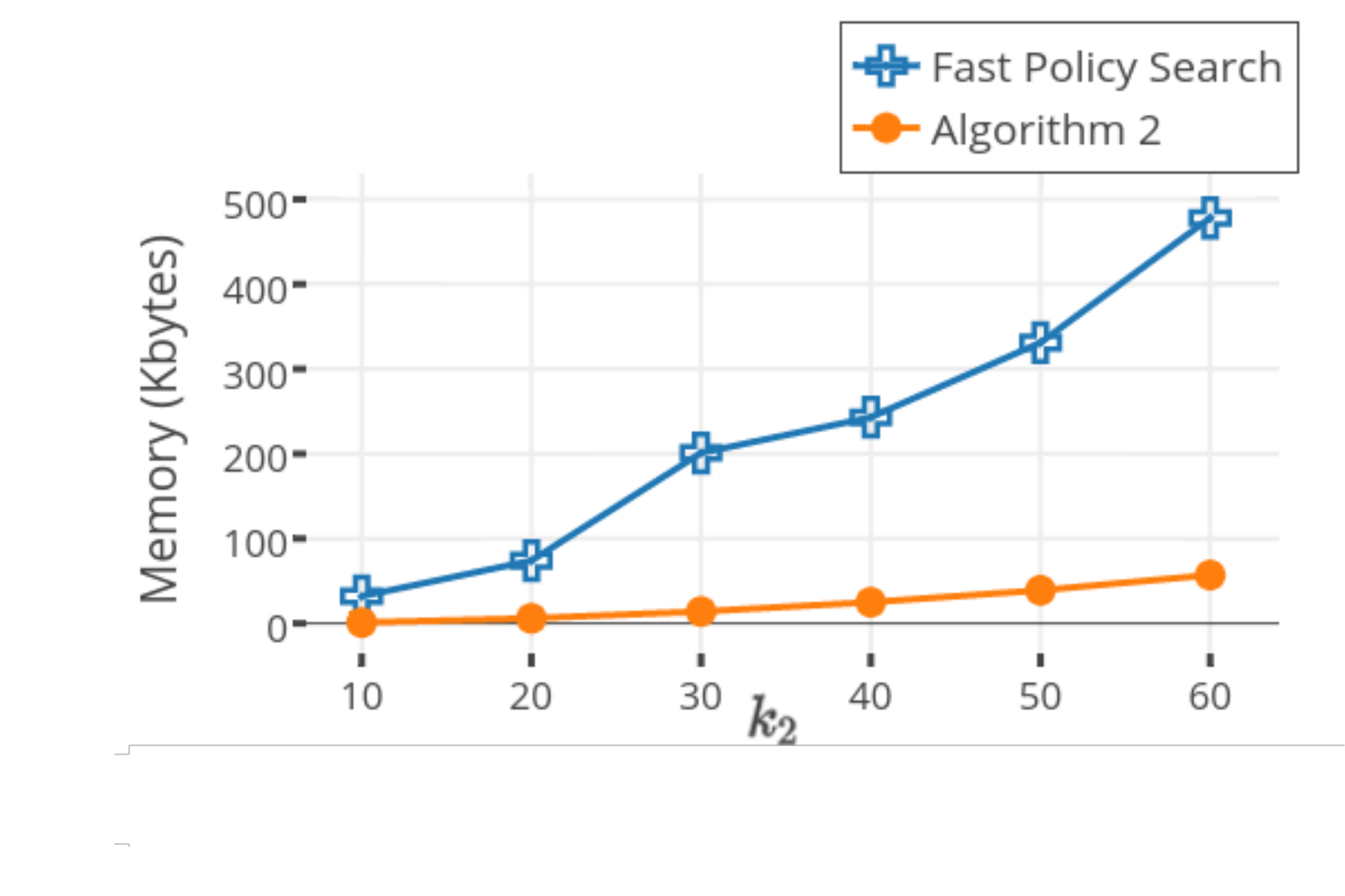}
	\end{center}
	\vspace{-40pt}
	\caption{Memory usage \emph{w.r.t.} $k_2$}\label{fig:k2vsmem}
	\vspace{-10pt}
\end{wrapfigure}

Here, we compare the efficiency of our algorithm with respect to the state-of-the-art algorithms. To measure the efficiency, we consider two benchmarks: \textit{system configuration count} and \textit{memory usage}. The system configuration count denotes the number of times the algorithm queries the generative model of the MDP with a policy to obtain sample trajectories. Memory usage denotes the average real time memory consumed by the algorithms. The results are shown in Figure \ref{fig:effcomp}. The performance of our algorithm with regard to the above benchmarks is commendable.
We also compare here the average memory usage of the fast policy search algorithm and our algorithm with respect to $k_2$ which is the dimension of the policy space. The results are shown in Fig. \ref{fig:k2vsmem}. The illustration shows that memory usage of our algorithms almost remains constant, however fast policy search is very sensitive to the parameter $k_2$. 
This non-dependency of our algorithm on the dimension of the policy space has a real pragmatic advantage since, as a result of this, our algorithm can be applied to very large and complex MDPs with wider policy spaces where fast policy search and MRAS might become intractable.

Another advantage of our approach is the application on legacy systems. In such systems, the information on the dynamics of the system in the form of bits or bytes or paper might be hard to find. However, human experience through long time interaction with the system is available in most cases. Utilizing this human experience to develop a generative model of the system might be hard, however using it to find a behaviour policy which can give average performance is more plausible, and which in turn can be exploited using our algorithm to find an optimal policy.
\section{Conclusion}
We presented an algorithm which solves the modified control problem in a model free MDP setting. We showed its convergence to the global optimal policy relative to the choice of the behaviour policy. The algorithm is data efficient, robust, stable as well as computationally and storage efficient. Using an appropriately chosen behaviour policy, it is also seen to consistently outperform or is competitive against the current state-of-the-art (both) off-policy and on-policy methods.


\bibliographystyle{spmpsci}      

\bibliography{mljrnlbib}
%
%

\end{document}